\PassOptionsToPackage{table,xcdraw}{xcolor}

\documentclass[10pt,logo,copyright]{nvidiatechreport}

\usepackage{subcaption}
\usepackage{makecell}
\usepackage{multirow}
\usepackage{cuted}    
\usepackage{needspace} 

\usepackage[numbers, sort]{natbib}
\definecolor{citecolor}{HTML}{0071BC}
\definecolor{linkcolor}{HTML}{ED1C24}

\usepackage{nicefrac}

\usepackage[capitalize]{cleveref}
\crefname{section}{\S}{\S\S}
\crefname{subsection}{\S}{\S\S}
\crefname{table}{\text{Tab.}}{\text{Tab.}}
\Crefname{table}{Table}{Tables}
\crefname{figure}{\text{Fig.}}{\text{Fig.}}
\Crefname{figure}{Figure}{Figures}
\crefname{equation}{\text{Eq}}{\text{Eq}}

\usepackage{listings}
\usepackage[percent]{overpic}
\usepackage{wrapfig}
\usepackage{tikz}
\usepackage{ragged2e} 

\usepackage[section]{placeins} 
\usepackage{dblfloatfix} 

\newcommand{\ours}{PiD\xspace}

\newcommand{\imgwithlabel}[3]{%
  \begin{overpic}[width=#1]{#2}
    \put(1,1){\makebox(0,0)[lb]{%
      \tikz[baseline]{\node[fill=white, fill opacity=0.75, text opacity=1, inner sep=1pt]{\scriptsize\color{black}#3};}%
    }}
  \end{overpic}%
}

\newsavebox{\infcosttblbox}
\newcommand{\cmark}{\ding{51}} 
\newcommand{\xmark}{\ding{55}} 

\title{\ours: Fast and High-Resolution Latent Decoding with Pixel Diffusion}

\author{%
Yifan Lu, Qi Wu, Jay Zhangjie Wu, Zian Wang, Huan Ling, Sanja Fidler, Xuanchi Ren\\
\small NVIDIA \\
\small \href{https://research.nvidia.com/labs/sil/projects/pid/}{https://research.nvidia.com/labs/sil/projects/pid/}
}

\begin{document}
\maketitle
\begin{figure}[ht!]
  \vspace{-2.5em}
  \centering
  \includegraphics[width=\linewidth]{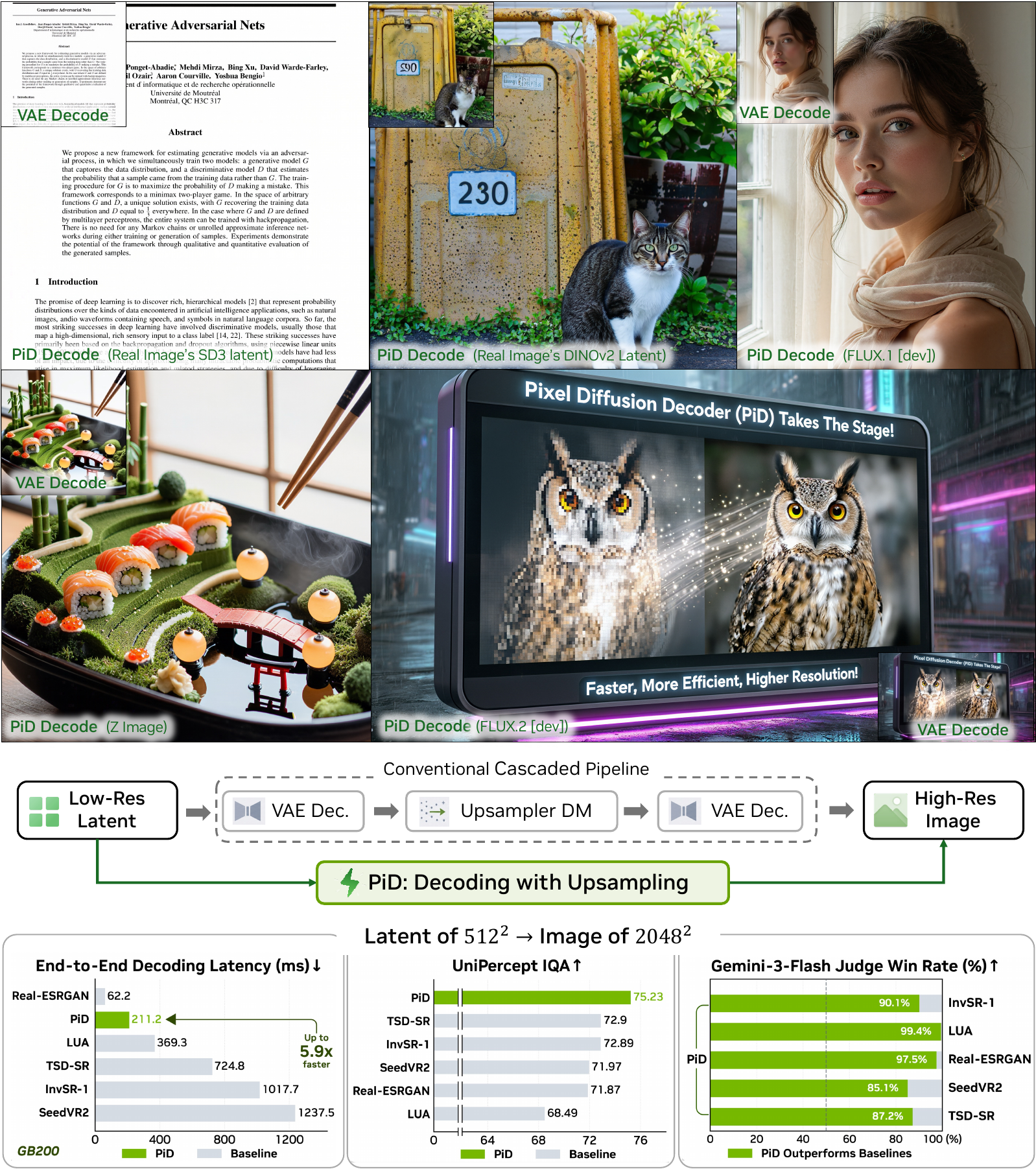}
  \vspace{-1\baselineskip}
  \caption{\ours \textbf{directly} decodes latents from VAE or vision encoders into higher-resolution images, replacing the decode--then--upsample cascade while achieving lower latency and higher visual quality.}
  \label{fig:teaser}
\end{figure}

\begin{figure*}[!t]
  \centering

  \begin{minipage}[t]{0.29\textwidth}
    \vspace{0pt}
    \imgwithlabel{\linewidth}{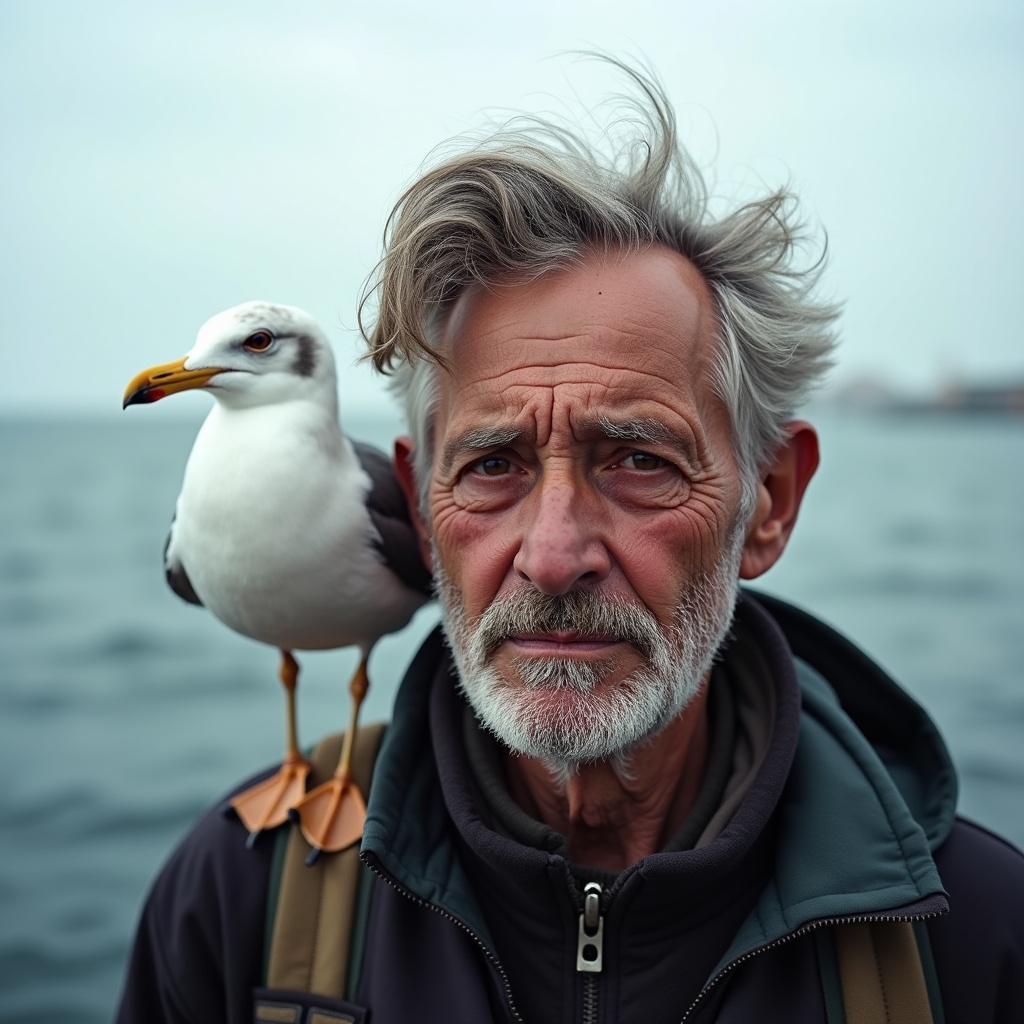}{VAE Decoding $(1024\times 1024)$}
    \vspace{-2pt}
    \centerline{\scriptsize \textit{FLUX.1[dev]}}
    \vspace{1pt}
    {\scriptsize\linespread{0.9}\selectfont\RaggedRight\sloppy\emergencystretch=2em\setlength{\parskip}{0pt}\setlength{\parindent}{0pt}%
    Prompt: A close portrait of a weathered fisherman's face beside a perched seagull on his shoulder, gray ocean behind, soft overcast light, simple documentary-style composition.}
  \end{minipage}\hfill
  \begin{minipage}[t]{0.70\textwidth}
    \vspace{0pt}
    \imgwithlabel{\linewidth}{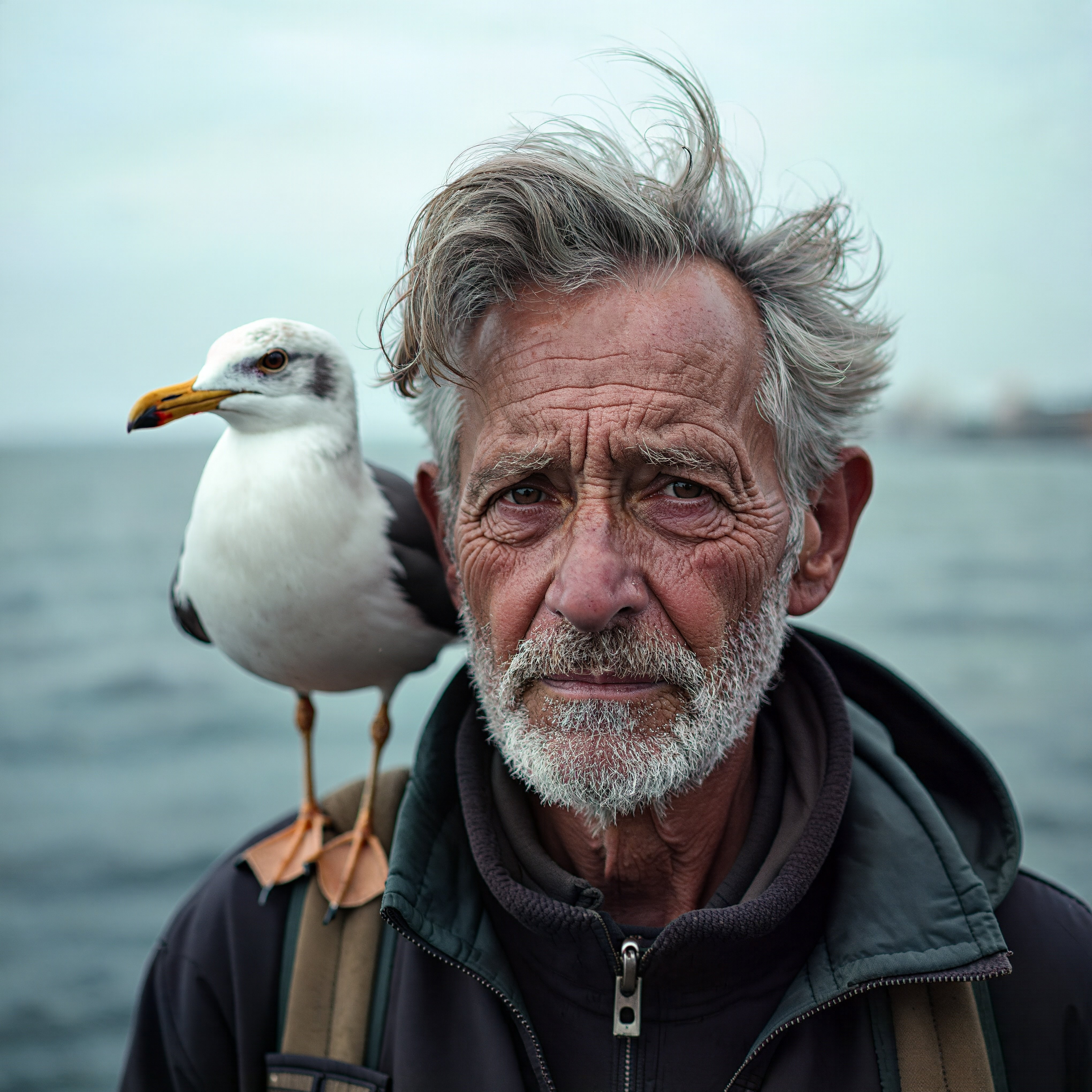}{PiD Decoding $(4096\times 4096)$}
  \end{minipage}

  \vspace{1em}

  \begin{minipage}[t]{0.29\textwidth}
    \vspace{0pt}
    \imgwithlabel{\linewidth}{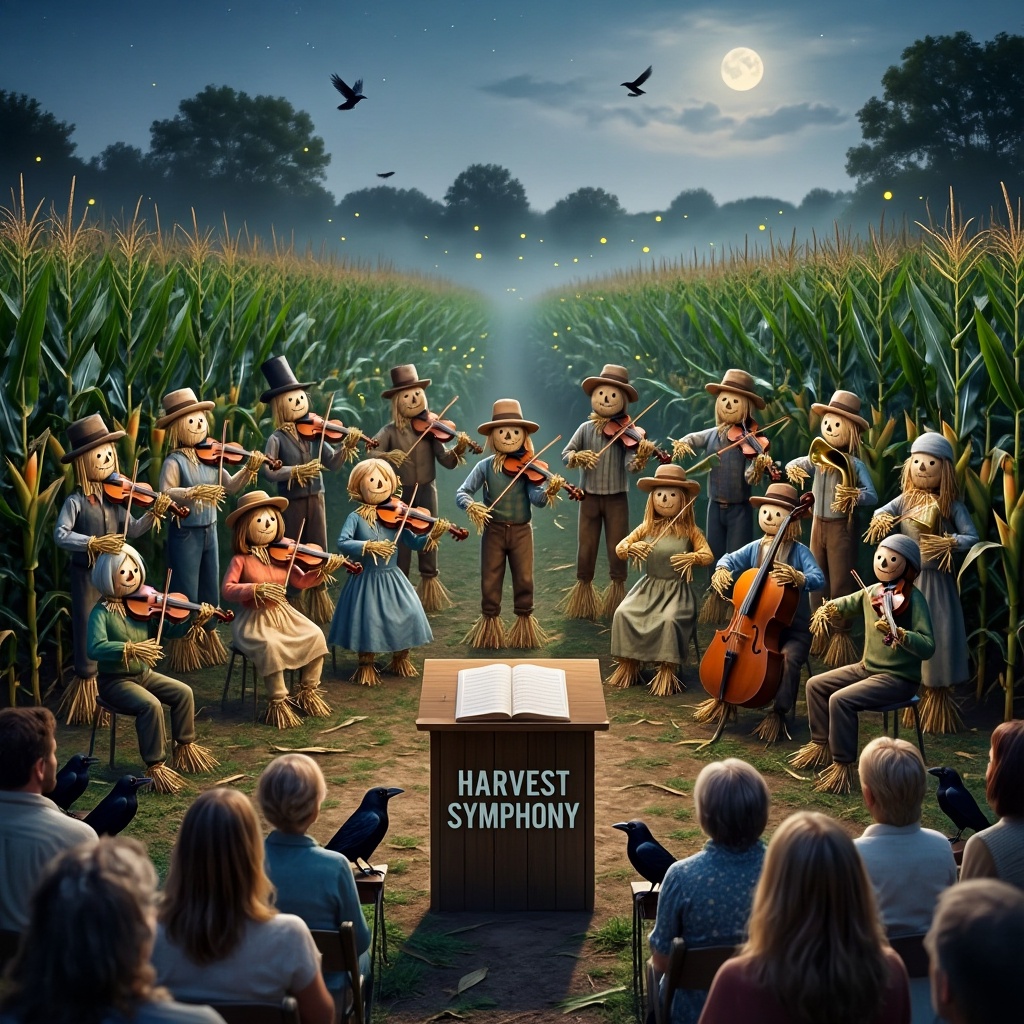}{VAE Decoding $(1024\times 1024)$}
    \vspace{-2pt}
    \centerline{\scriptsize \textit{Z-Image}}
    \vspace{1pt}
    {\scriptsize\linespread{0.9}\selectfont\RaggedRight\sloppy\emergencystretch=2em\setlength{\parskip}{0pt}\setlength{\parindent}{0pt}%
    Prompt: A cinematic rural scene where scarecrows gather at night for a secret orchestra concert in a cornfield, straw hands playing violins and brass, moonlit rows as aisles, crows sitting respectfully in the audience, conductor's stand labeled ``HARVEST SYMPHONY'', mist, fireflies, gentle spooky charm without horror, richly detailed realism.}
  \end{minipage}\hfill
  \begin{minipage}[t]{0.70\textwidth}
    \vspace{0pt}
    \imgwithlabel{\linewidth}{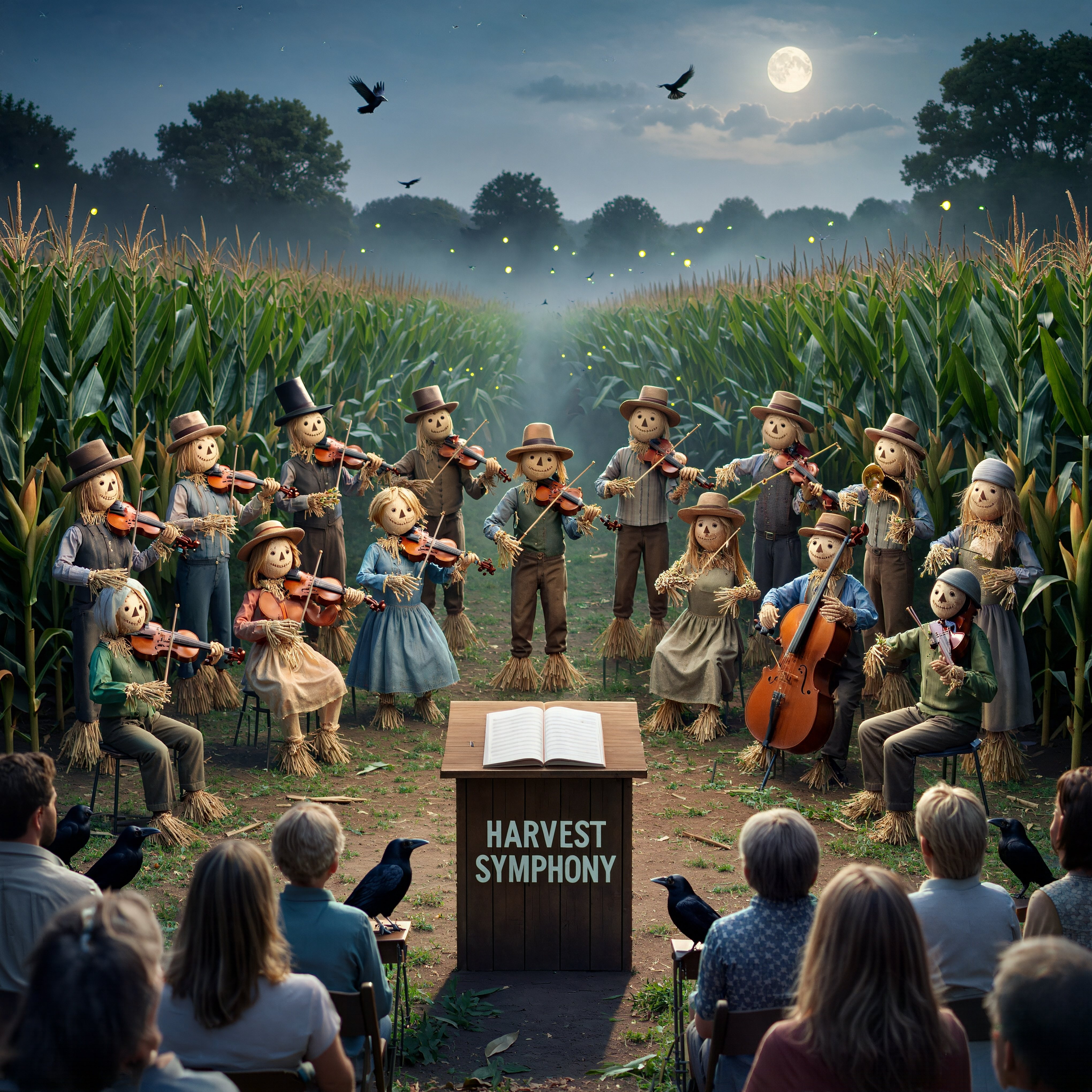}{PiD Decoding $(4096\times 4096)$}
  \end{minipage}

  \caption{\textbf{4K decoding results}. PiD synthesize more details at 4k resolution.}
  \label{fig:4k-decoding}
  \vspace{-1em}
\end{figure*}

\clearpage
\begin{abstract}
Most practical high-resolution text-to-image systems, including latent diffusion and autoregressive models, perform generation in a compact latent space, and a decoder maps the generated latents back to pixels. Yet the latent-to-pixel decoder is reconstruction-oriented, optimized to invert the encoder rather than synthesize more details, and becomes increasingly costly at megapixel scale. This drawback calls for a more expressive and efficient decoding paradigm. Motivated by recent progress in scalable pixel-space diffusion, we introduce \textbf{\ours}, a \textbf{Pi}xel diffusion \textbf{D}ecoder that reformulates latent decoding as conditional pixel diffusion, unifying decoding and upsampling into one generative module. By denoising directly in high-resolution pixel space, \ours synthesizes \(4\times\) and even \(8\times\) upscaled images with low latency. For latent conditioning, a lightweight sigma-aware adapter injects noise-corrupted latents into the pixel diffusion backbone, enabling \ours to decode partially denoised latents and terminate the latent diffusion process early.
To further improve efficiency, we distill the model using DMD2, reducing inference to just $4$ steps. \ours applies to both conventional VAE latents and semantic latents (e.g., SigLIP, DINOv2) used in recent RAE-based models. \ours decodes latents of $512{\times}512$ images into $2048{\times}2048$ pixels in under $1$ second with $13$ GB peak memory on a consumer RTX 5090, and as fast as $210$ ms on a GB200 GPU, about $6{\times}$ faster than cascaded diffusion-based super-resolution pipelines with better visual fidelity.

\end{abstract}

\abscontent

\vspace{-6pt}
\section{Introduction}
\label{sec:intro}
\vspace{-6pt}

Latent diffusion models (LDMs) have become the standard paradigm for text-to-image generation~\citep{rombach2022high}. 
They denoise in a compact latent space, which is more computationally efficient and statistically easier than working in the pixel domain, followed by a decoder that maps the generated latent to pixels.
Beyond diffusion models, modern autoregressive image generators~\citep{yu2022parti,sun2024llamagen,tian2024var,li2024mar} also generate images through visual tokens that must ultimately be decoded back to pixels.
As the only pathway from latent space back to image space, this latent-to-pixel interface is a key factor in final visual fidelity.

Despite its importance, the decoder in LDMs has received far less attention than the diffusion backbone. Most pipelines adopt a convolutional variational autoencoder~\citep{kingma2014autoencoding,rezende2014stochastic} (VAE), but encoder–decoder reconstruction is never perfect, inevitably introducing some inherent losses of fine-grained details. Furthermore, since traditional VAE decoders are designed solely to recover stored information in latents, they tend to pass through or even amplify artifacts in the generated latents rather than correcting them.
These limitations are particularly evident in representation autoencoders (RAEs)~\citep{zheng2025rae}, where semantically rich latents preserve high-level structure but under-specify low-level appearance, requiring the decoder to synthesize missing texture—a capability that standard reconstruction-based decoders lack.

Recent work has begun to treat decoding as a generative problem.
$\epsilon$-VAE~\citep{zhao2024epsilonvae} and SSDD~\citep{vallaeys2025ssdd} replace conventional VAE decoding with pixel-space denoising and distill it into fast single-step decoders. 
However, these approaches remain limited: they do not scale to high-resolution generation due to network architecture, remain dominated by reconstruction objectives, and still lack enough generative capacity to rectify latent-level artifacts or synthesize missing high-frequency details.

At the same time, recent pixel-space diffusion models~\citep{jit2025,yu2025pixeldit} show the power of generating high-resolution images directly in pixel space, excelling at fine detail synthesis. 
This suggests a natural hybrid: use the sampled latent as a structural condition, and use a strong pixel-space diffusion model as the decoder-side generative prior.
In this formulation, the latent preserves the base model's global layout and semantics, while the pixel prior supplies the high-frequency appearance details that reconstruction-oriented decoders struggle to generate.

In this work, we introduce \textbf{\ours}, a \textbf{Pi}xel diffusion \textbf{D}ecoder that reformulates latent decoding as conditional pixel diffusion. \ours unifies visual decoding and super-resolution in a single generative module. By directly denoising in target high-resolution image space, it generates $4{\times}$ or even $8{\times}$ higher-resolution images with low latency.
To incorporate latent conditioning, we introduce a lightweight adapter that injects noise-corrupted latents into the pixel diffusion backbone through a sigma-aware gate. This noise-aware conditioning allows our diffusion decoder to handle both fully denoised and partially denoised latents.
We further distill the model with DMD2~\cite{yin2024improved} technique, reducing inference to only \(4\) denoising steps. This streamlined decoding strategy supersedes the conventional high-resolution cascade---low-resolution VAE decoding, super-resolution diffusion, and high-resolution VAE decoding---with lower latency, lower memory, and better visual fidelity.

\ours decodes a latent corresponding to $512{\times}512$ into a $2048{\times}2048$ image in under $1$ second with $13$ GB peak memory on a consumer RTX 5090, and in $210\,$ms on a GB200 GPU, delivering superior speed and visual quality compared with the cascaded baseline. Additionally, \ours can consume partially denoised latents, enabling early termination of the base latent diffusion process. Beyond VAE latents, \ours generalizes to semantic latents such as DINOv2 features and is particularly well matched to RAE: where RAE latents retain strong semantics but underdetermine fine appearance, \ours supplies the missing generative capacity at decoding time. We further demonstrate that \ours can scale up to 4K resolution decoding while generating more fine-grained details.

In summary, our contributions are as follows:
\begin{itemize}[leftmargin=*,itemsep=2pt,topsep=2pt]
    \item We introduce \textbf{\ours}, a pixel diffusion decoder that reformulates latent decoding as conditional pixel generation, unifying decoding and high-resolution upsampling in a single diffusion decoding stage.

    \item We propose a lightweight sigma-aware latent adapter with noisy latent condition training, allowing early termination of LDM models and decoding on partially denoised latents.
    
    \item We show that \ours extends beyond VAE latents to semantic representations such as DINOv2 features, providing generative, high-resolution decoding capability for RAE-style diffusion models.
    
    \item We demonstrate that \ours outperforms the conventional super-resolution pipeline in both latency and visual quality, decoding $2048{\times}2048$ images in under $1$ second with $13$ GB peak memory on a consumer RTX 5090 and in $210\,$ms on a GB200 GPU.
\end{itemize}

\vspace{-6pt}
\section{Related Work}
\label{sec:related}
\vspace{-6pt}

\noindent\textbf{Latent representations for image generation.}
Latent diffusion models~\citep{rombach2022high} generate images in the latent space of an autoencoder, greatly reducing training and inference cost of the diffusion network.
Most existing systems use reconstruction-oriented autoencoders, such as VAEs~\citep{kingma2014autoencoding,rezende2014stochastic}, VQ-VAEs~\citep{van2017neural}, and more recent tokenizer and autoencoder designs~\citep{chen2024deep,yu2024image,yu2023language}, further improving this paradigm by increasing compression efficiency and reconstruction fidelity. 
Beyond latent diffusion, autoregressive image generators including Parti~\citep{yu2022parti}, LlamaGen~\citep{sun2024llamagen}, VAR~\citep{tian2024var}, and MAR~\citep{li2024mar} likewise model discrete or continuous visual latents produced by such autoencoders, and rely on the same autoencoder decoder to map the generated latents back to pixels.
More recently, works such as VFM-VAE~\citep{bi2025vision}, RAE~\citep{zheng2025rae}, and Scale-RAE~\citep{tong2026scalerae} replace reconstruction-centric encoders with pretrained vision encoders such as DINOv2~\citep{dinov2} and SigLIP~\citep{siglip}, producing semantically richer latent representations.
These approaches shift the role of the latent space: reconstruction autoencoders emphasize pixel-level invertibility, whereas pretrained vision encoders preserve stronger semantic structure while under-specifying fine appearance.
Across all of these settings, generated latents must ultimately be converted back to pixels through a decoder that is largely reconstruction-oriented.
Our work is complementary to advances in latent representation learning: rather than designing a new latent space, we improve the latent-to-pixel interface through latent-conditioned generative decoding that applies to latents produced by either diffusion or autoregressive generators.

\noindent\textbf{Diffusion decoders.}
Several works replace conventional reconstruction decoders with generative diffusion processes in pixel space.
DiVAE~\citep{wu2022divae}, $\epsilon$-VAE~\citep{zhao2024epsilonvae}, and SSDD~\citep{vallaeys2025ssdd} formulate decoding as iterative denoising to improve reconstruction quality and downstream generation, while SSDD further distills diffusion decoding into a fast single-step model.
DALL-E~3~\citep{betker2023improving} reports using a diffusion decoder on top of an LDM latent space, and Unified Latents~\citep{heek2026unified} jointly optimizes the encoder, latent prior, and diffusion decoder.
These works indicate diffusion models can serve as effective decoders, but they still remain primarily reconstruction-oriented: they focus on same-resolution decoding, are evaluated at relatively low resolutions, and remain tied to separate super-resolution stages in high-resolution pipelines.
In contrast, \ours incorporates a pixel diffusion generative prior into the decoder and formulates decoding and upsampling as a single conditional generation task.

\noindent\textbf{High-resolution image synthesis.}
A widely used strategy for high-resolution image synthesis is cascaded super-resolution, which decomposes generation into a base model followed by one or more upsampling stages.
Classical methods such as ESRGAN~\citep{wang2018esrgan} and Real-ESRGAN~\citep{wang2021realesrgan} use adversarial and perceptual objectives for visually sharp outputs, while diffusion-based cascades~\citep{ho2022cascaded,saharia2021sr3,yue2023resshift,wang2023stablesr,lin2024diffbir,wu2024seesr,yu2024supir} leverage large generative priors for higher-fidelity texture synthesis.
Recent distilled variants~\citep{yue2025arbitrary,dong2025tsd,kawai2025efficient,sun2025one,wu2026one,arora2025guidesr,you2025consistency} reduce sampling cost, and latent-space approaches such as LUA~\citep{razin2025one} and LSRNA~\citep{jeong2025latent} avoid intermediate pixel decoding.

In parallel, recent work explores direct high-resolution generation.
PixArt-$\Sigma$~\citep{chen2024pixartsigma}, SANA~\citep{xie2024sana}, and UltraFlux~\citep{ye2025ultraflux} improve scalable high-resolution synthesis through advances in training, architecture, and compression.
Recent pixel-space diffusion transformers such as JiT~\citep{jit2025} and PixelDiT~\citep{yu2025pixeldit} further demonstrate that raw-pixel generation can scale to high resolutions while synthesizing fine-grained detail.
In contrast, \ours unifies latent decoding and spatial upsampling within a single latent-conditioned pixel diffusion model, directly using a scalable pixel diffusion prior as the decoder.

\vspace{-6pt}
\section{Pixel Diffusion Decoder}
\label{sec:method}
\vspace{-6pt}

\begin{figure}
    \centering
    \includegraphics[width=\linewidth]{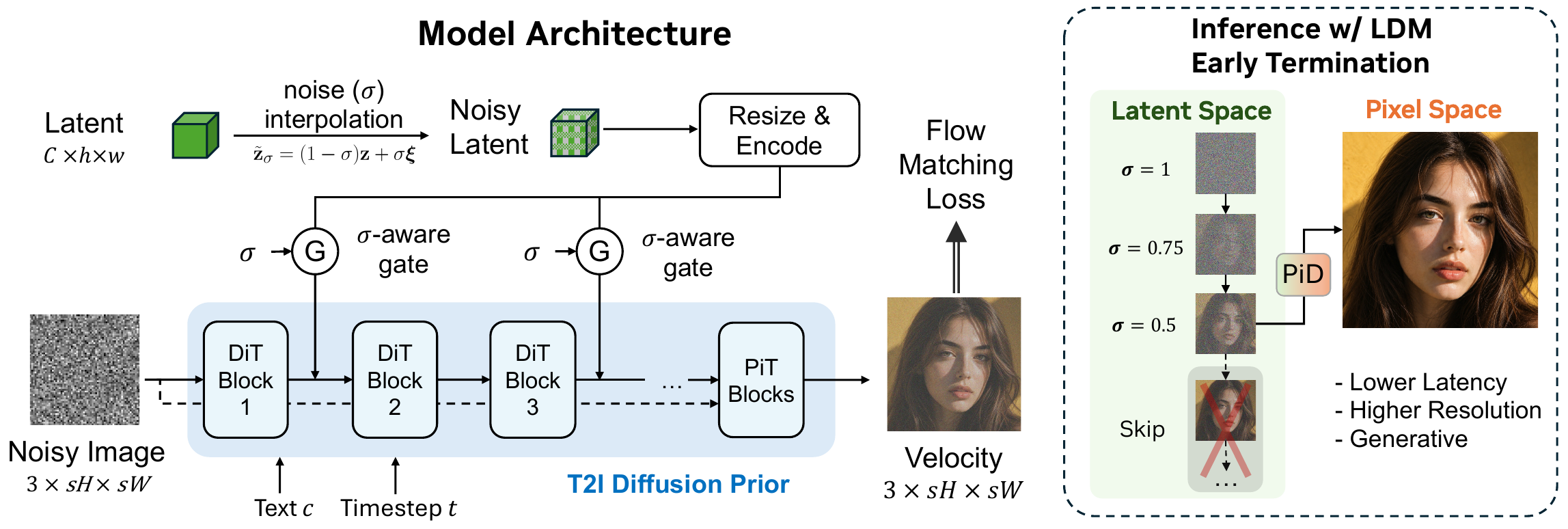}
    \caption{\textbf{Overview of \ours.}
    \ours unifies latent decoding and upsampling as a single latent-conditioned pixel diffusion model that predicts the target-resolution pixel-space velocity field.
    Noise-corrupted latent training and sigma-aware gating make the decoder robust to partially denoised latents, enabling early exit from the base LDM while preserving high-resolution output quality.}
    \label{fig:method}
    \vspace{-1em}
\end{figure}

We reformulate latent decoding as latent-conditioned pixel diffusion.
Instead of first decoding a latent to a low-resolution image and then applying separate super-resolution stages, our model directly generates high-resolution images from sampled latents with a pixel-space text-to-image prior.
This unifies latent decoding and spatial upsampling within a single generative process.

\vspace{-4pt}
\subsection{Problem Formulation}
\label{subsec:method_formulation}
\vspace{-4pt}

Let $\mathbf{z} \in \mathbb{R}^{C\times h\times w}$ denote a sampled latent under text condition $c$ in an autoencoder latent space, such as a VAE or representation autoencoder (RAE) latent space.
Conventional latent diffusion pipelines first decode the latent into an image
$\mathbf{x}_{\mathrm{dec}} = \mathcal{D}(\mathbf{z})\in \mathbb{R}^{3\times H\times W}$ using a decoder $\mathcal{D}$,
and then apply a separate upsampler $\mathcal{U}_s$ to produce higher-resolution outputs:
\begin{equation}
\hat{\mathbf{x}}_0
=
\mathcal{U}_s(\mathbf{x}_{\mathrm{dec}})
\in
\mathbb{R}^{3\times (sH)\times (sW)},
\end{equation}
where $s>1$ is the target upsampling factor.
This formulation decomposes latent decoding and spatial upsampling into separate stages.

In contrast, \ours directly models the target-resolution image distribution with conditional pixel diffusion:
\begin{equation}
\hat{\mathbf{x}}_0 \sim p_{\theta}^{(s)}(\mathbf{x}_0 \mid \mathbf{z}, c), \qquad \mathbf{x}_0 \in \mathbb{R}^{3\times (sH)\times (sW)}.
\end{equation}
In this work, we focus on high-resolution decoding with $s=4$ or $s=8$.
The latent provides global structure and semantic hints, while the pixel diffusion prior synthesizes fine image detail directly at the target resolution.
This replaces native reconstruction and decode-then-upsample cascades with a single conditional pixel-diffusion decoder.

\vspace{-4pt}
\subsection{Latent-Conditioned Pixel Decoder}
\label{subsec:method_stage2}
\vspace{-4pt}

\noindent\textbf{Pixel-space generative prior.} 
To provide strong generative capability for decoding and upsampling, we design \ours to build on top of a text-conditioned pixel diffusion prior.
We adopt PixelDiT~\citep{yu2025pixeldit}, a pixel-space diffusion transformer with an MMDiT-style~\cite{esser2024scaling} backbone, as our base architecture.

Starting from a pretrained $1024{\times}1024$ checkpoint, we further scale the model to 2K resolution and beyond.
Scaling to these resolutions substantially increases the patch-token sequence length, so we replace the original rotary positional encoding (RoPE)~\citep{su2024roformer} with an NTK-aware RoPE variant~\citep{ntk_rope} for improved spatial extrapolation.
The resulting model serves as a high-resolution pixel diffusion prior that is later conditioned on sampled latents.

\noindent\textbf{From pixel prior to latent decoder.}
We convert this high-resolution text-to-image prior into a latent-conditioned pixel decoder by adding a lightweight ControlNet-style~\citep{zhang2023adding} adapter and jointly fine-tuning it with the pixel diffusion backbone.
The adapter injects latent information into the pixel diffusion model while the text-to-image prior from previous stage provides a strong initialization.

At each denoising step, the model takes a noisy target-resolution image $\mathbf{x}_t$, text condition $c$, and latent condition $\mathbf{z}$, and predicts the rectified-flow velocity in pixel space.
The latent condition provides global layout and semantic guidance, while the pixel prior synthesizes high-resolution appearance.

\noindent\textbf{Noisy latent conditioning.}
Rather than conditioning only on clean latents, we expose the decoder to latents corrupted by varying noise levels:
\begin{equation}
\tilde{\mathbf{z}}_{\sigma} = (1-\sigma)\mathbf{z} + \sigma \boldsymbol{\xi},
\qquad
\boldsymbol{\xi} \sim \mathcal{N}(\mathbf{0}, \mathbf{I}),
\qquad
\sigma \sim \mathcal{U}(0, \sigma_{\max}).
\label{eq:latent_noise}
\end{equation}
This serves two purposes.
First, it prevents the decoder from over-trusting the latent, which can suppress generative detail synthesis. 
Second, it exposes the decoder to latents of varying quality, enabling early-exit inference from partially denoised latents as described in Sec.~\ref{subsec:method_stage3}.
The noise level $\sigma$ is also provided to the injection modules so that the decoder can adapt how strongly it uses the latent condition.

\noindent\textbf{Latent projection and injection.}
The pixel diffusion decoder operates on the noisy target-resolution image $\mathbf{x}_t$, patchified into $16\times16$ image patches.
To inject the latent condition, we align the noisy latent $\tilde{\mathbf{z}}_{\sigma}$ to this patch-token grid using a lightweight projection module.
Concretely, $\tilde{\mathbf{z}}_{\sigma}$ is spatially upscaled with nearest-neighbor upsampling to match the patch-token grid, encoded by convolutional residual blocks, flattened into tokens, and linearly projected to the PixelDiT hidden dimension:
\begin{equation}
\hat{\mathbf{z}}_{\sigma}=\operatorname{Resize}(\tilde{\mathbf{z}}_{\sigma}),\qquad
\mathbf{l}_i=\operatorname{Linear}_i\!\left(\operatorname{Flatten}\left(\operatorname{ResBlock}(\hat{\mathbf{z}}_{\sigma})\right)\right),
\end{equation}
where $\operatorname{Resize}(\cdot)$ denotes spatial alignment to the image patch grid and $\operatorname{ResBlock}(\cdot)$ is the convolutional feature extractor.
This produces latent-conditioning tokens $\mathbf{l}_i\in\mathbb{R}^{B\times N\times d}$ aligned with the image tokens at transformer block $i$, where $N$ is the number of image patch tokens and $d$ is the block hidden dimension.
Let $\mathbf{h}_i$ denote the hidden tokens at transformer block $i$. We inject the latent condition every two backbone blocks:
\begin{equation}
\mathbf{h}_i \leftarrow \mathbf{h}_i + g_i(\mathbf{h}_i, \mathbf{l}_i, \sigma) \odot \mathbf{l}_i,
\label{eq:sigma_gate}
\end{equation}
where $g_i(\cdot)$ controls the strength of latent injection.

\noindent\textbf{Sigma-aware gating.}
The reliability of the latent condition depends on its noise level $\sigma$: clean latents provide strong layout and semantic cues, while noisier latents should be trusted less.
We therefore modulate the injection in Eq.~\eqref{eq:sigma_gate} with a sigma-aware gate that outputs a per-token per-channel scalar:
\begin{equation}
 g_i(\mathbf{h}_i, \mathbf{l}_i, \sigma)
 = \operatorname{sigmoid}\!\left(\operatorname{Linear}_i([\mathbf{h}_i,\mathbf{l}_i]) - \alpha\,\sigma\right),
\label{eq:sigmoid_sigma_gate}
\end{equation}
where the first term predicts a content-dependent injection strength and the learned $\alpha>0$ introduces a monotonic sigma-dependent bias that encourages weaker latent injection as the latent becomes noisier.
We zero-initialize the latent injection heads so that training starts from the pretrained pixel prior and gradually learns to use the latent condition.

\vspace{-4pt}
\subsection{Model Training}
\label{subsec:method_training}
\vspace{-4pt}
\noindent\textbf{Training the pixel diffusion prior.}
We first train the high-resolution pixel diffusion prior using a standard rectified-flow~\citep{lipman2022flow,liu2022flow} objective in pixel space.
Given a clean image $\mathbf{x}_0$ and text condition $c$, we sample
$t\sim\mathcal{U}(0,1)$
and construct the noisy image
\begin{equation}
\mathbf{x}_t
=
t\mathbf{x}_0
+
(1-t)\boldsymbol{\epsilon},
\qquad
\boldsymbol{\epsilon}
\sim
\mathcal{N}(\mathbf{0},\mathbf{I}).
\end{equation}

The model predicts the rectified-flow velocity field
\begin{equation}
\mathbf{v}_{\theta}(\mathbf{x}_t,t,c)
\approx
\mathbf{x}_0-\boldsymbol{\epsilon},
\end{equation}
and is optimized with
\begin{equation}
\mathcal{L}_{\mathrm{FM}}
=
\mathbb{E}
\left[
\left\|
\mathbf{v}_{\theta}(\mathbf{x}_t,t,c)
-
(\mathbf{x}_0-\boldsymbol{\epsilon})
\right\|_2^2
\right].
\end{equation}

\noindent\textbf{Training the latent-conditioned decoder.}
Starting from the pretrained pixel prior, we jointly fine-tune the diffusion backbone and latent injection modules using the same rectified-flow objective, now conditioned on the noisy latent $\tilde{\mathbf{z}}_{\sigma}$ and its noise level $\sigma$ defined in Eq.~\eqref{eq:latent_noise}:
\begin{equation}
\mathcal{L}_{\mathrm{FM}}
=
\mathbb{E}
\left[
\left\|
\mathbf{v}_{\theta}
(
\mathbf{x}_t,
t,
c,
\tilde{\mathbf{z}}_{\sigma},
\sigma
)
-
(\mathbf{x}_0-\boldsymbol{\epsilon})
\right\|_2^2
\right].
\end{equation}

During training, the model learns to balance latent fidelity and generative synthesis across different latent noise levels.

\vspace{-4pt}
\subsection{Fast Inference with Distillation and Early Termination}
\label{subsec:method_stage3}
\vspace{-4pt}

\noindent\textbf{Few-step distillation.}
We further accelerate inference by distilling our model with Distribution Matching Distillation (DMD2)~\citep{yin2024improved}.
Starting from the teacher trained in the previous stage, we train a student decoder that performs inference in only four sampling steps.
We additionally distill classifier-free guidance into the student, eliminating the need for separate conditional and unconditional forward passes at inference time.
We retain the noisy latent conditioning strategy so that the student preserves its ability to decode latents at varying noise levels.

\noindent\textbf{Early termination of base latent diffusion}
At inference time, the base latent diffusion model can be stopped before completing all denoising steps, yielding a partially denoised latent with residual noise level $\sigma$.
Note that this latent has the same intermediate-noise form used in noisy latent conditioning (Eq.~\eqref{eq:latent_noise}), so it can be passed directly to \ours for pixel-space decoding.
\section{Experiments}
\label{sec:experiments}

In this section, we introduce implementation details of \ours. We also conduct experiments across diverse latent spaces and base generators to evaluate \ours's visual quality, latency, and generalization ability.

\subsection{Data}
\label{sec:implementation-details-data}

We train on MultiAspect-4K-1M~\cite{ye2025ultraflux}, rendered PDF data, and internally procured high-resolution images.
To improve data quality, we filter low-quality samples with Q-Align~\cite{wu2023qalign}, resulting in $2.6$M high-quality images. We organize image data into separate buckets based on aspect ratio, specifically 16:9, 4:3, 1:1, 3:4, and 9:16. During training, images from each bucket are center-cropped to a fixed resolution: $2048\times2048$ for 1:1, $2304\times1728$ for 4:3, $1728\times2304$ for 3:4, $2688\times1536$ for 16:9, and $1536\times2688$ for 9:16. This allows our model to decode varying aspect-ratio images.

We label three captions per training image: a long caption (\texttt{prompt}, 200--300 words), a medium caption (\texttt{prompt\_medium}, 50--200 words), and a short caption (\texttt{prompt\_short}, $<$50 words). During training, captions of different length are uniformly sampled so the model sees text ranging from detailed scene descriptions to concise one-liners. We use Qwen3-VL-8B-Instruct~\cite{Qwen3-VL} via LMDeploy's TurboMind engine for high-throughput batched captioning.

\subsection{Implementation Details}
\label{sec:implementation-details}

\noindent\textbf{Pixel-space generative prior.}
To have a pixel-space diffusion generator at 2K resolution, we start from the official pretrained PixelDiT checkpoint (1.3B parameters) and finetune it with our dataset. To adapt to higher resolution, we use timestep shift $6$, while the original checkpoint uses shift $4$ for 1K resolution. The PixelDiT backbone's architecture uses patch size 16, hidden size 1536, 24 attention heads, 14 MM-DiT image-text blocks, and 2 PiT pixel blocks for pixel decoding. The PiT branch uses 16-dimensional pixel tokens, attention width 1152, and 16 attention heads. Text conditioning uses the frozen Gemma-2-2B-it~\cite{team2024gemma} encoder with 2304-dimensional text features and maximum sequence length 300. We use NTK-aware RoPE with a $1024\times1024$ reference resolution. We use batch size $128$ and learning rate $2\times10^{-5}$ for $20{,}000$ iterations, which takes about $1$ day on $128$ H100 GPUs.

\noindent\textbf{Latent projection and injection.}
For noisy latent conditioning, we set $\sigma_{\max}=0.8$. The latent adapter is a ControlNet-style 2D convolutional path that converts the noisy low-resolution latent into per-block token features for injection into the PixelDiT backbone.

After resizing latent to patch grid's shape, the latent passes through: (i)~\texttt{Conv2d}($16\!\to\!512$, $3\!\times\!3$, padding $1$); (ii)~\texttt{SiLU}; (iii)~\texttt{Conv2d}($512\!\to\!512$, $3\!\times\!3$, padding $1$); followed by (iv)~four pre-activation residual blocks of the form $\textsc{GN}_{4}\!\to\!\textsc{SiLU}\!\to\!\texttt{Conv}_{3\!\times\!3}\!\to\!\textsc{GN}_{4}\!\to\!\textsc{SiLU}\!\to\!\texttt{Conv}_{3\!\times\!3}$ with a residual skip (all $512$-channel, GroupNorm group count $4$). For $2048^2$ resolution decoding target, it generates $[B,512,128,128]$ feature map, and is then flattened to $[B,16384,512]$ tokens.

\emph{Per-block output heads.} We inject latent-conditioning tokens into every two DiT blocks via an independent head per injection point, while leaving the PiT pixel blocks untouched. Each head is a single $\texttt{Linear}(512\!\to\!1536)$ layer, weight- and bias-zero-initialized so that training starts from the pretrained text-to-image behavior. Tokens from each head then go through the sigma-aware gate as described in Sec.~\ref{subsec:method_stage2}. The sigma-aware gate is initialized with bias $2.0$ and $\alpha\approx5$, formulating $\operatorname{sigmoid}(2 - 5\sigma)$ at the beginning of training.

We jointly fine-tune the pixel diffusion backbone and latent projection modules with batch size $64$ and learning rate $5\times10^{-5}$ for $30{,}000$ iterations, which takes about half a day on $64$ H100 GPUs. We apply 10\% caption dropout and 10\% latent-condition dropout for classifier-free guidance training. For the vision encoder latents, our experiments show that keeping the PixelDiT backbone frozen results in less color drift than fully finetuning it, so we choose to freeze the backbone in these cases. 

\noindent\textbf{Distillation.}
We distill the multi-step latent-conditioned decoder into a $4$-step student using the sigma schedule $\{0.999, 0.866,$ $ 0.634, 0.342\}$. Following DMD2~\cite{yin2024improved}, we add projected GAN regularization on intermediate model features. The discriminator is a DiT with $26$ blocks and hidden dimension $1536$. We set DMD loss weight to 1.0, and denoising score matching loss weight 1.0. The student and fake-score network is initialized from the same teacher architecture and checkpoint, and optimized with learning rate $10^{-5}$ and weight decay $10^{-3}$. We set the GAN loss weight to $0.05$ and the R1 regularization weight to $200.0$. The student, fake score model, and discriminator are trained with AdamW using a constant learning rate of $1\times10^{-5}$ and batch size $16$ for $3{,}000$ iterations, taking about $2$ hours on $128$ H100 GPUs with context parallelism 8.

All training uses mixed precision: forward passes run in bfloat16, while gradients and optimizer states remain in float32. We maintain an exponential moving average (EMA) of model weights and use EMA weight in inference.

\subsection{Quantitative Evaluation}

\noindent\textbf{Experimental setup.}
We conduct experiments on three types of VAE latents (FLUX.1~\cite{flux2024}, FLUX.2~\cite{flux-2-2025}, Stable Diffusion 3 (SD3)~\cite{esser2024scaling}) and two types of vision-encoder latents (DINOv2~\cite{dinov2} and SigLIP~\cite{siglip}). We evaluate them in \textit{generation} scenarios, where the latents are sampled from a latent diffusion model rather than encoded from real images.

Specifically, we test on the FLUX.1 VAE with \textit{FLUX.1 [dev]}~\cite{flux2024} and \textit{Z-Image}~\cite{cai2025z}, the SD3 VAE with \textit{Stable Diffusion 3 Medium}~\cite{esser2024scaling}, the FLUX.2 VAE with \textit{FLUX.2 [dev]}~\cite{flux-2-2025}, DINOv2-B with \textit{DiT$^\text{DH}$}~\cite{zheng2025rae}, and SigLIP with \textit{Scale-RAE (DiT 2.8B)}~\cite{tong2026scalerae}. 
We use upsampling scale $s=8$ for SigLIP and $s=4$ for the rest. 

We decode partially denoised latents from the $M$-th denoising step out of $N$ total denoising steps with LDM early termination, denoted as PiD\texttt{(M/N)}. For DINOv2, we generate latents from ImageNet-1k classes since \textit{DiT$^\text{DH}$} is trained for class-conditional generation; for VAE latents and SigLIP experiments, we use $1{,}000$ prompts from DPG-Bench~\cite{hu2024ella} and run the latent diffusion model to obtain latents.

For baseline comparison, we use the original VAE / RAE decoder and SSDD~\cite{vallaeys2025ssdd} for latent-to-pixel decoding, and then apply state-of-the-art 1-step super-resolution models~\cite{wang2021realesrgan,yue2025arbitrary,dong2025tsd,wang2025seedvr2} to match our target output resolution ($4\times$ upsampling for VAE and DINOv2 latents, $8\times$ for SigLIP). We also include the latent-space upsampler LUA~\cite{razin2025one} in the comparison.

\noindent\textbf{Evaluation metrics.}
We assess decoded images with eight no-reference image-quality metrics that jointly capture perceptual fidelity, naturalness, and aesthetic quality without requiring ground-truth references: MUSIQ~\cite{ke2021musiq} (PaQ-2-PiQ variant), NIQE~\cite{mittal2012making}, DEQA~\cite{you2025teaching}, MANIQA~\cite{yang2022maniqa}, Q-Align~\cite{wu2023qalign}, Unipercept~\cite{cao2025unipercept} (IAA for image-aesthetic assessment and IQA for image-quality assessment), and VisualQuality-R1~\cite{wu2025visualquality}. We additionally report end-to-end decoder latency (in \textit{ms}) under both eager execution and \texttt{torch.compile} on a single GB200 GPU. We benchmark the latency using Docker \texttt{nvcr.io/nvidian/pytorch:26.02-py3}, which equips with NVIDIA CUDA 13.1.1, PyTorch 2.11.0. The latency of baseline methods include low-resolution decoding, upsampling and potential high-resolution decoding for diffusion-based super-resolution methods.

\noindent\textbf{Quantitative results.}
\Cref{tab:main-table-vae} shows that \ours achieves the best scores on most metrics across all six latent settings.
On VAE latents, \ours reduces NIQE from $4.04 / 3.76 / 3.50 / 4.05$ (best baseline) to $3.50 / 3.11 / 3.12 / 3.26$ on FLUX.1, SD3, FLUX.2, and Z-Image, and consistently leads on Unipercept-IAA, Unipercept-IQA, and VisualQuality-R1.
The gap is most pronounced on RAE-style semantic latents: on SigLIP, \ours raises MUSIQ from $73.68$ to $74.03$, DEQA from $4.00$ to $4.17$, and Unipercept-IAA from $59.95$ to $64.94$, since Scale-RAE~\cite{tong2026scalerae} struggles to generate well-structured latents and its decoder cannot correct these artifacts.
At $\sim\!210\,$ms with \texttt{torch.compile}, \ours is $3$--$6\times$ faster than diffusion-based one-step SR baselines (SeedVR2, TSD-SR, InvSR) while delivering higher quality. Lighter upsamplers like Real-ESRGAN run faster because of their lightweight network architecture, but they produce noticeably worse visual qualities.

\begin{table}[h]
\centering
\caption{\textbf{Image quality and latency across decoding pipelines.} For each cascaded baseline, we apply a state-of-the-art upsampler to match \ours's target output resolution. \ours attains the highest visual quality while running substantially faster than diffusion-based upsamplers. {\color[HTML]{CB0000} \textbf{Red}}, {\color[HTML]{3166FF} \textbf{blue}}, and {\color[HTML]{00B050} \textbf{green}} denote the best, second-best, and third-best per metric. ``QA.'' refers to Q-Align~\cite{wu2023qalign}, ``Uni.'' to Unipercept~\cite{cao2025unipercept}, and ``VQ-R1'' to VisualQuality-R1~\cite{wu2025visualquality}. Latency is reported under both eager execution and \texttt{torch.compile} on a single GB200 GPU.}
\label{tab:main-table-vae}
\resizebox{\columnwidth}{!}{%
\begin{tabular}{@{}c|l|cccccccc|cc@{}}
\toprule
 &
   &
   &
   &
   &
   &
   &
   &
   &
   &
  \multicolumn{2}{c}{Latency(ms) $\downarrow$} \\
\multirow{-2}{*}{\begin{tabular}[c]{@{}c@{}}Latent\\ (\textit{from LDM})\end{tabular}} &
  \multirow{-2}{*}{Method} &
  \multirow{-2}{*}{\begin{tabular}[c]{@{}c@{}}MUSIQ $\uparrow$\\ (paq2piq)\end{tabular}} &
  \multirow{-2}{*}{NIQE$\downarrow$} &
  \multirow{-2}{*}{DEQA $\uparrow$} &
  \multirow{-2}{*}{MANIQA$\uparrow$} &
  \multirow{-2}{*}{QA.$\uparrow$} &
  \multirow{-2}{*}{\begin{tabular}[c]{@{}c@{}}Uni.$\uparrow$ \\ (IAA)\end{tabular}} &
  \multirow{-2}{*}{\begin{tabular}[c]{@{}c@{}}Uni.$\uparrow$\\ (IQA)\end{tabular}} &
  \multirow{-2}{*}{VQ-R1$\uparrow$} &
  Eager &
  Compile \\ \midrule
 &
  LUA~\cite{razin2025one} &
  67.14 &
  7.33 &
  4.20 &
  0.36 &
  4.71 &
  59.13 &
  70.15 &
  4.66 &
  730.5 &
  369.3 \\
 &
  VAE Dec. + Real-ESRGAN~\cite{wang2021realesrgan} &
  71.65 &
  4.93 &
  4.25 &
  0.50 &
  {\color[HTML]{CB0000} \textbf{4.77}} &
  62.95 &
  73.77 &
  4.65 &
  {\color[HTML]{CB0000} \textbf{100.5}} &
  {\color[HTML]{CB0000} \textbf{62.2}} \\
 &
  VAE Dec. + SeedVR2-3B~\cite{wang2025seedvr2} &
  72.98 &
  4.05 &
  4.22 &
  0.52 &
  4.71 &
  64.02 &
  73.84 &
  4.64 &
  2084.2 &
  1237.5 \\
 &
  VAE Dec. + TSD-SR~\cite{dong2025tsd} &
  73.35 &
  4.15 &
  4.23 &
  0.53 &
  4.76 &
  62.81 &
  74.22 &
  4.67 &
  2113.2 &
  724.8 \\
 &
  VAE Dec.  + InvSR-1~\cite{yue2025arbitrary} &
  {\color[HTML]{CB0000} \textbf{73.40}} &
  4.23 &
  4.23 &
  0.56 &
  4.75 &
  64.12 &
  73.95 &
  4.65 &
  1404.1 &
  1017.7 \\
 &
  SSDD + Real-ESRGAN~\cite{wang2021realesrgan} &
  71.05 &
  5.08 &
  4.20 &
  0.49 &
  4.75 &
  62.19 &
  73.36 &
  4.63 &
  {\color[HTML]{3166FF} \textbf{143.0}} &
  {\color[HTML]{3166FF} \textbf{93.4}} \\
 &
  SSDD + SeedVR2-3B~\cite{wang2025seedvr2} &
  72.39 &
  4.23 &
  4.23 &
  0.50 &
  4.74 &
  62.97 &
  73.50 &
  4.64 &
  2126.7 &
  1268.6 \\
 &
  SSDD + TSD-SR~\cite{dong2025tsd} &
  73.06 &
  4.16 &
  4.19 &
  0.52 &
  4.75 &
  62.24 &
  73.93 &
  4.66 &
  2155.7 &
  755.9 \\
 &
  SSDD + InvSR-1~\cite{yue2025arbitrary} &
  73.07 &
  4.04 &
  4.28 &
  {\color[HTML]{CB0000} \textbf{0.55}} &
  4.76 &
  63.94 &
  73.98 &
  4.65 &
  1445.7 &
  1048.9 \\
\multirow{-10}{*}{\begin{tabular}[c]{@{}c@{}}FLUX.1 VAE \\ (\textit{FLUX.1[dev]})\end{tabular}} &
  \textbf{PiD}\texttt{(24/28)} &
  73.26 &
  {\color[HTML]{CB0000} \textbf{3.50}} &
  {\color[HTML]{CB0000} \textbf{4.31}} &
  0.54 &
  4.74 &
  {\color[HTML]{CB0000} \textbf{66.21}} &
  {\color[HTML]{CB0000} \textbf{75.21}} &
  {\color[HTML]{CB0000} \textbf{4.68}} &
  {\color[HTML]{009901} \textbf{512.7}} &
  {\color[HTML]{009901} \textbf{211.2}} \\ \midrule
 &
  VAE Dec. + Real-ESRGAN~\cite{wang2021realesrgan} &
  72.22 &
  4.59 &
  3.97 &
  0.49 &
  4.50 &
  55.72 &
  70.65 &
  4.39 &
  {\color[HTML]{CB0000} \textbf{100.7}} &
  {\color[HTML]{CB0000} \textbf{62.2}} \\
 &
  VAE Dec. + SeedVR2-3B~\cite{wang2025seedvr2} &
  73.55 &
  3.87 &
  3.91 &
  0.52 &
  4.39 &
  56.26 &
  70.14 &
  4.37 &
  2084.4 &
  1237.5 \\
 &
  VAE Dec. + TSD-SR~\cite{dong2025tsd} &
  73.93 &
  3.76 &
  3.98 &
  0.53 &
  4.52 &
  56.55 &
  71.16 &
  4.43 &
  2113.4 &
  724.8 \\
 &
  VAE Dec.  + InvSR-1~\cite{yue2025arbitrary} &
  {\color[HTML]{CB0000} \textbf{74.11}} &
  3.82 &
  4.05 &
  0.56 &
  4.56 &
  57.09 &
  71.40 &
  4.41 &
  1404.3 &
  1017.7 \\
 &
  SSDD + Real-ESRGAN~\cite{wang2021realesrgan} &
  71.58 &
  4.90 &
  3.88 &
  0.48 &
  4.45 &
  53.99 &
  69.18 &
  4.35 &
  {\color[HTML]{3166FF} \textbf{143.0}} &
  {\color[HTML]{3166FF} \textbf{93.4}} \\
 &
  SSDD + SeedVR2-3B~\cite{wang2025seedvr2} &
  72.10 &
  4.59 &
  3.86 &
  0.47 &
  4.36 &
  54.11 &
  68.10 &
  4.33 &
  2126.7 &
  1268.6 \\
 &
  SSDD + TSD-SR~\cite{dong2025tsd} &
  73.56 &
  3.89 &
  3.90 &
  0.52 &
  4.46 &
  55.30 &
  69.91 &
  4.39 &
  2155.7 &
  755.9 \\
 &
  SSDD + InvSR-1~\cite{yue2025arbitrary} &
  73.83 &
  4.22 &
  4.09 &
  0.56 &
  4.56 &
  55.47 &
  70.16 &
  4.38 &
  1445.7 &
  1048.9 \\
\multirow{-9}{*}{\begin{tabular}[c]{@{}c@{}}SD3 VAE \\ (\textit{SD3-medium})\end{tabular}} &
  \textbf{PiD} \texttt{(24/28)} &
  74.00 &
  {\color[HTML]{CB0000} \textbf{3.11}} &
  {\color[HTML]{CB0000} \textbf{4.26}} &
  {\color[HTML]{CB0000} \textbf{0.56}} &
  {\color[HTML]{CB0000} \textbf{4.66}} &
  {\color[HTML]{CB0000} \textbf{62.57}} &
  {\color[HTML]{CB0000} \textbf{74.22}} &
  {\color[HTML]{CB0000} \textbf{4.59}} &
  {\color[HTML]{009901} \textbf{501.4}} &
  {\color[HTML]{009901} \textbf{214.0}} \\ \midrule
 &
  VAE Dec. + Real-ESRGAN~\cite{wang2021realesrgan} &
  72.19 &
  4.44 &
  4.19 &
  0.52 &
  4.69 &
  61.96 &
  74.77 &
  4.58 &
  {\color[HTML]{CB0000} \textbf{100.8}} &
  {\color[HTML]{CB0000} \textbf{62.2}} \\
 &
  VAE Dec. + SeedVR2-3B~\cite{wang2025seedvr2} &
  73.41 &
  3.55 &
  4.18 &
  0.54 &
  4.60 &
  62.85 &
  74.72 &
  4.58 &
  2084.5 &
  1237.4 \\
 &
  VAE Dec. + TSD-SR~\cite{dong2025tsd} &
  74.07 &
  3.59 &
  4.21 &
  0.56 &
  4.69 &
  61.72 &
  75.28 &
  4.62 &
  2113.5 &
  724.7 \\
 &
  VAE Dec.  + InvSR-1~\cite{yue2025arbitrary} &
  {\color[HTML]{CB0000} \textbf{74.13}} &
  3.65 &
  4.25 &
  {\color[HTML]{CB0000} \textbf{0.58}} &
  4.72 &
  63.52 &
  75.47 &
  4.60 &
  1404.4 &
  1017.7 \\
 &
  SSDD + Real-ESRGAN~\cite{wang2021realesrgan} &
  68.95 &
  4.48 &
  3.14 &
  0.48 &
  3.46 &
  43.94 &
  49.30 &
  3.47 &
  {\color[HTML]{3166FF} \textbf{143.0}} &
  {\color[HTML]{3166FF} \textbf{93.4}} \\
 &
  SSDD + SeedVR2-3B~\cite{wang2025seedvr2} &
  64.23 &
  3.55 &
  3.05 &
  0.49 &
  3.21 &
  42.91 &
  45.54 &
  3.00 &
  2126.7 &
  1268.6 \\
 &
  SSDD + TSD-SR~\cite{dong2025tsd} &
  73.76 &
  3.50 &
  3.72 &
  0.55 &
  4.07 &
  54.21 &
  62.25 &
  3.98 &
  2155.7 &
  755.9 \\
 &
  SSDD + InvSR-1~\cite{yue2025arbitrary} &
  73.32 &
  3.80 &
  3.76 &
  0.56 &
  4.12 &
  50.39 &
  60.25 &
  3.96 &
  1445.7 &
  1048.9 \\
\multirow{-9}{*}{\begin{tabular}[c]{@{}c@{}}FLUX.2 VAE\\ (\textit{FLUX.2[dev]})\end{tabular}} &
  \textbf{PiD} \texttt{(45/50)} &
  73.79 &
  {\color[HTML]{CB0000} \textbf{3.12}} &
  {\color[HTML]{CB0000} \textbf{4.30}} &
  0.56 &
  {\color[HTML]{CB0000} \textbf{4.70}} &
  {\color[HTML]{CB0000} \textbf{66.01}} &
  {\color[HTML]{CB0000} \textbf{75.71}} &
  {\color[HTML]{CB0000} \textbf{4.66}} &
  {\color[HTML]{009901} \textbf{508.3}} &
  {\color[HTML]{009901} \textbf{206.1}} \\ \midrule
 &
  LUA~\cite{razin2025one} &
  66.92 &
  7.39 &
  4.04 &
  0.36 &
  4.49 &
  54.70 &
  68.49 &
  4.52 &
  730.5 &
  369.3 \\
 &
  VAE Dec. + Real-ESRGAN~\cite{wang2021realesrgan} &
  71.56 &
  4.99 &
  4.09 &
  0.50 &
  4.60 &
  56.92 &
  71.87 &
  4.48 &
  {\color[HTML]{CB0000} \textbf{100.5}} &
  {\color[HTML]{CB0000} \textbf{62.2}} \\
 &
  VAE Dec. + SeedVR2-3B~\cite{wang2025seedvr2} &
  73.01 &
  4.08 &
  4.05 &
  0.51 &
  4.51 &
  57.97 &
  71.97 &
  4.50 &
  2084.2 &
  1237.5 \\
 &
  VAE Dec. + TSD-SR~\cite{dong2025tsd} &
  73.67 &
  4.09 &
  4.10 &
  0.53 &
  4.63 &
  57.69 &
  72.90 &
  4.54 &
  2113.2 &
  724.8 \\
 &
  VAE Dec.  + InvSR-1~\cite{yue2025arbitrary} &
  73.85 &
  4.23 &
  4.16 &
  0.56 &
  4.64 &
  59.21 &
  72.89 &
  4.51 &
  1404.1 &
  1017.7 \\
 &
  SSDD + Real-ESRGAN~\cite{wang2021realesrgan} &
  70.96 &
  5.11 &
  4.02 &
  0.49 &
  4.56 &
  56.29 &
  71.42 &
  4.47 &
  {\color[HTML]{3166FF} \textbf{143.0}} &
  {\color[HTML]{3166FF} \textbf{93.4}} \\
 &
  SSDD + SeedVR2-3B~\cite{wang2025seedvr2} &
  72.14 &
  4.36 &
  4.04 &
  0.48 &
  4.51 &
  57.37 &
  71.47 &
  4.50 &
  2126.7 &
  1268.6 \\
 &
  SSDD + TSD-SR~\cite{dong2025tsd} &
  73.36 &
  4.11 &
  4.04 &
  0.52 &
  4.60 &
  57.26 &
  72.43 &
  4.53 &
  2155.7 &
  755.9 \\
 &
  SSDD + InvSR-1~\cite{yue2025arbitrary} &
  73.54 &
  4.05 &
  4.20 &
  0.56 &
  4.64 &
  58.87 &
  72.87 &
  4.52 &
  1445.7 &
  1048.9 \\
\multirow{-10}{*}{\begin{tabular}[c]{@{}c@{}}FLUX.1 VAE \\ (\textit{Z-Image})\end{tabular}} &
  \textbf{PiD} \texttt{(45/50)} &
  {\color[HTML]{CB0000} \textbf{74.08}} &
  {\color[HTML]{CB0000} \textbf{3.26}} &
  {\color[HTML]{CB0000} \textbf{4.29}} &
  {\color[HTML]{CB0000} \textbf{0.56}} &
  {\color[HTML]{CB0000} \textbf{4.68}} &
  {\color[HTML]{CB0000} \textbf{63.96}} &
  {\color[HTML]{CB0000} \textbf{75.23}} &
  {\color[HTML]{CB0000} \textbf{4.64}} &
  {\color[HTML]{009901} \textbf{498.3}} &
  {\color[HTML]{009901} \textbf{211.2}} \\ \midrule
 &
  RAE Dec. + RealESRGAN &
  68.10 &
  4.36 &
  3.67 &
  0.44 &
  3.81 &
  50.52 &
  60.18 &
  4.17 &
  {\color[HTML]{CB0000} \textbf{94.0}} &
  {\color[HTML]{CB0000} \textbf{67.0}} \\
 &
  RAE Dec. + SeedVR2-3B~\cite{wang2025seedvr2} &
  69.78 &
  3.60 &
  3.89 &
  0.49 &
  4.01 &
  55.04 &
  64.46 &
  4.27 &
  2077.7 &
  1242.2 \\
 &
  RAE Dec. + TSD-SR~\cite{dong2025tsd} &
  {\color[HTML]{CB0000} \textbf{73.65}} &
  {\color[HTML]{CB0000} \textbf{3.22}} &
  4.06 &
  0.53 &
  4.23 &
  58.50 &
  70.16 &
  4.38 &
  2106.7 &
  729.5 \\
 &
  RAE Dec. + InvSR-1~\cite{yue2025arbitrary} &
  71.85 &
  3.44 &
  3.95 &
  0.53 &
  4.02 &
  52.94 &
  64.32 &
  4.19 &
  1396.7 &
  1022.5 \\
\multirow{-5}{*}{\begin{tabular}[c]{@{}c@{}}DINOv2\\ (\textit{DiT$^{\text{DH}}$})\end{tabular}} &
  \textbf{PiD} \texttt{(50/50)}&
  73.31 &
  3.38 &
  {\color[HTML]{CB0000} \textbf{4.27}} &
  {\color[HTML]{CB0000} \textbf{0.54}} &
  {\color[HTML]{CB0000} \textbf{4.55}} &
  {\color[HTML]{CB0000} \textbf{69.81}} &
  {\color[HTML]{CB0000} \textbf{76.52}} &
  {\color[HTML]{CB0000} \textbf{4.63}} &
  {\color[HTML]{3166FF} \textbf{499.5}} &
  {\color[HTML]{3166FF} \textbf{212.4}} \\ \midrule
 &
  RAE Dec. + Real-ESRGAN~\cite{wang2021realesrgan} &
  64.07 &
  4.58 &
  3.57 &
  0.43 &
  3.79 &
  52.05 &
  58.54 &
  4.14 &
  {\color[HTML]{CB0000} \textbf{94.8}} &
  {\color[HTML]{CB0000} \textbf{67.4}} \\
 &
  RAE Dec. + SeedVR2-3B~\cite{wang2025seedvr2} &
  68.57 &
  4.00 &
  3.65 &
  0.49 &
  3.88 &
  51.71 &
  57.96 &
  4.13 &
  2078.5 &
  1242.6 \\
 &
  RAE Dec. + TSD-SR~\cite{dong2025tsd} &
  73.68 &
  3.82 &
  4.00 &
  0.50 &
  4.42 &
  59.95 &
  68.03 &
  4.40 &
  2107.5 &
  730.0 \\
 &
  RAE Dec. + InvSR-1~\cite{yue2025arbitrary} &
  71.47 &
  {\color[HTML]{CB0000} \textbf{3.30}} &
  3.93 &
  0.54 &
  4.03 &
  56.43 &
  63.35 &
  4.31 &
  1397.5 &
  1022.9 \\
\multirow{-5}{*}{\begin{tabular}[c]{@{}c@{}}SigLIP\\ (\textit{Scale-RAE})\end{tabular}} &
  \textbf{PiD}\texttt{(50/50)} &
  {\color[HTML]{CB0000} \textbf{74.03}} &
  3.34 &
  {\color[HTML]{CB0000} \textbf{4.17}} &
  {\color[HTML]{CB0000} \textbf{0.56}} &
  {\color[HTML]{CB0000} \textbf{4.43}} &
  {\color[HTML]{CB0000} \textbf{64.94}} &
  {\color[HTML]{CB0000} \textbf{72.78}} &
  {\color[HTML]{CB0000} \textbf{4.45}} &
  {\color[HTML]{3166FF} \textbf{501.5}} &
  {\color[HTML]{3166FF} \textbf{208.7}} \\ \bottomrule
\end{tabular}%
}
\vspace{-1.5em}
\end{table}
\FloatBarrier

\begin{figure}[th]
  \centering
  \includegraphics[width=\linewidth]{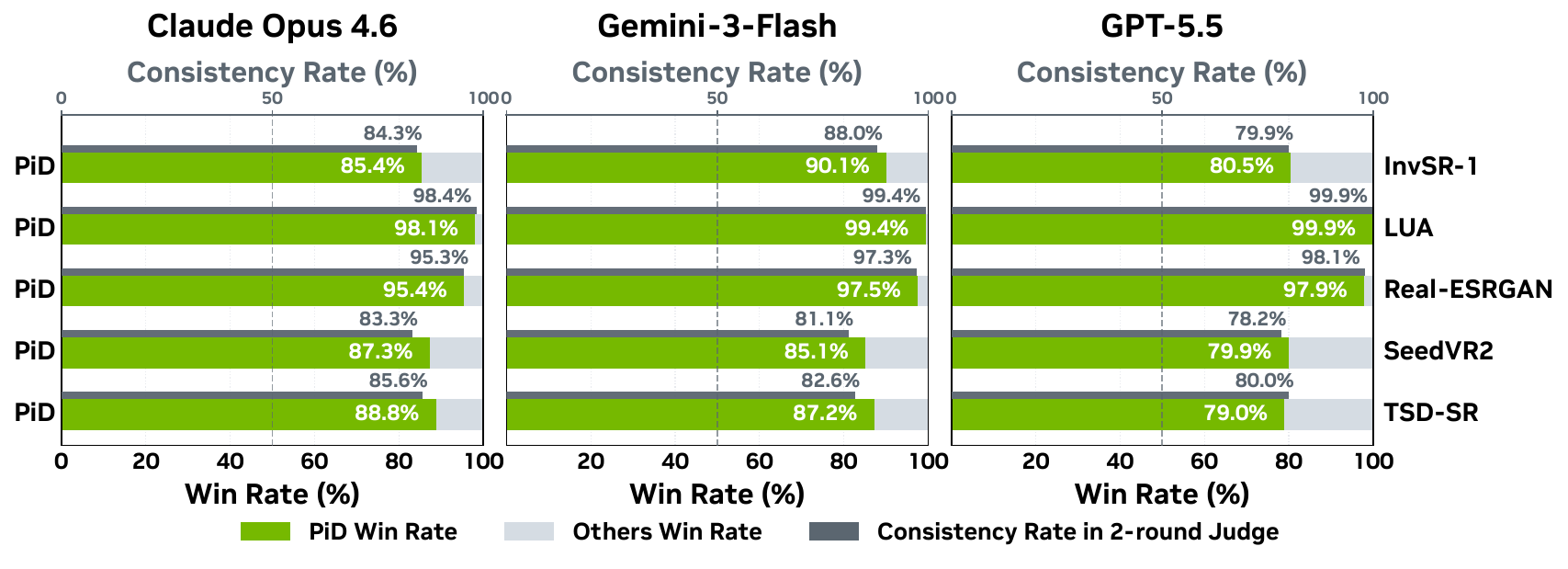}
  \vspace{-2em}
  \caption{\textbf{Pairwise image quality preference judged by closed-source MLLMs.} Three MLLMs compare \ours-decoded images against cascaded baselines (original VAE decoder followed by SR). All three MLLMs consistently prefer \ours, with high 2-round consistency under image order swap.}
  \label{fig:mllm-judge-winrate-pid-consistency}
  \vspace{-0.5em}
\end{figure}
\FloatBarrier

\begin{table}[h]
\centering
\caption{Performance of different inference steps of teacher and student model. For teacher models, more inference step leads to better image quality. However, few-step student model can surpass multiple-step teacher model in generated latent decoding cases.}
\label{tab:step-ablation}
\resizebox{\columnwidth}{!}{%
\begin{tabular}{@{}c|c|cccccccc|ccc@{}}
\toprule
\multirow{2}{*}{Model} &
  \multirow{2}{*}{\begin{tabular}[c]{@{}c@{}}Num of \\ Inference Step\end{tabular}} &
  \multicolumn{8}{c|}{\textbf{PiD}\texttt{(24/28)} (\textit{FLUX.1 [dev]})} &
  \multicolumn{3}{c}{\textbf{Small Text Reconstruction}} \\ \cmidrule(l){3-13} 
 &
   &
  MUSIQ $\uparrow$ &
  NIQE$\downarrow$ &
  DEQA$\uparrow$ &
  MANIQA$\uparrow$ &
  QA.$\uparrow$ &
  Uni. (IAA)$\uparrow$ &
  Uni.IQA $\uparrow$ &
  VQ-R1$\uparrow$ &
  PSNR$\uparrow$ &
  SSIM$\uparrow$ &
  LPIPS$\downarrow$ \\ \midrule
Teacher & 50 & 71.79 & 4.92 & 4.28 & 0.49 & 4.74          & 63.82 & 73.35 & 4.64 & 24.96          & \textbf{0.966} & 0.16 \\
Teacher & 25 & 71.63 & 5.43 & 4.29 & 0.49 & 4.75          & 63.36 & 73.26 & 4.65 & 25.00          & 0.965          & 0.18 \\
Teacher & 12 & 70.95 & 6.02 & 4.29 & 0.48 & \textbf{4.76} & 62.68 & 72.90 & 4.64 & 25.12          & \textbf{0.966} & 0.18 \\
Teacher & 8  & 70.32 & 6.31 & 4.29 & 0.47 & 4.75          & 62.15 & 72.51 & 4.64 & 25.24          & 0.964          & 0.19 \\
Teacher & 4  & 68.32 & 7.00 & 4.24 & 0.45 & 4.72          & 60.50 & 71.13 & 4.63 & \textbf{25.70} & 0.960          & 0.21 \\
Student &
  4 &
  \textbf{73.26} &
  \textbf{3.50} &
  \textbf{4.31} &
  \textbf{0.54} &
  4.74 &
  \textbf{66.21} &
  \textbf{75.21} &
  \textbf{4.68} &
  24.19 &
  0.964 &
  \textbf{0.09} \\ \bottomrule
\end{tabular}%
}
\end{table}

\noindent\textbf{Closed-source multimodal LLM judgment.}
We adopt three of the most powerful closed-source multimodal LLMs (MLLMs) for pairwise comparison: Gemini 3 Flash, GPT 5.5, and Claude Opus 4.6. Given image A and image B, the MLLM is asked to judge which one is better in quality and detail. To avoid order bias, we use a two-round evaluation by switching the input order of A and B, and report agreement across the two rounds to measure the model's confidence. We call it consistency rate.
A higher consistency rate indicates that the model is more confident in its assessment and suggests a greater difference in image quality.

As shown in~\Cref{fig:mllm-judge-winrate-pid-consistency}, all three MLLMs consistently judge \ours to outperform the baseline methods, with high \ours win rate and high 2-round consistency.

\noindent\textbf{Decoding Step Analysis}
We evaluate the teacher model with \{50, 25, 12, 8, 4\} inference steps, and compare it with our distilled 4-step student using the latent generated by FLUX.1 [dev]. As expected, the teacher exhibits a quality--cost trade-off: reducing steps gradually degrades perceptual metrics (e.g., MUSIQ decreases and NIQE increases), with the 4-step teacher being the weakest. But the 4-step student not only closes the gap but surpasses all multi-step teacher variants on IQA metrics (best MUSIQ/NIQE/DEQA/MANIQA and the highest uniformity/IQA scores in Table~\ref{tab:step-ablation}), indicating that distillation successfully transfers multi-step behavior into a few-step decoder.

For small-text reconstruction, the picture is more nuanced: while the multi-step teacher variants achieve higher PSNR/SSIM, the student substantially improves perceptual similarity (lowest LPIPS). This suggests that the student prioritizes visually plausible character strokes and local textures, but the pixel-wise alignment is not maximized.

\begin{figure*}[t]
  \centering
  \setlength{\tabcolsep}{0pt}
  \renewcommand{\arraystretch}{0}
  \begin{tabular}{@{}cccccc@{}}
    \includegraphics[width=0.1667\textwidth]{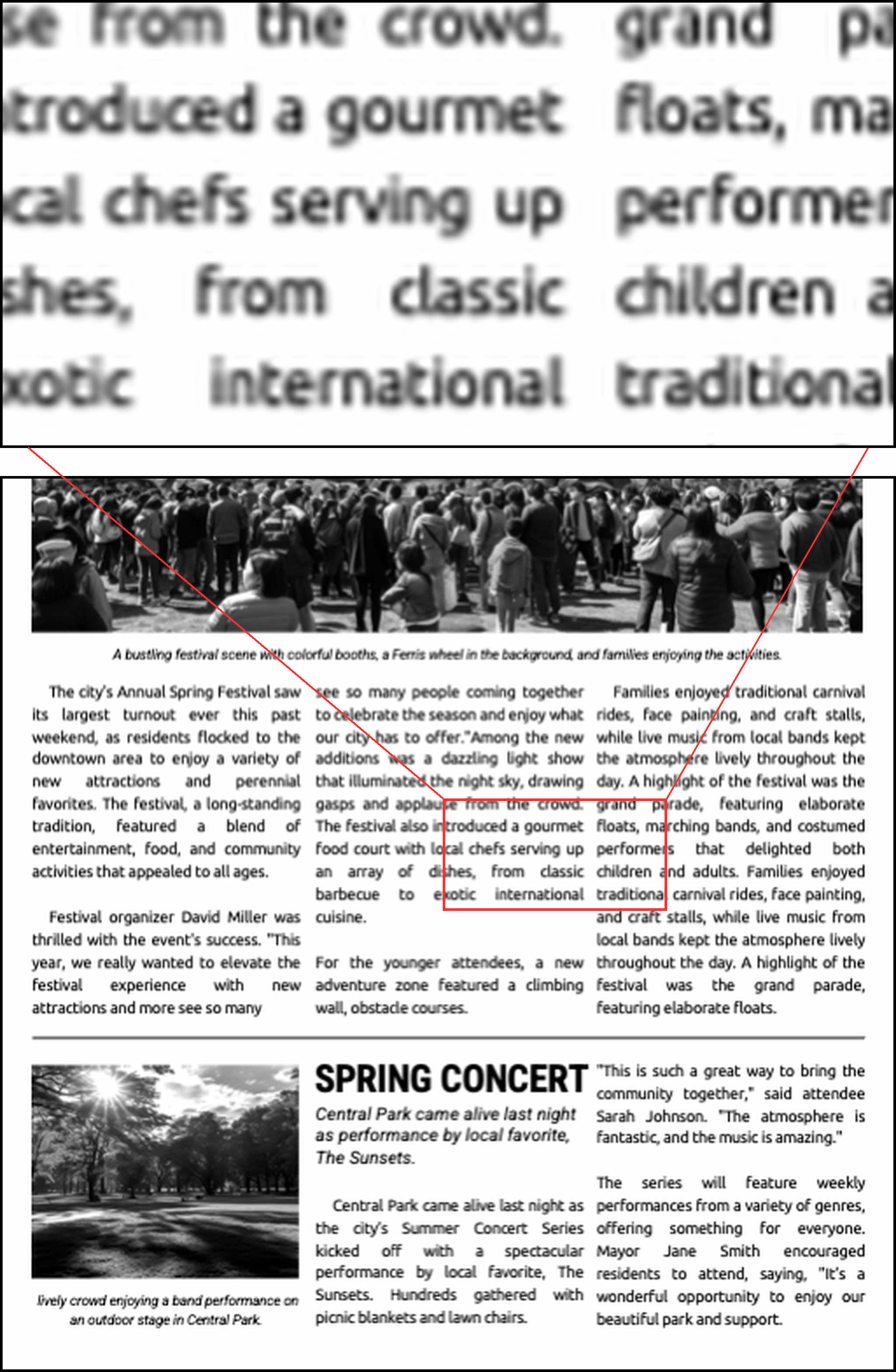} &
    \includegraphics[width=0.1667\textwidth]{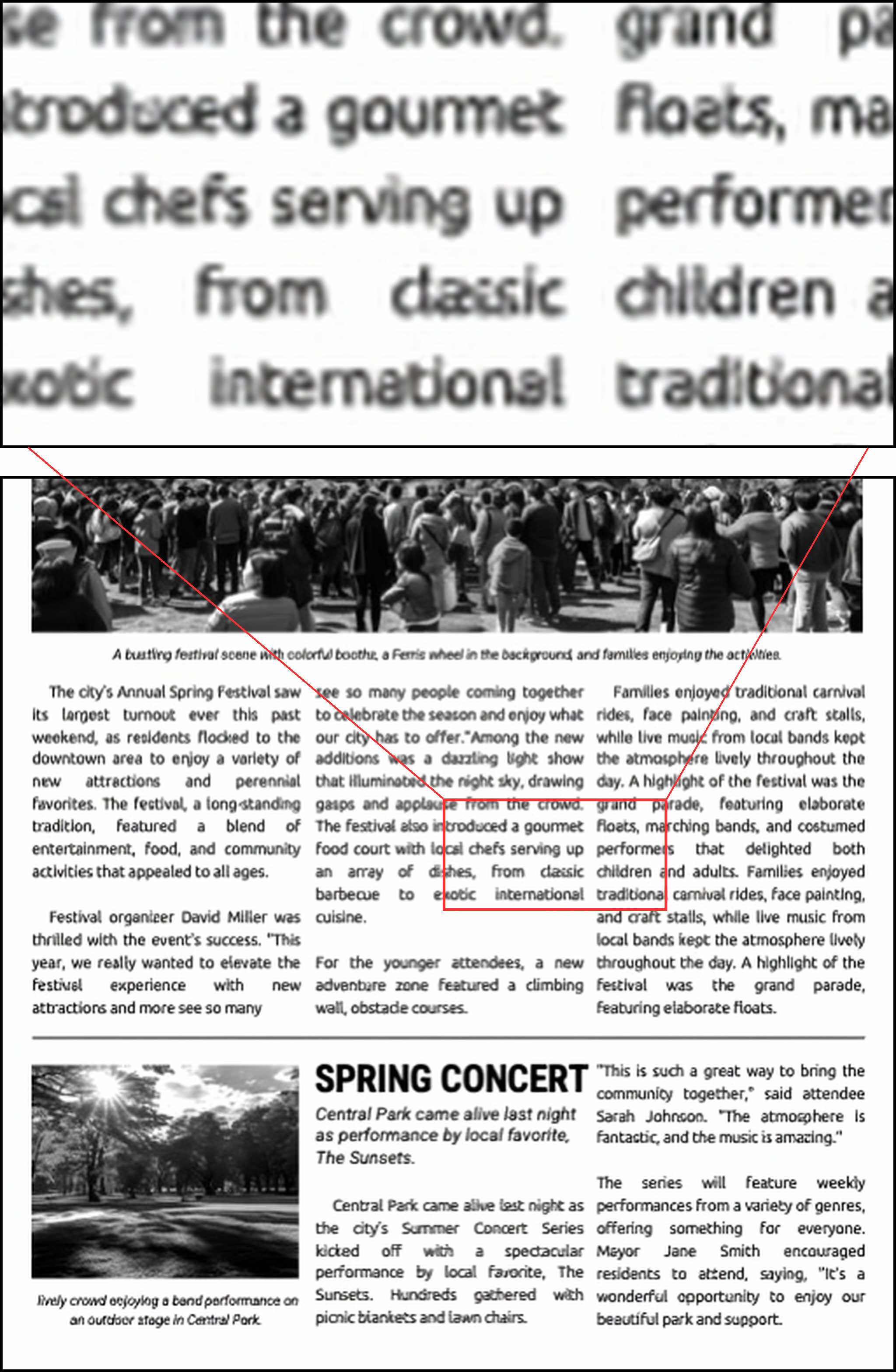} &
    \includegraphics[width=0.1667\textwidth]{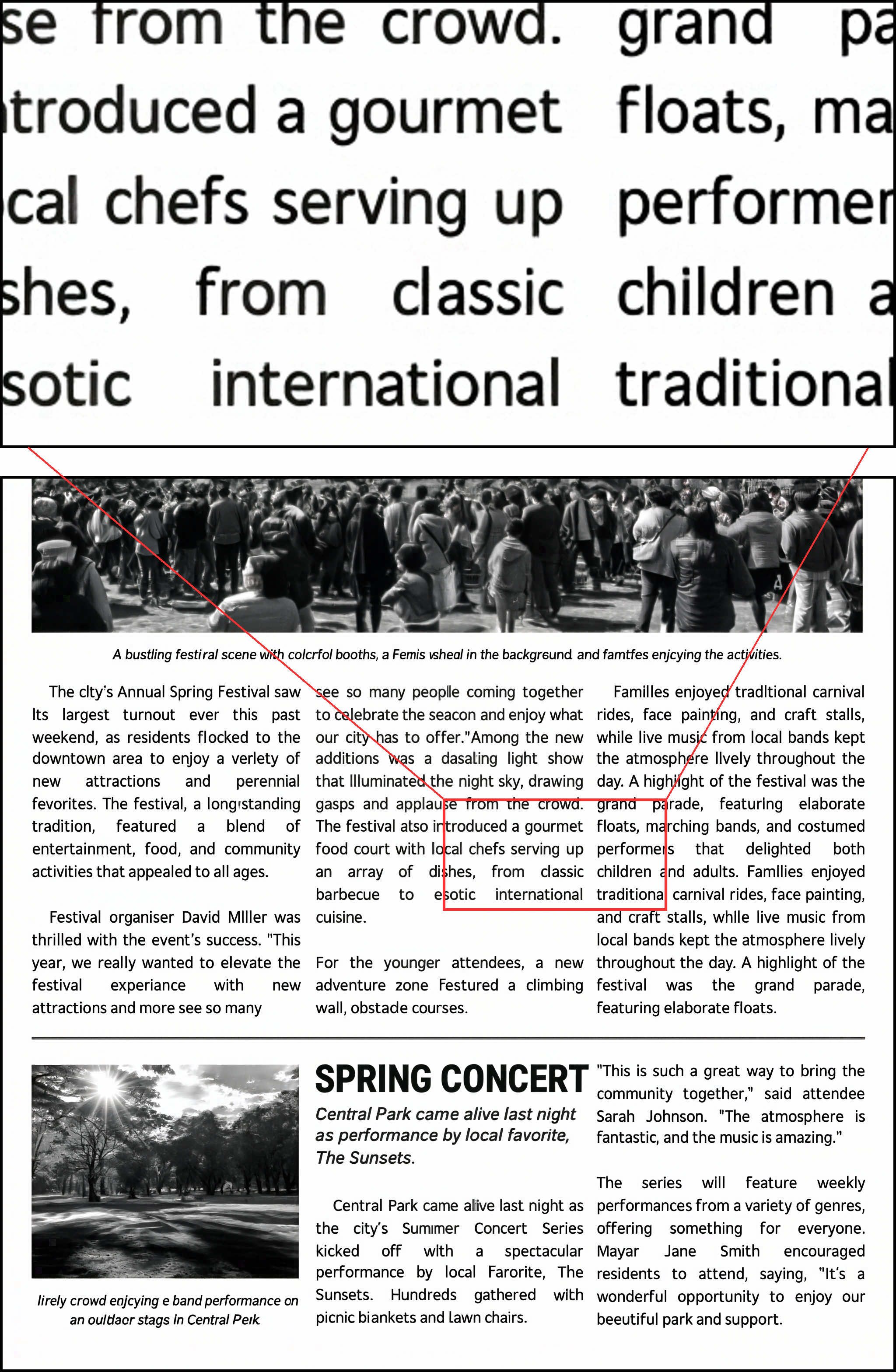} &
    \includegraphics[width=0.1667\textwidth]{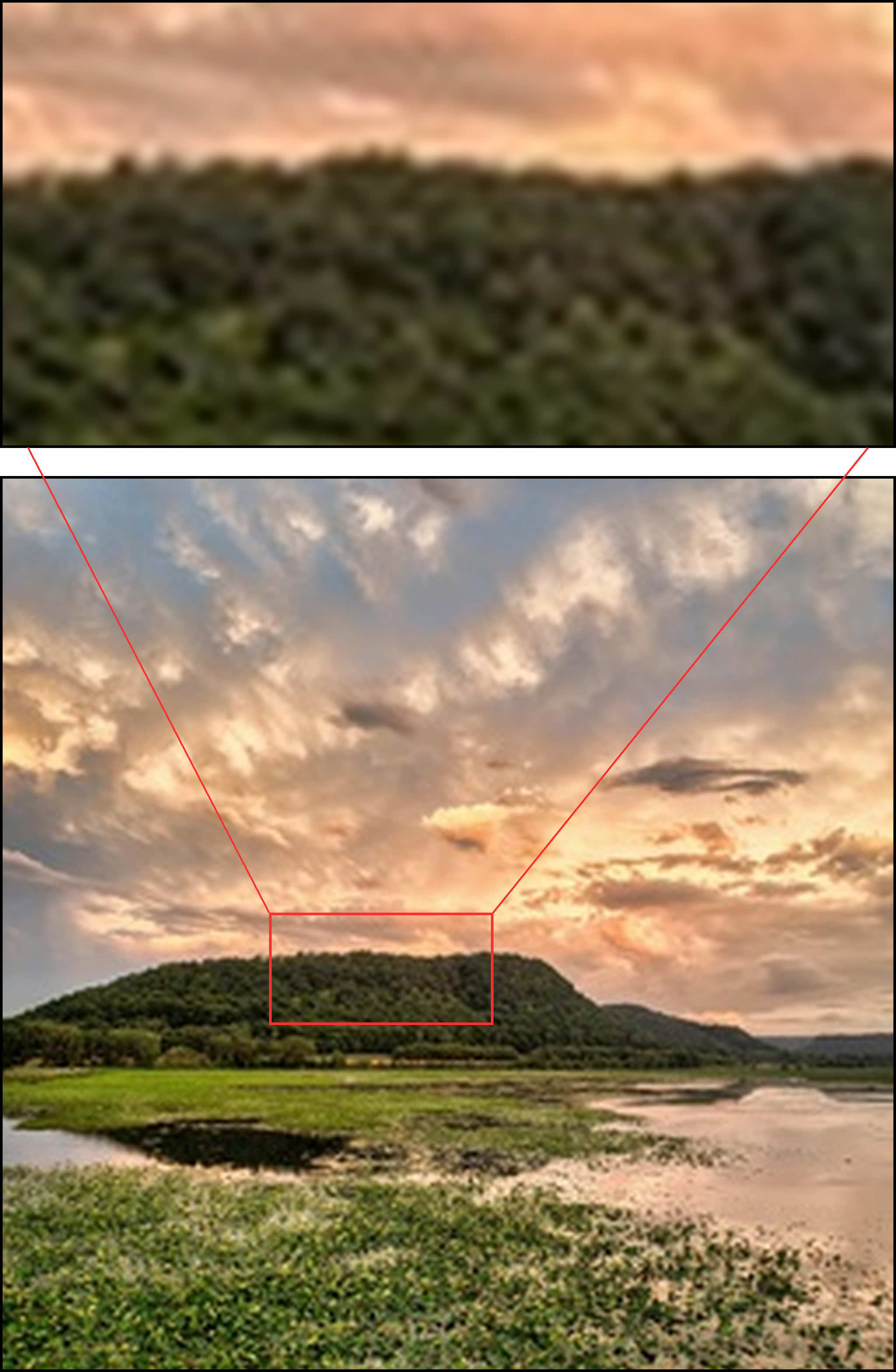} &
    \includegraphics[width=0.1667\textwidth]{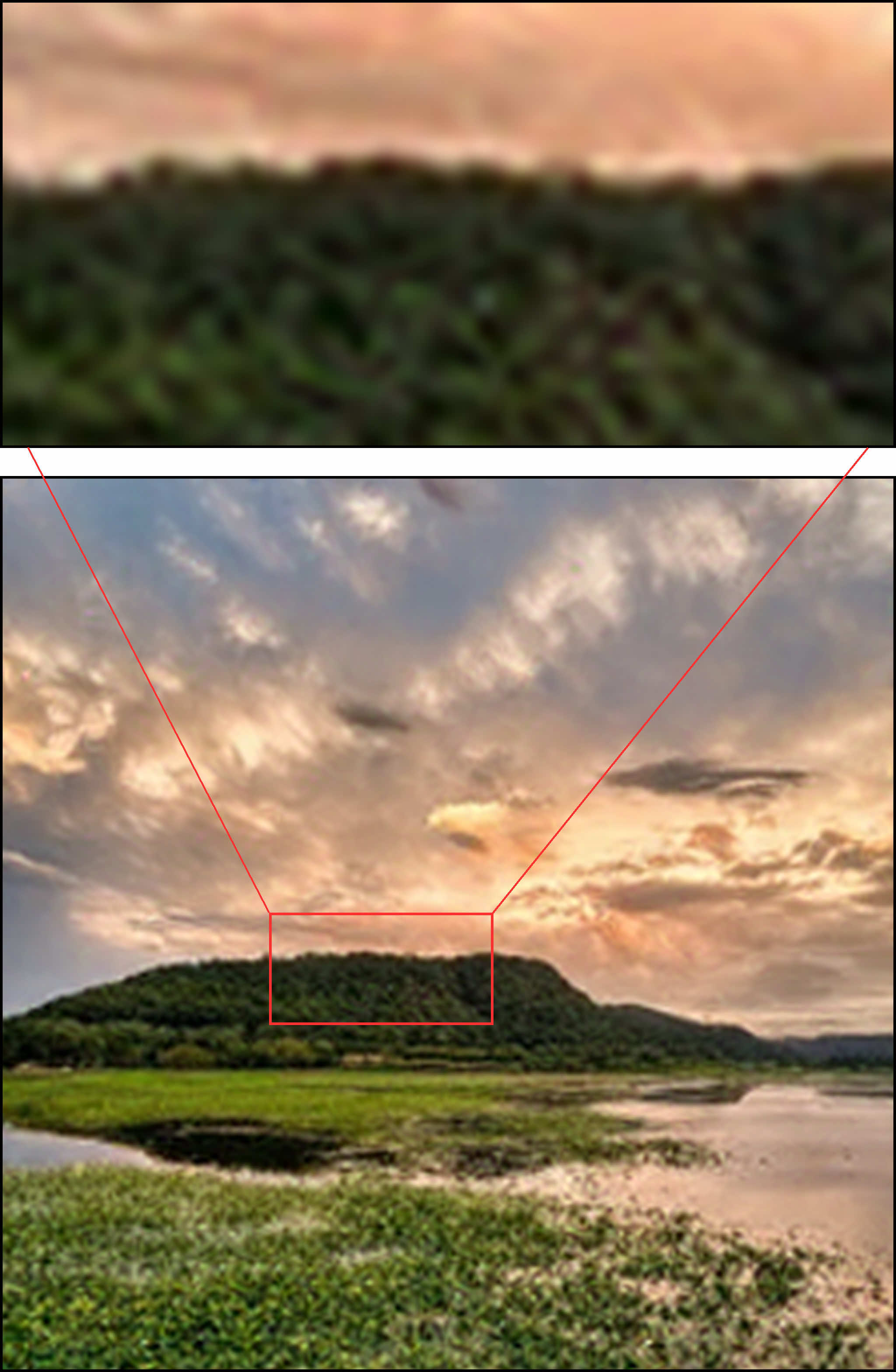} &
    \includegraphics[width=0.1667\textwidth]{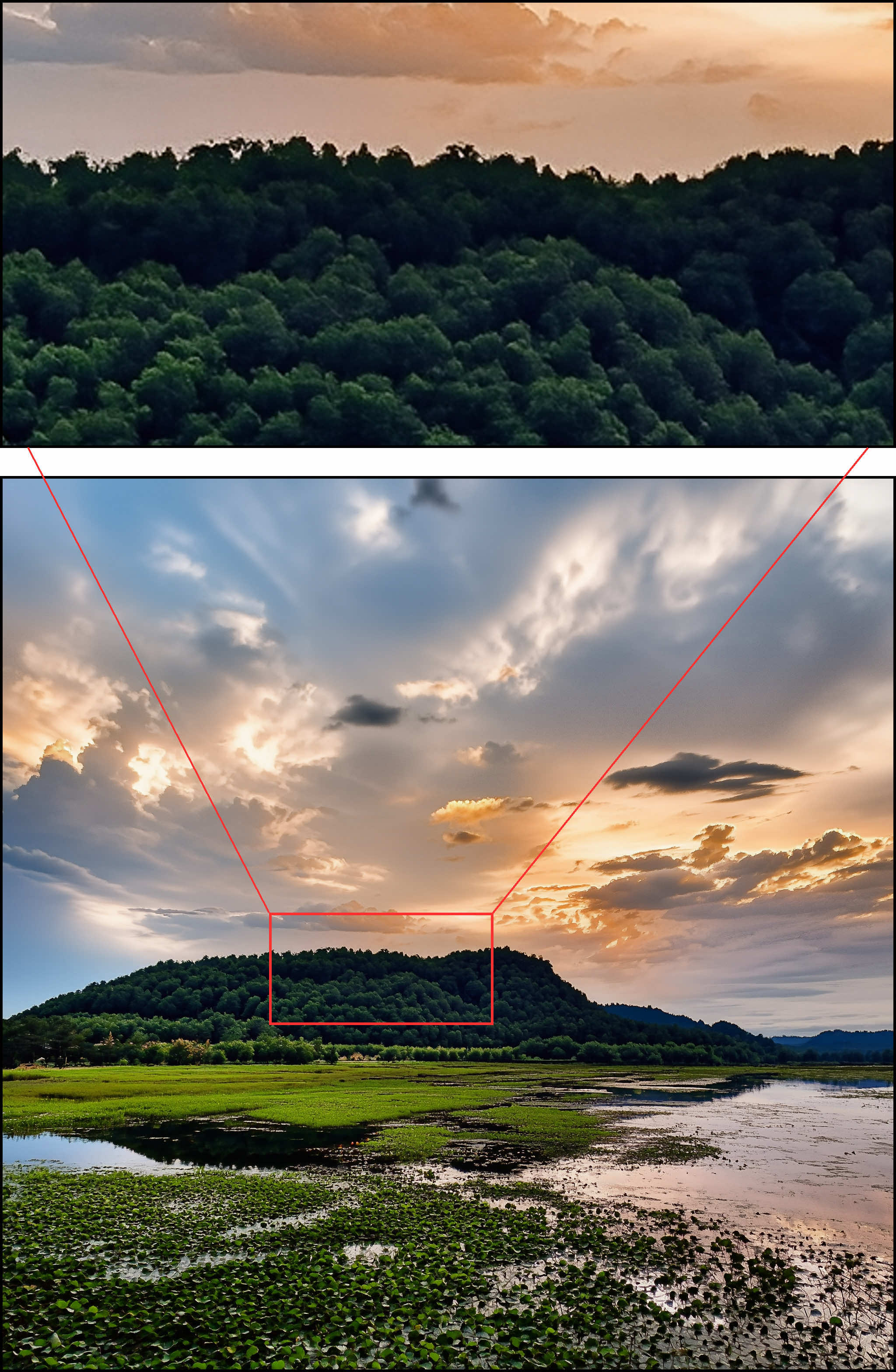}\\
    \\[0.2em]
    \multicolumn{1}{c}{\scriptsize Input $(512^2)$} &
    \multicolumn{1}{c}{\scriptsize FLUX.1 VAE $(512^2)$} &
    \multicolumn{1}{c}{\scriptsize \ours $(2048^2)$} &
    \multicolumn{1}{c}{\scriptsize Input $(512^2)$} &
    \multicolumn{1}{c}{\scriptsize Scale-RAE $(512^2)$} &
    \multicolumn{1}{c}{\scriptsize \ours $(2048^2)$}\\
  \end{tabular}
  \caption{\textbf{Image reconstruction comparison.} Given a latent encoded from a clean image, \ours reconstructs the image at higher resolution with sharper details than the original VAE / RAE decoder.}
  \label{fig:recon-vis}
\end{figure*}

\begin{figure*}[h]
  \centering
  \setlength{\tabcolsep}{0pt}
  \renewcommand{\arraystretch}{1.0}
  \begin{tabular}{@{}ccccc@{}}
    \imgwithlabel{0.2\textwidth}{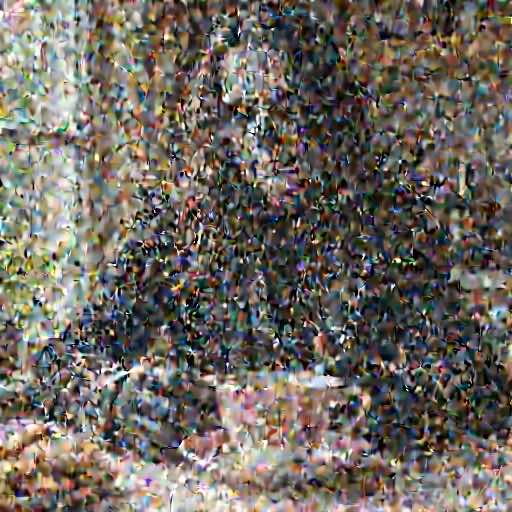}{VAE decoding} &
    \includegraphics[width=0.2\textwidth]{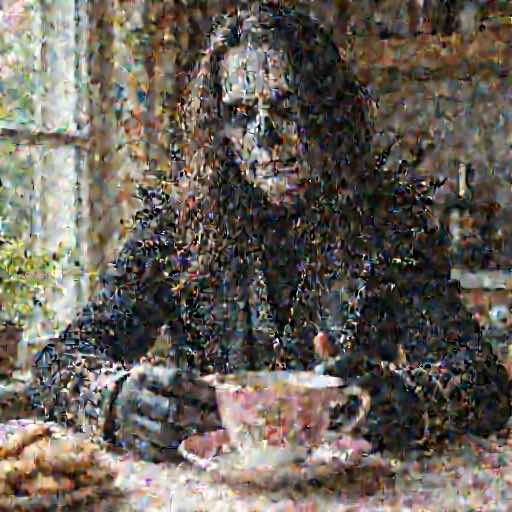} &
    \includegraphics[width=0.2\textwidth]{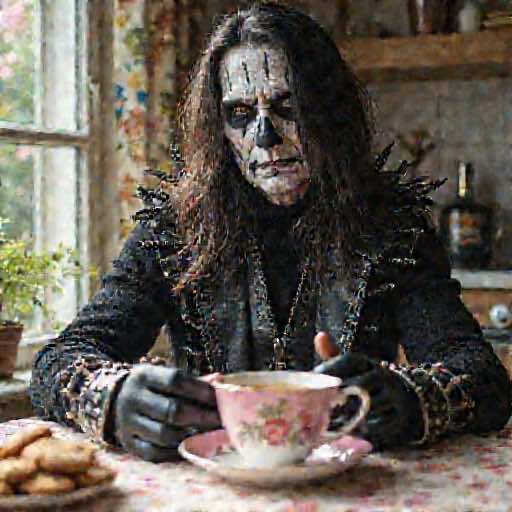} &
    \includegraphics[width=0.2\textwidth]{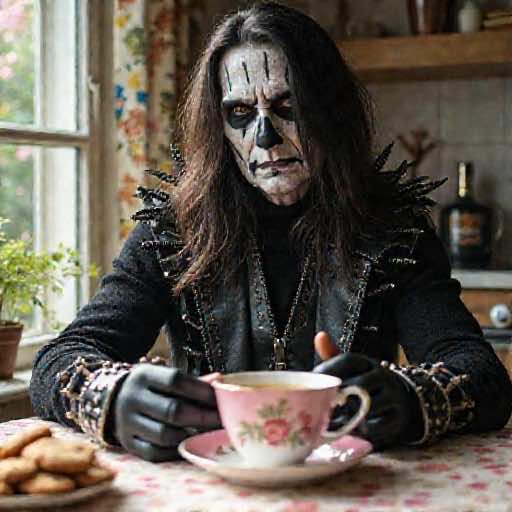} &
    \includegraphics[width=0.2\textwidth]{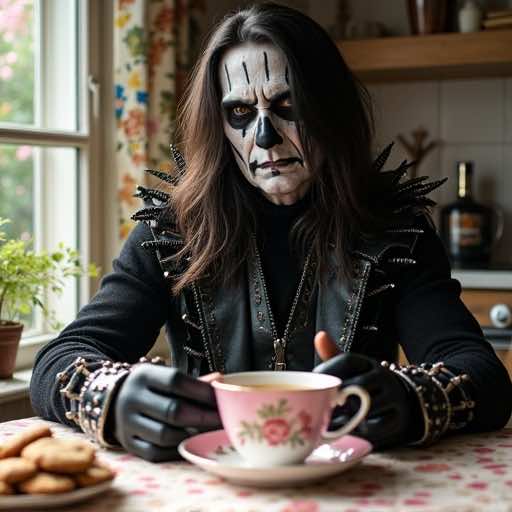} \\[-2pt]
    \imgwithlabel{0.2\textwidth}{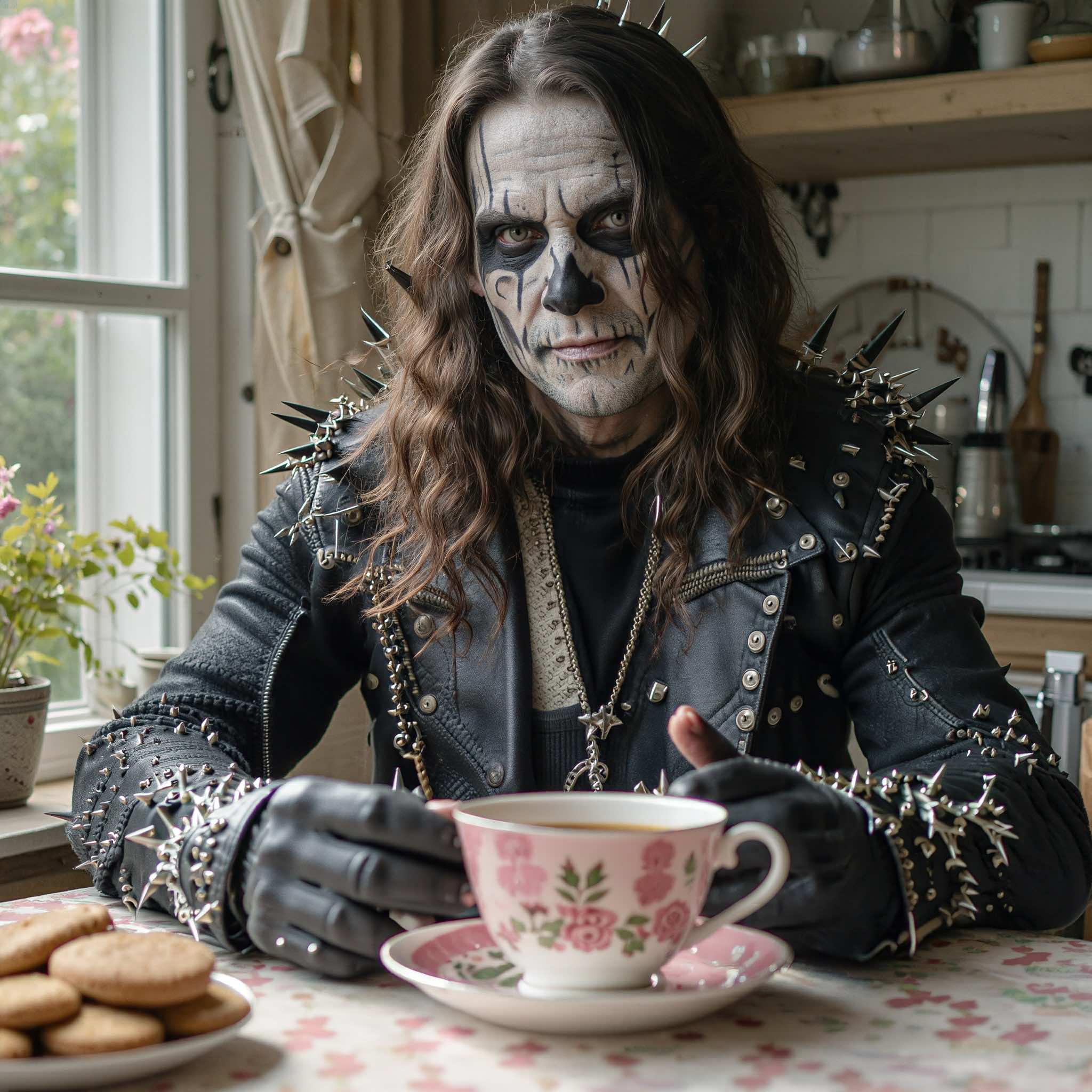}{PiD decoding} &
    \includegraphics[width=0.2\textwidth]{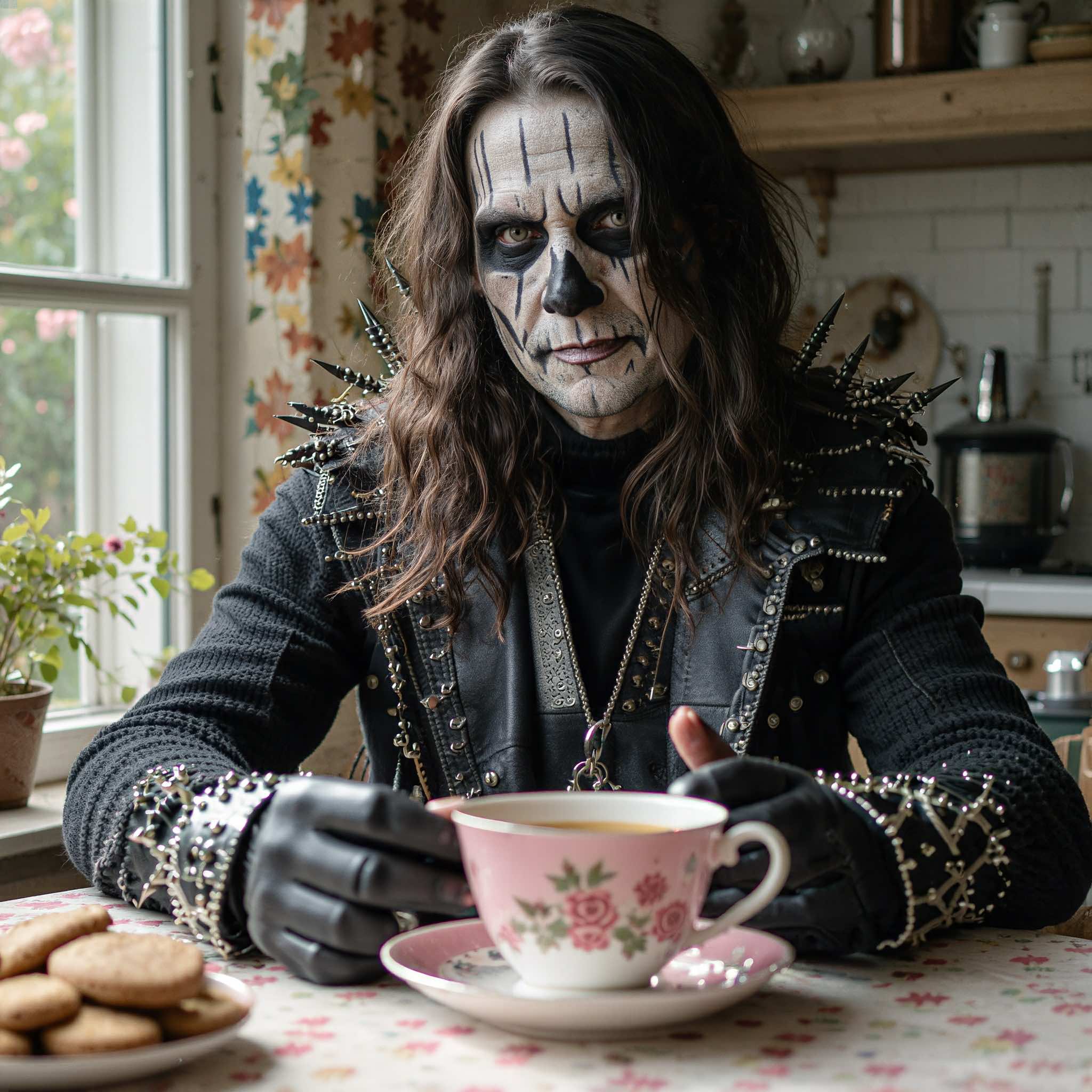} &
    \includegraphics[width=0.2\textwidth]{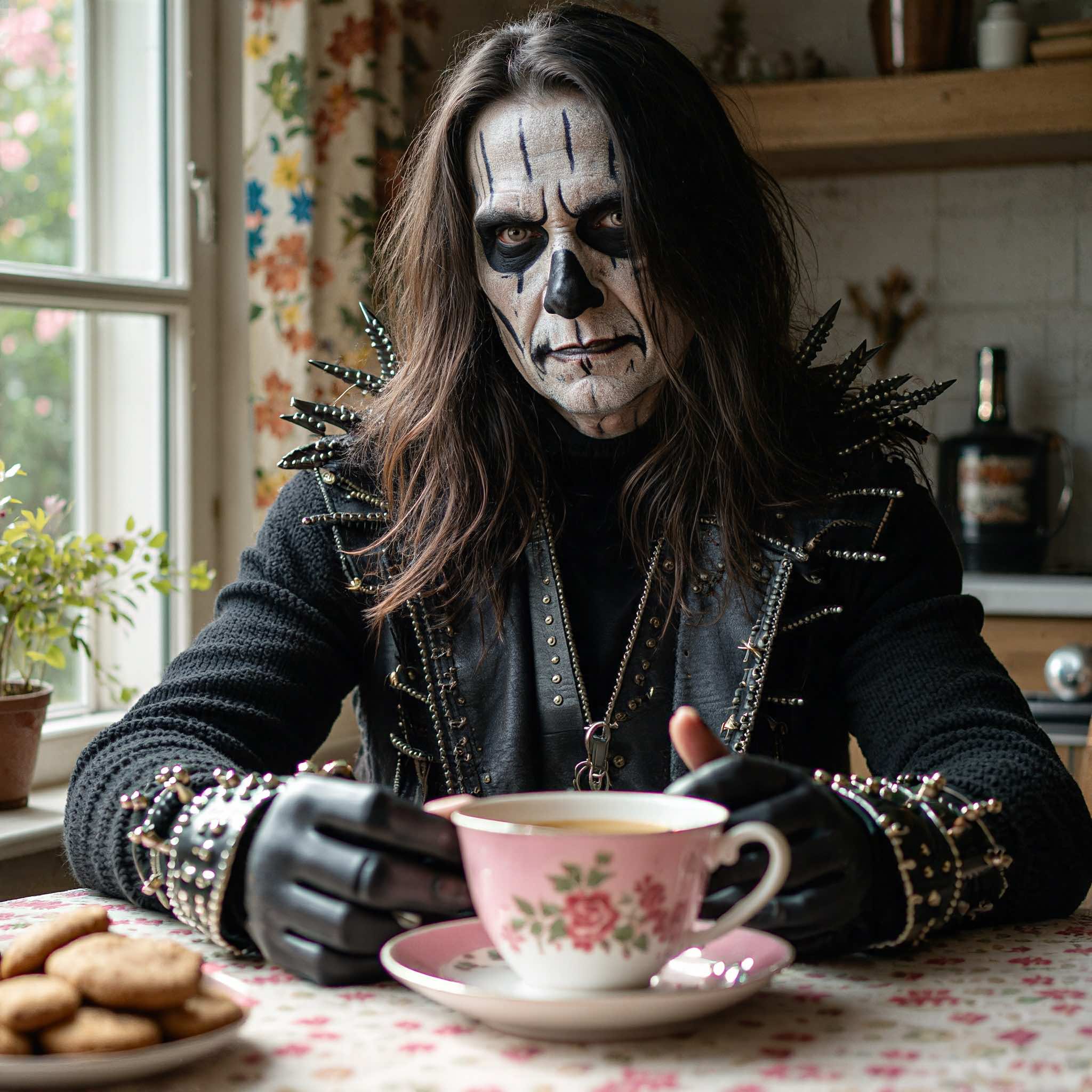} &
    \includegraphics[width=0.2\textwidth]{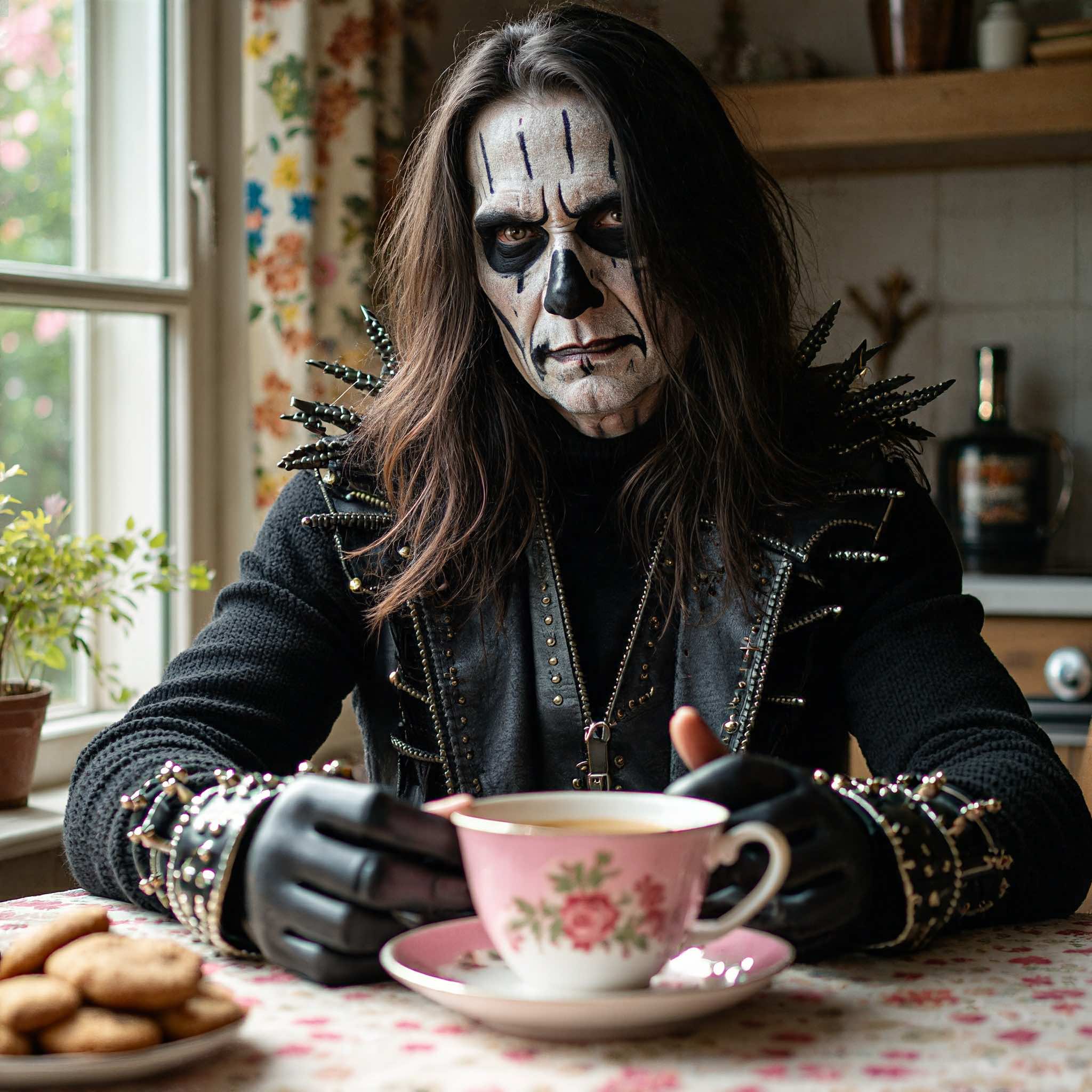} &
    \includegraphics[width=0.2\textwidth]{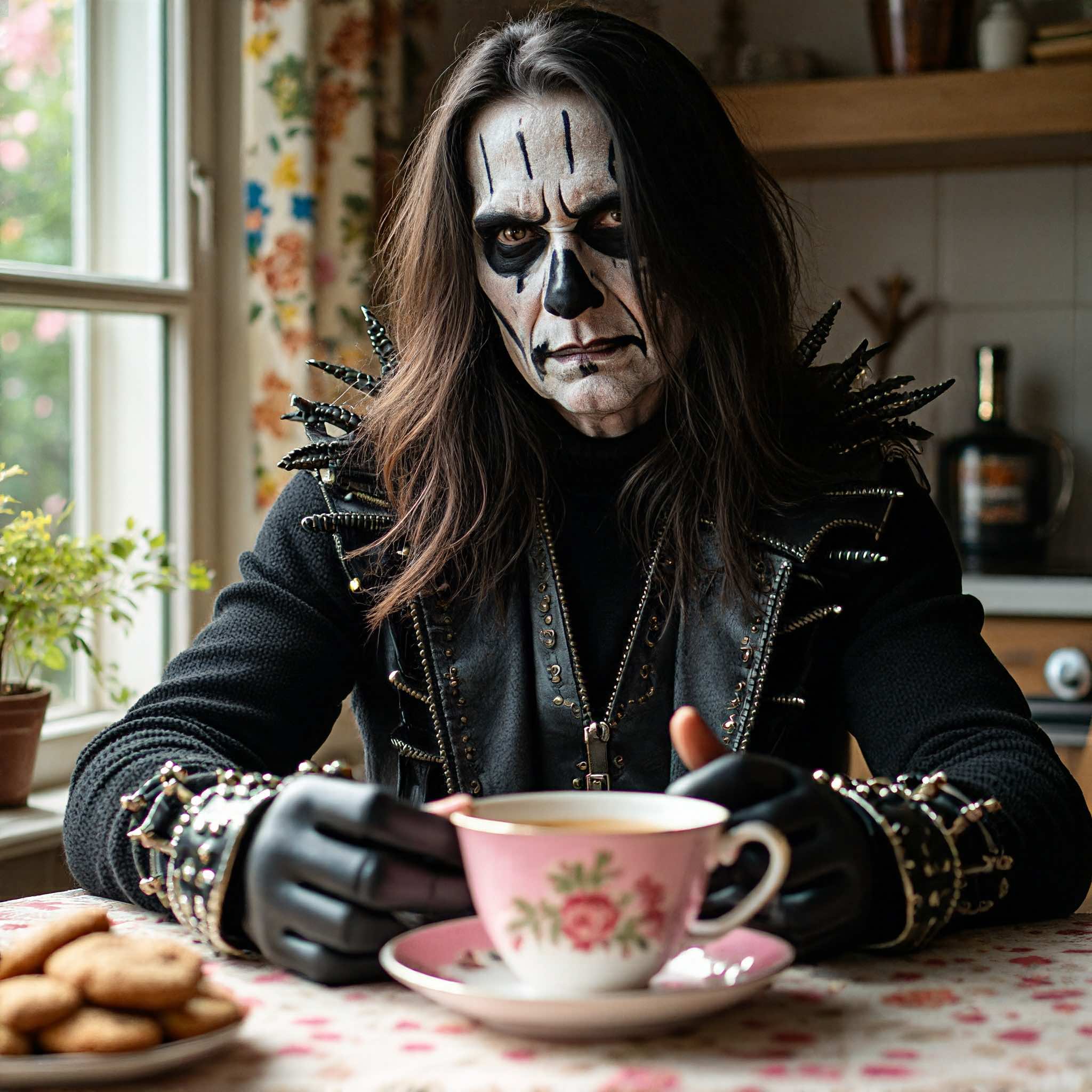} \\[-2pt]
    \scriptsize\texttt{16/28 step} & \scriptsize\texttt{20/28 step} & \scriptsize\texttt{24/28 step} & \scriptsize\texttt{26/28 step} & \scriptsize\texttt{full (28) step} 
  \end{tabular}
  \vspace{-1em}
  \caption{\textbf{VAE decoding and \ours decoding at different LDM termination steps.} \textbf{Top}: VAE decoding. \textbf{Bottom}: \ours decoding. With the full LDM denoising steps, \ours is faithful to the latent's VAE decoding results; at intermediate steps, because the base latent diffusion model has not denoised all the subtle details, it allows \ours to imagine additional details.}
  \label{fig:pid-decoding-trajectory}
  \vspace{-1\baselineskip}
\end{figure*}

\subsection{Qualitative Evaluation}
\noindent\textbf{Reconstruction on real-world image's latent.}
We visualize reconstruction quality in \Cref{fig:recon-vis} by decoding latents encoded from clean images and comparing vanilla reconstructions (FLUX.1 VAE / Scale-RAE decoder) against our \ours reconstruction. Results show that \ours preserves finer and sharper details at higher resolution. The small text is corrupted by VAE encoding and VAE decoding, however, with PiD's generative ability, it can reconstruct the correct text with high sharpness.

\noindent\textbf{Decoding at different LDM termination step.}
We show the VAE decoding and \ours decoding results at different LDM termination steps in \Cref{fig:pid-decoding-trajectory}. With the full 28 denoising steps, \ours remains closer to the original latent representation, while at intermediate step counts, \ours tends to generate additional content and can produce sharper details.

\begin{figure*}[t]
  \centering
  \setlength{\tabcolsep}{0pt}
  \renewcommand{\arraystretch}{1}
  \vspace{-0.5em}
  \begin{tabular}{@{}ccccc@{}}
    \scriptsize\strut VAE Decode $(512^2)$ & \scriptsize\strut +SeedVR2-3B $(2048^2)$ & \scriptsize\strut +InvSR $(2048^2)$ & \scriptsize\strut +TSD-SR $(2048^2)$ & \scriptsize\strut Ours $(2048^2)$\\
    \imgwithlabel{0.2\textwidth}{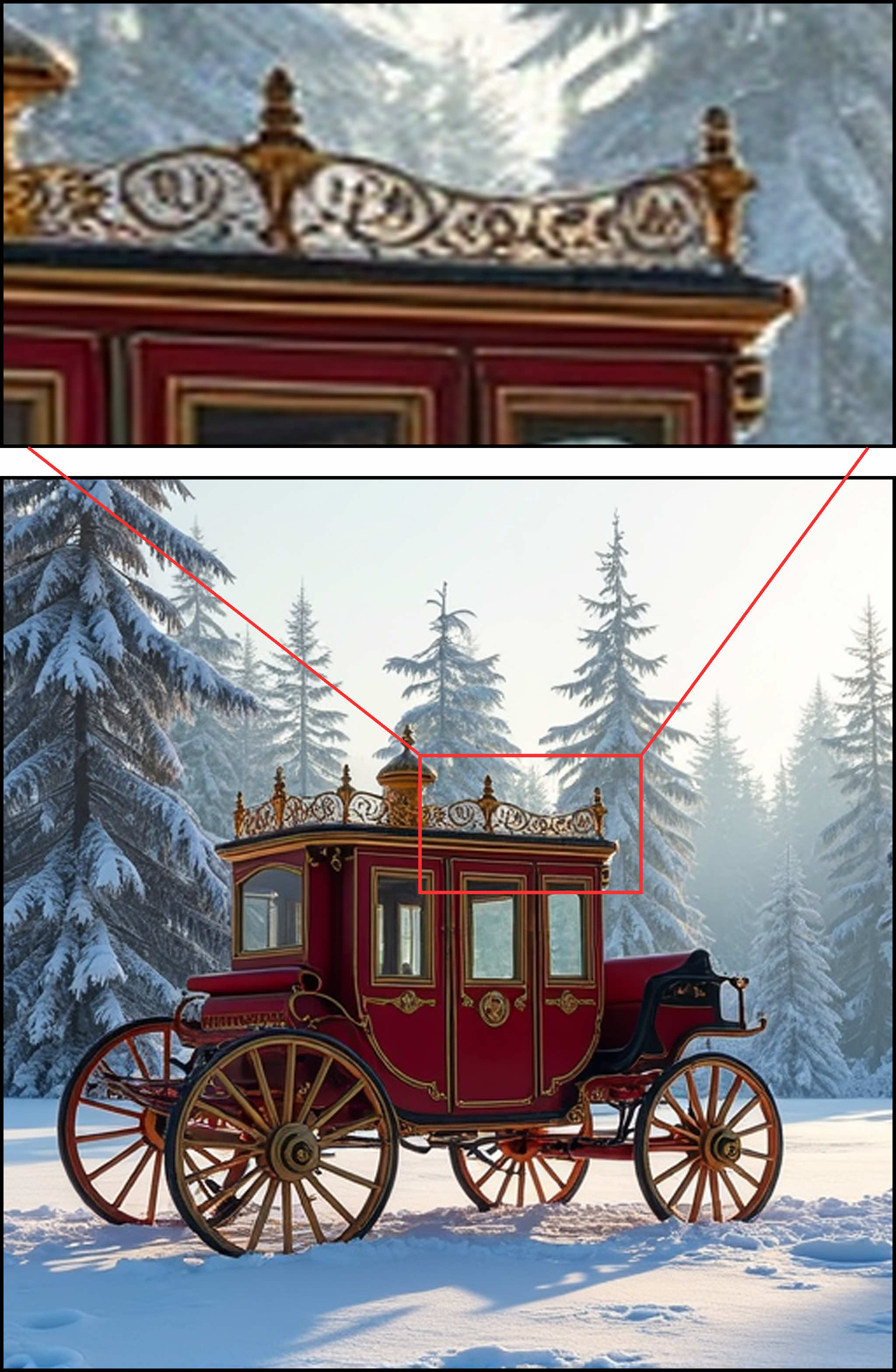}{18.25 ms}&
    \imgwithlabel{0.2\textwidth}{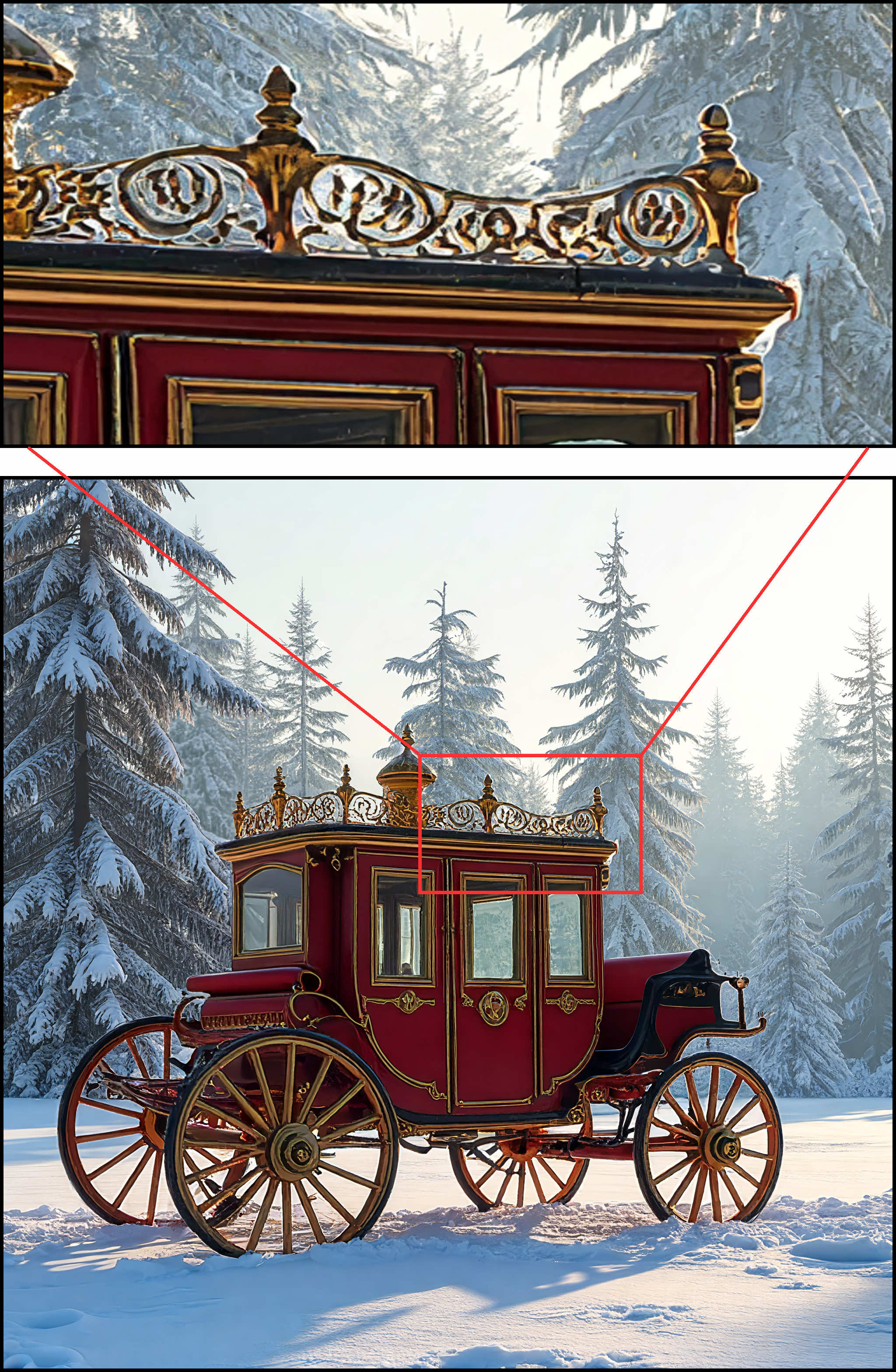}{1237.4 ms}&
    \imgwithlabel{0.2\textwidth}{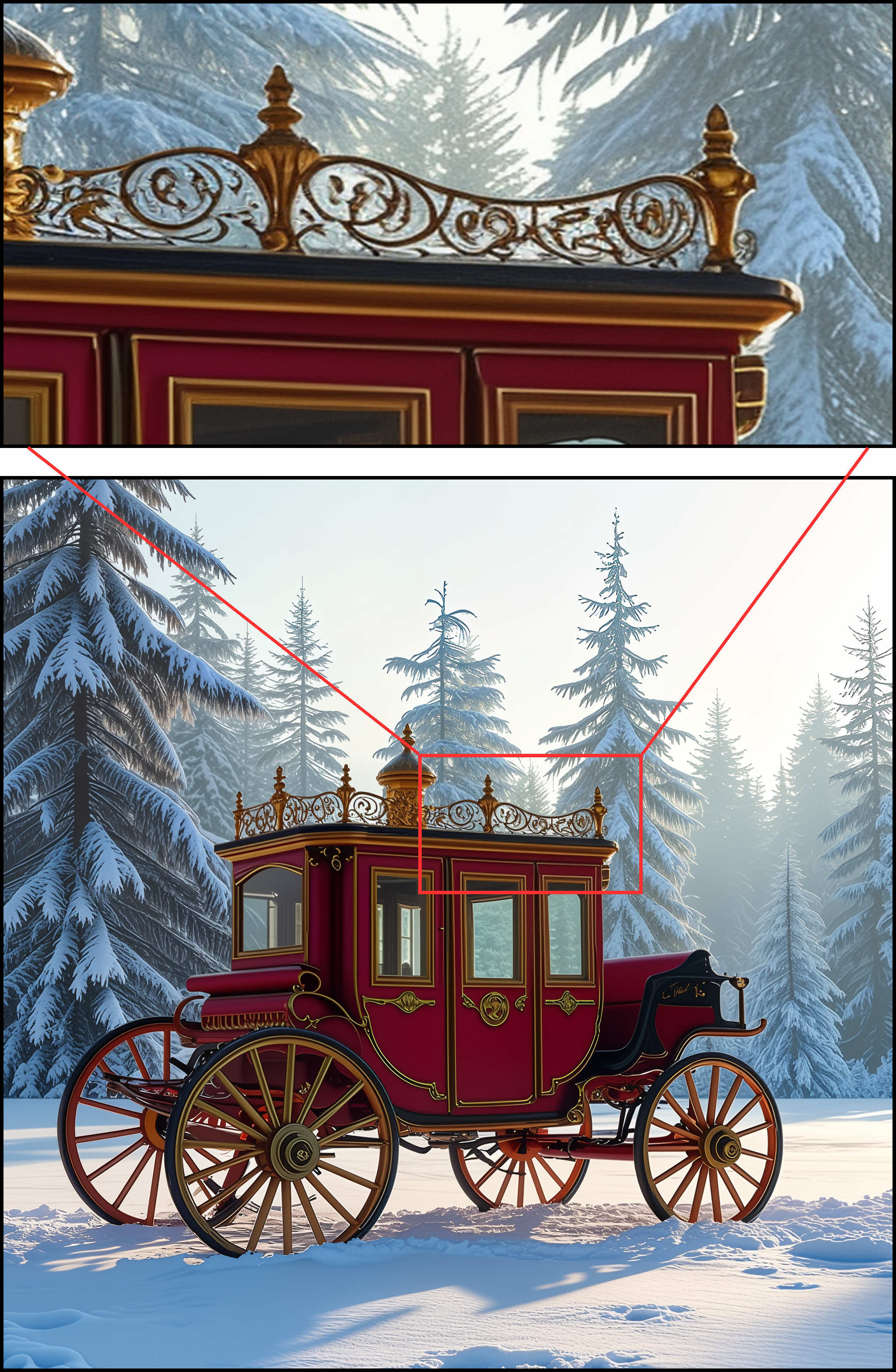}{1017.7 ms}&
    \imgwithlabel{0.2\textwidth}{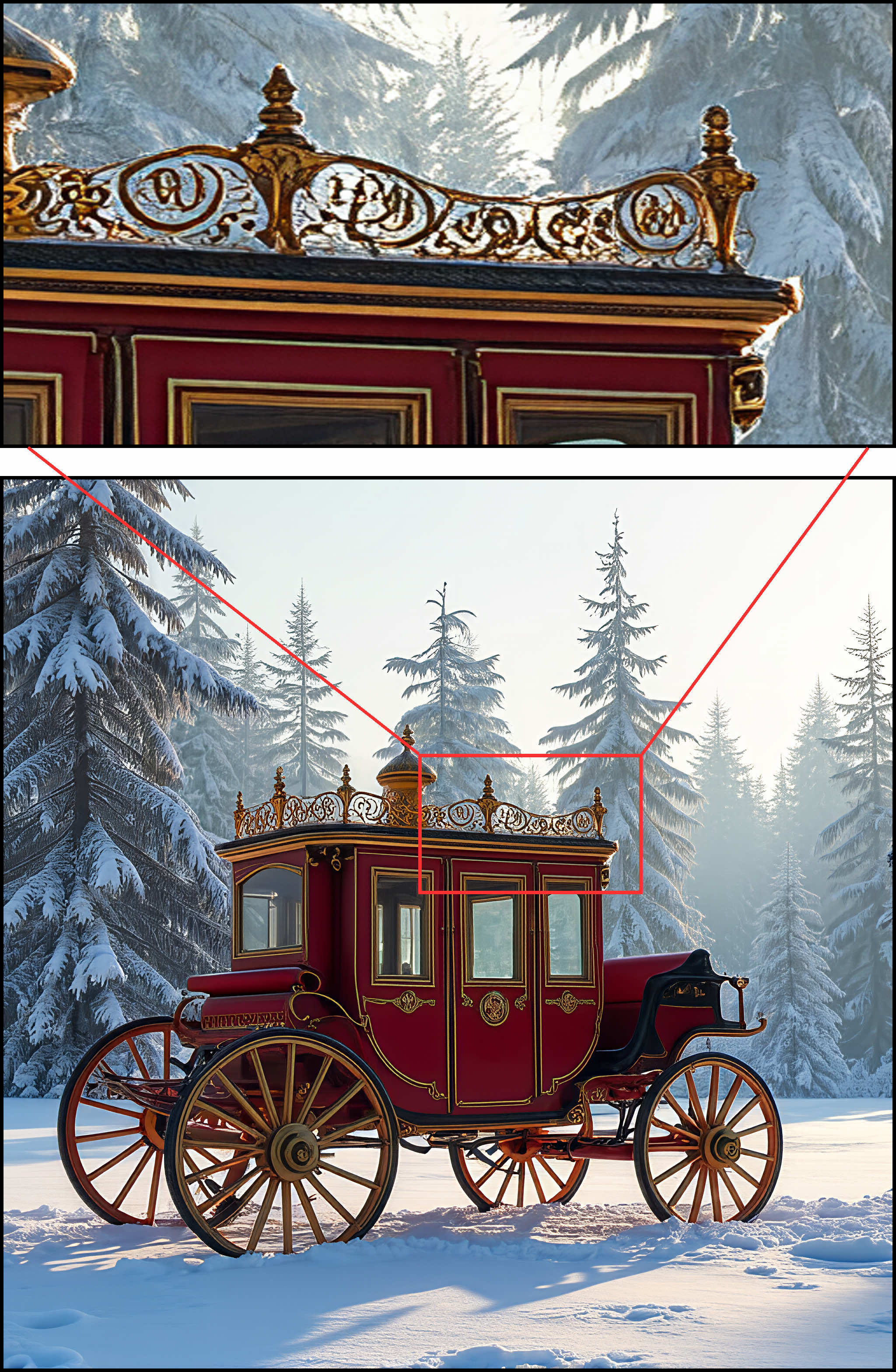}{724.7 ms}&
    \imgwithlabel{0.2\textwidth}{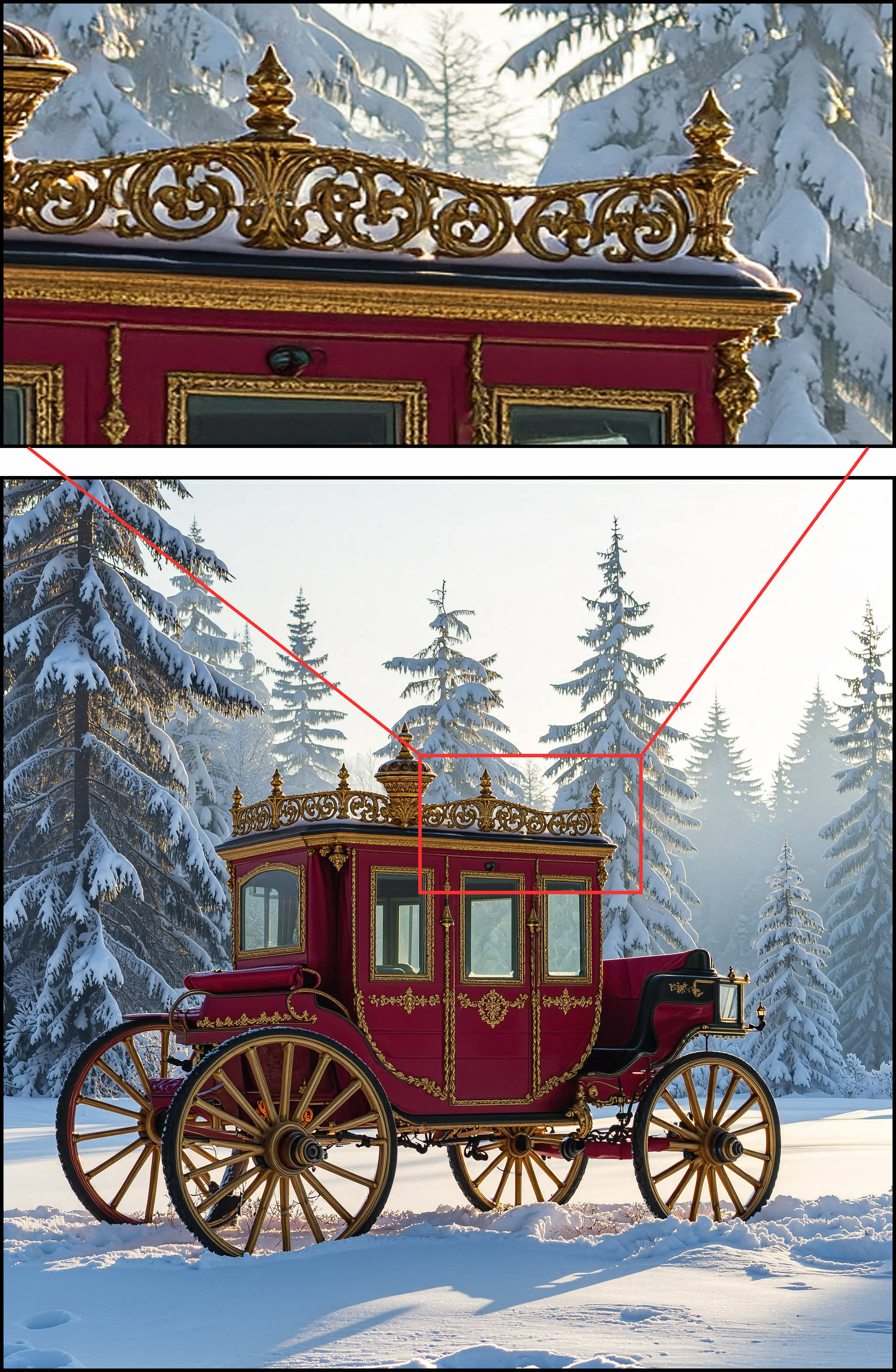}{211.2 ms}\\[0.5em]
    \imgwithlabel{0.2\textwidth}{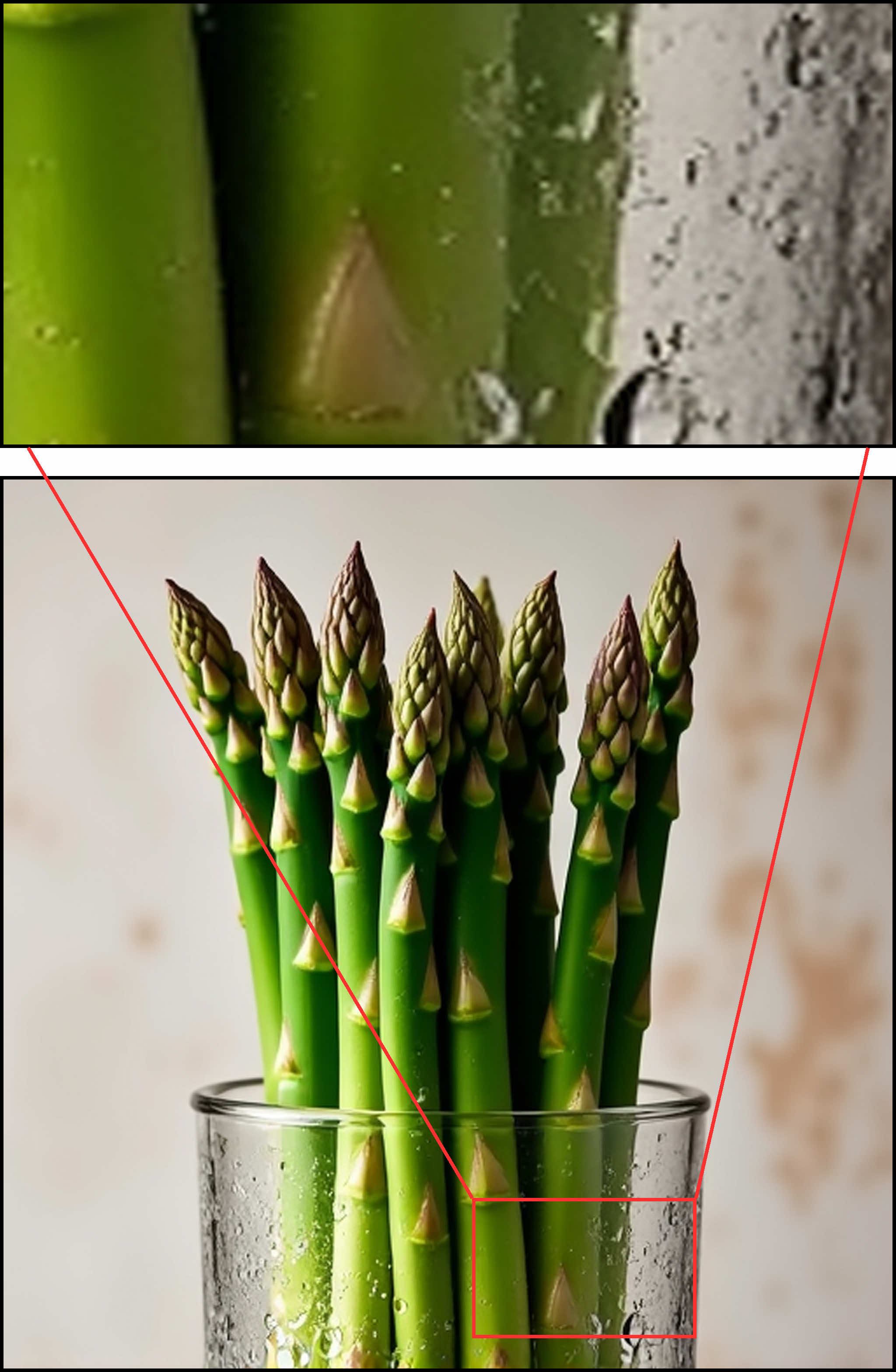}{18.25 ms}&
    \imgwithlabel{0.2\textwidth}{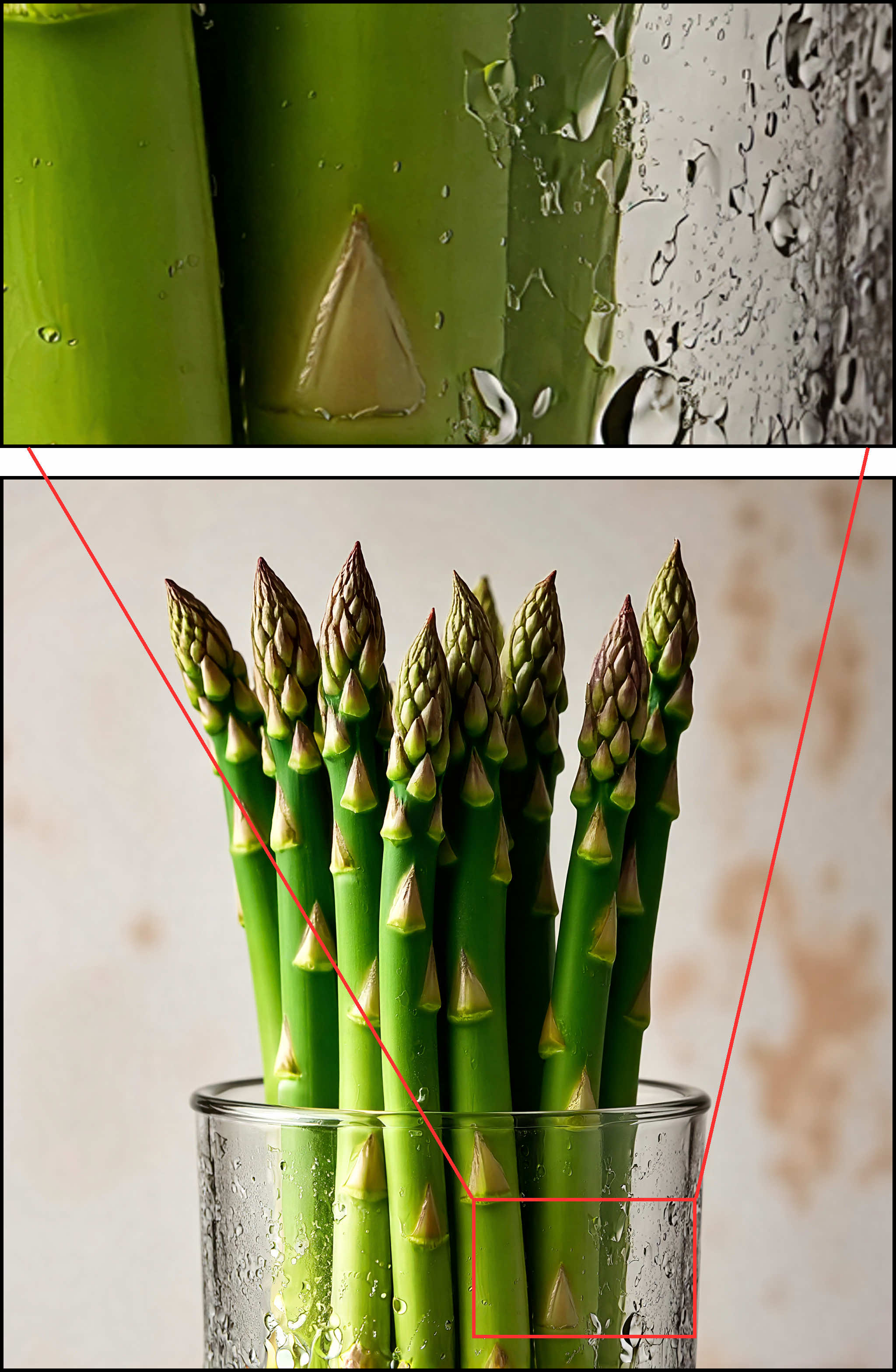}{1237.4 ms}&
    \imgwithlabel{0.2\textwidth}{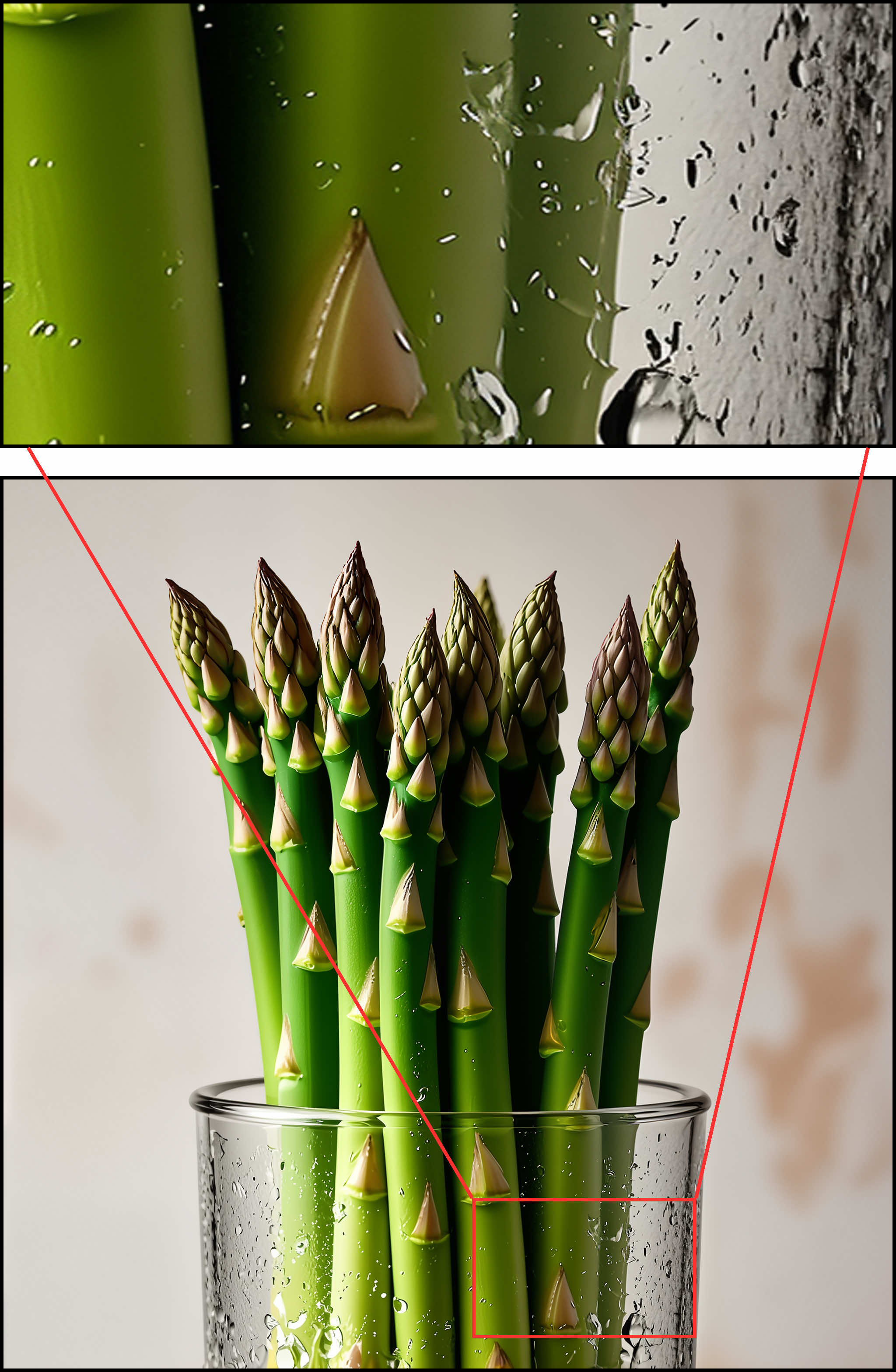}{1017.7 ms}&
    \imgwithlabel{0.2\textwidth}{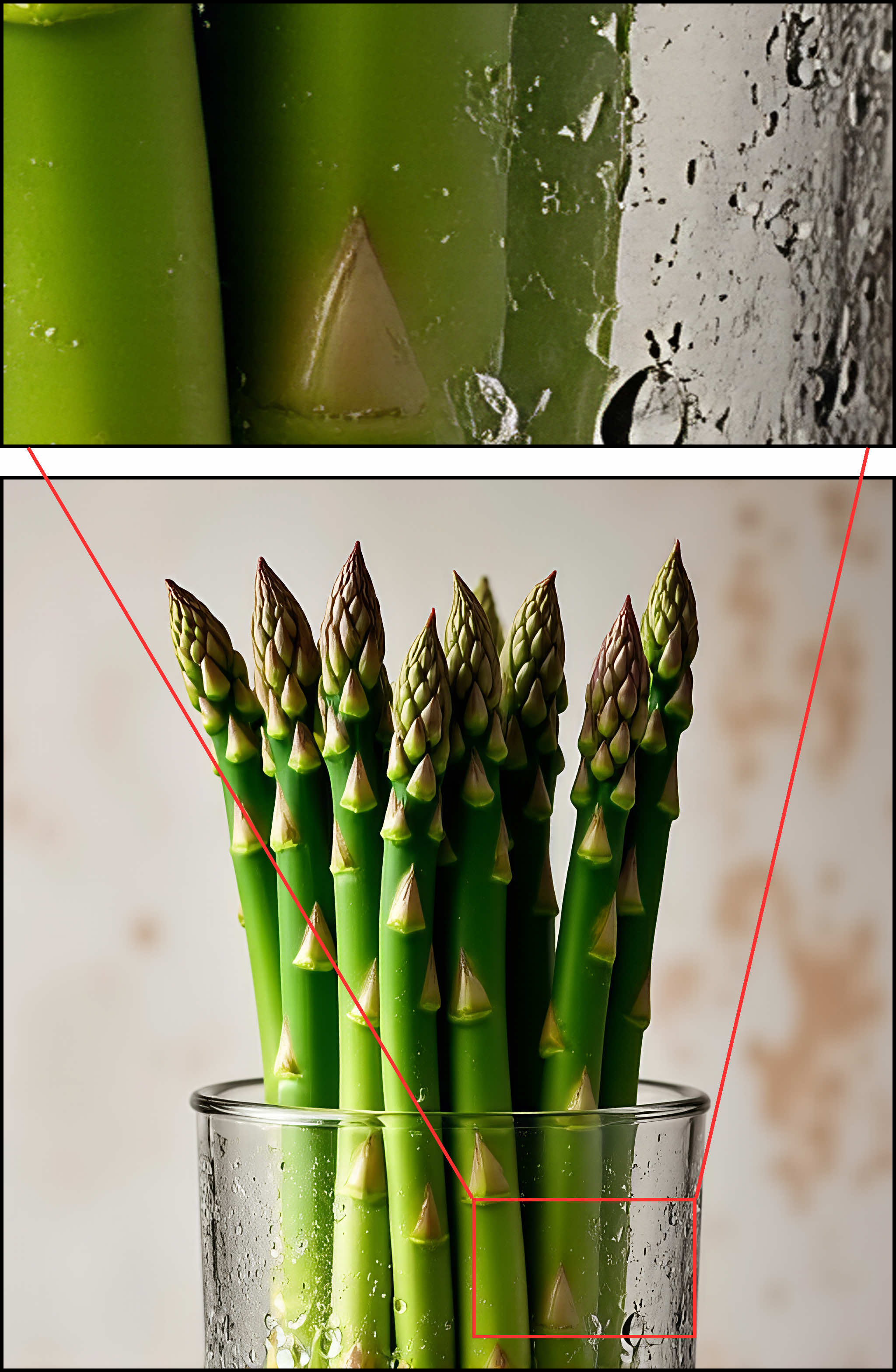}{724.7 ms}&
    \imgwithlabel{0.2\textwidth}{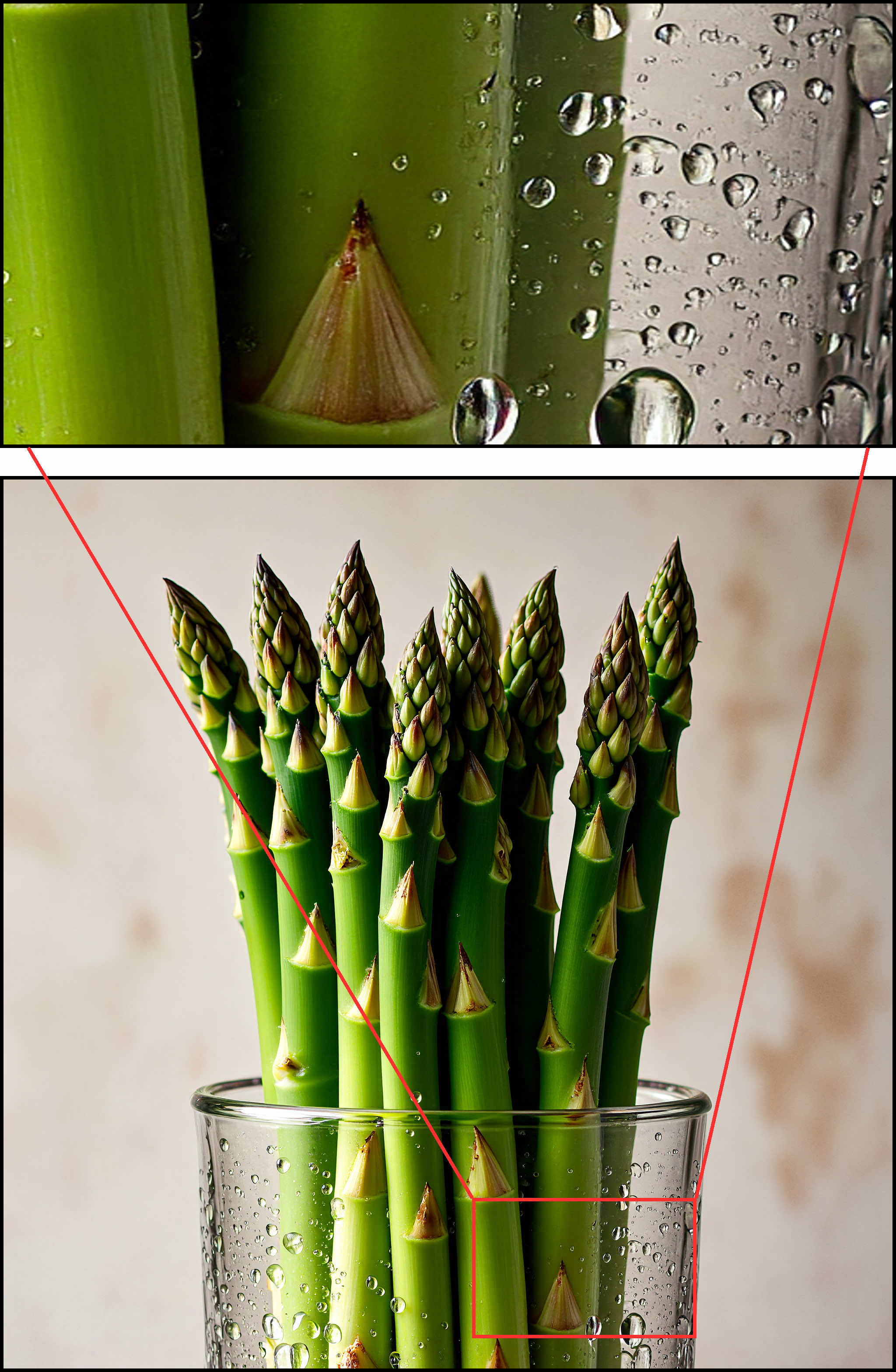}{211.2 ms}\\[0.5em]
    \imgwithlabel{0.2\textwidth}{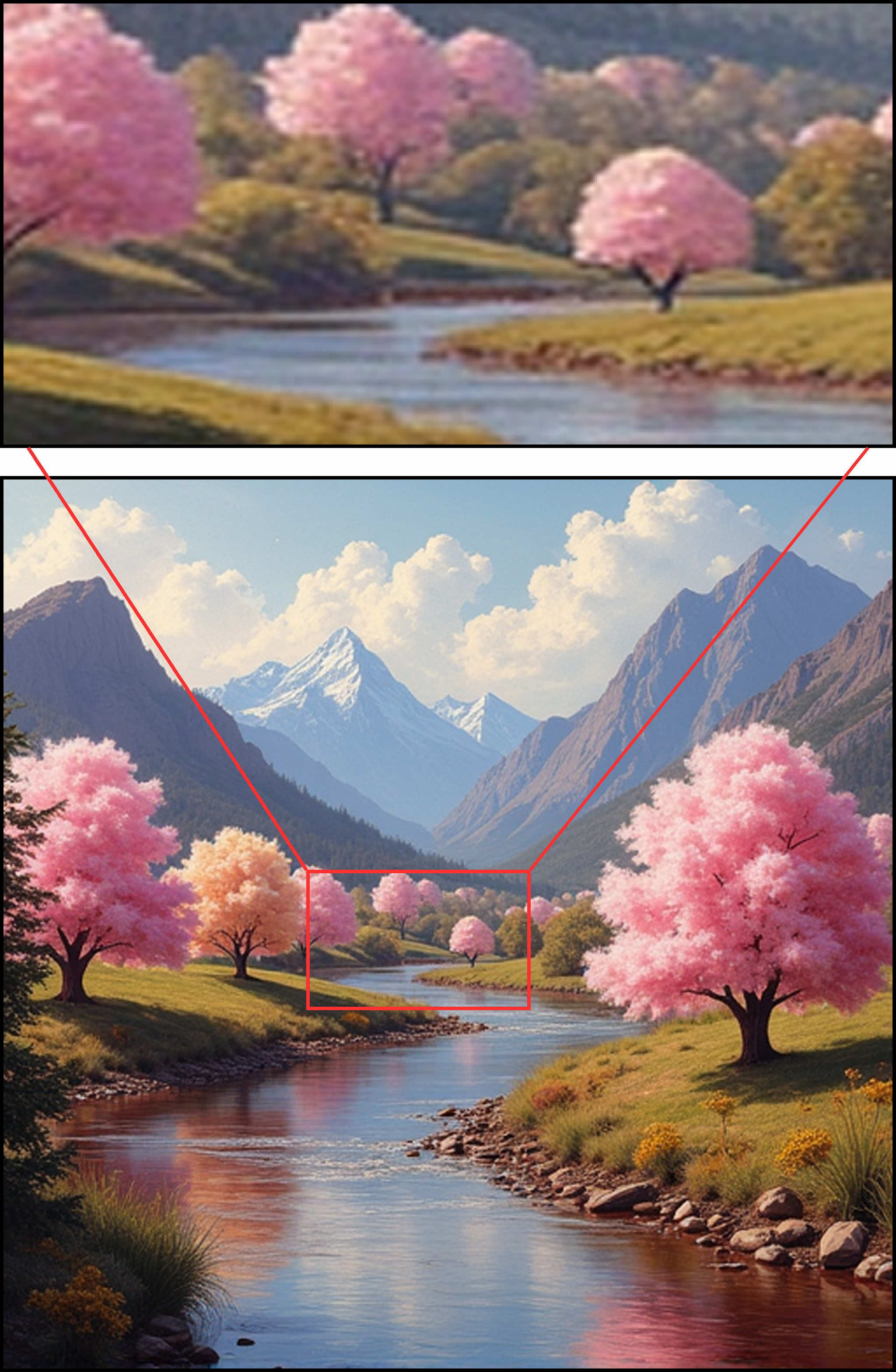}{18.25 ms}&
    \imgwithlabel{0.2\textwidth}{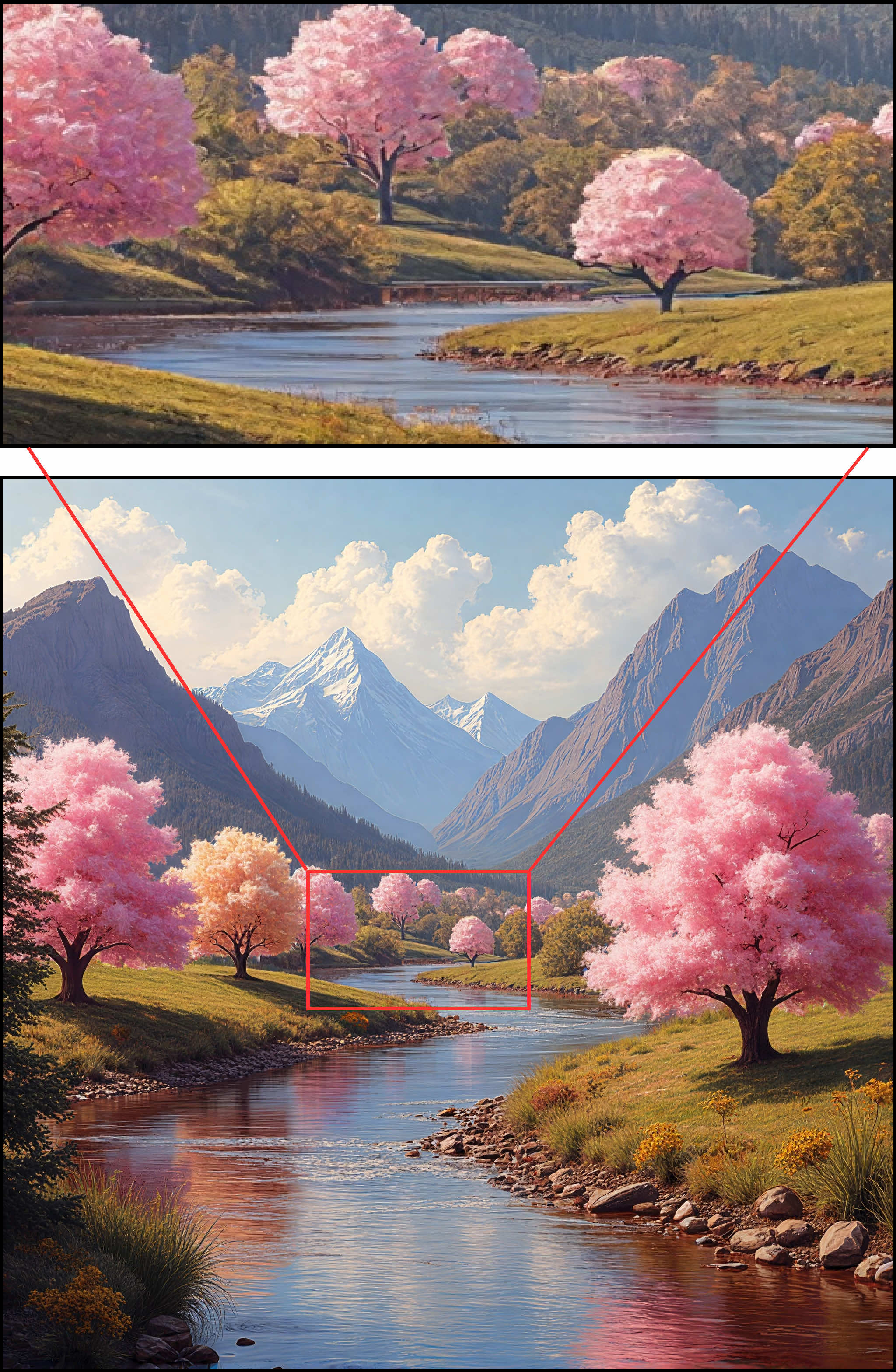}{1237.4 ms}&
    \imgwithlabel{0.2\textwidth}{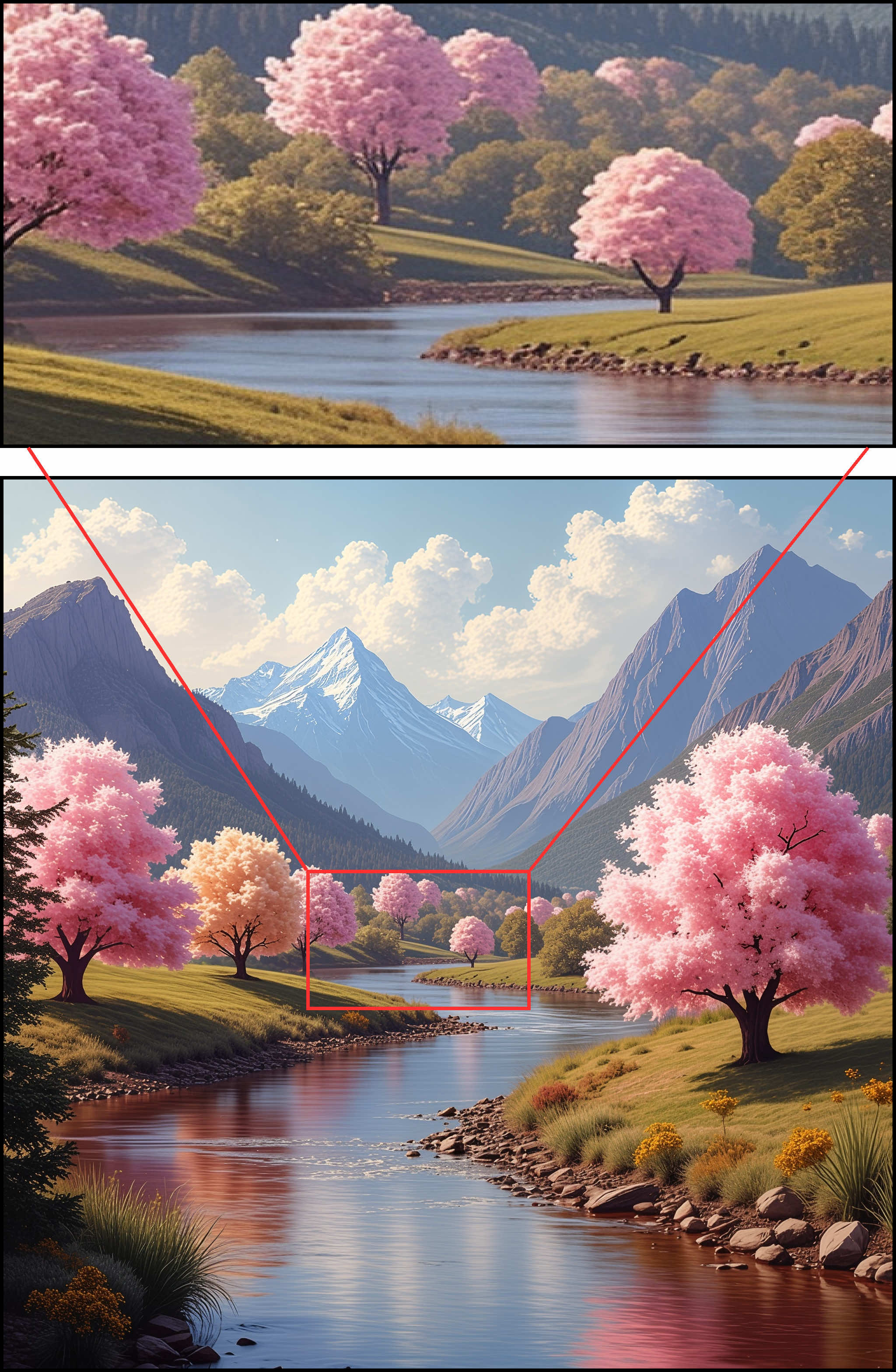}{1017.7 ms}&
    \imgwithlabel{0.2\textwidth}{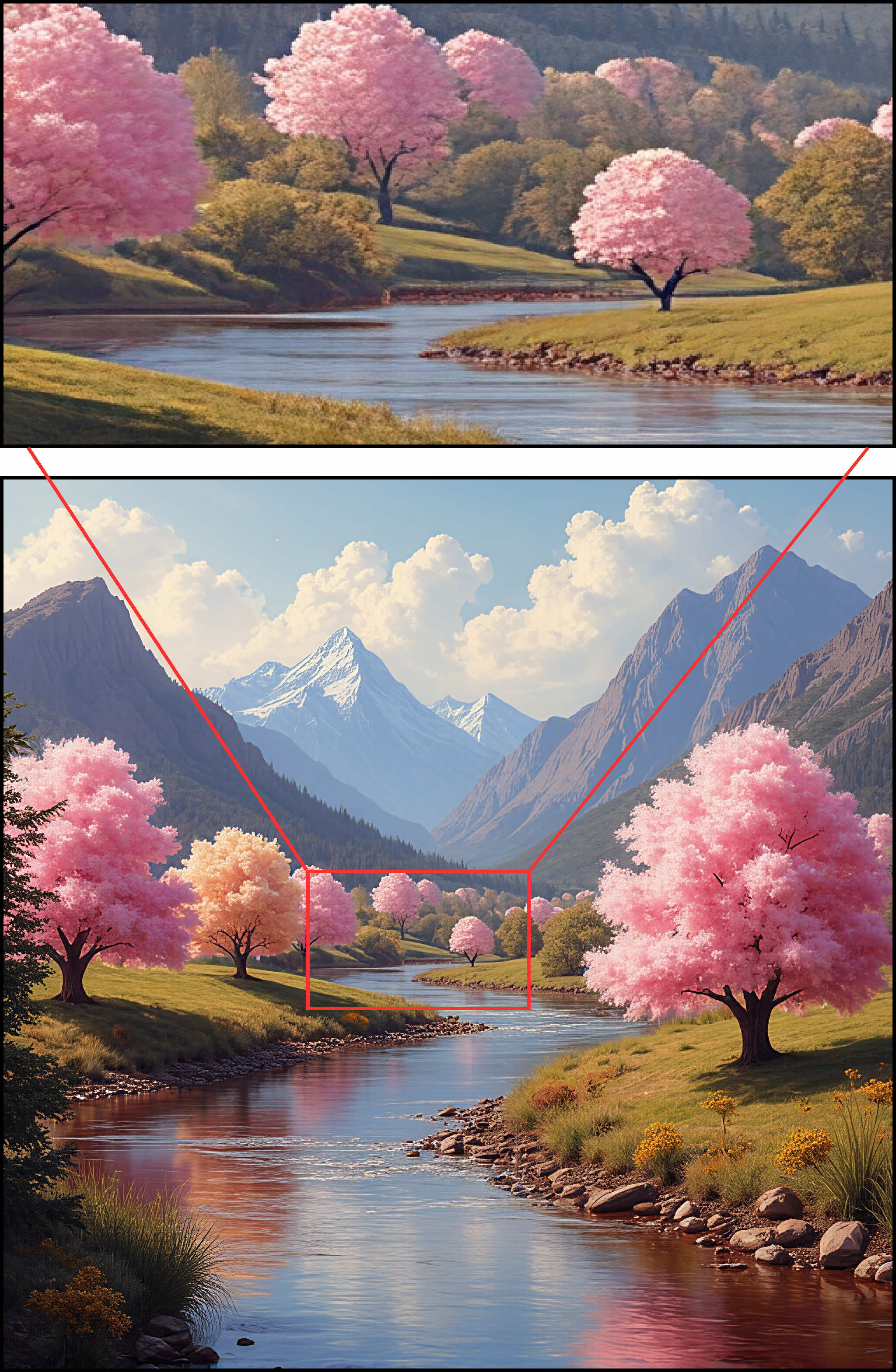}{724.7 ms}&
    \imgwithlabel{0.2\textwidth}{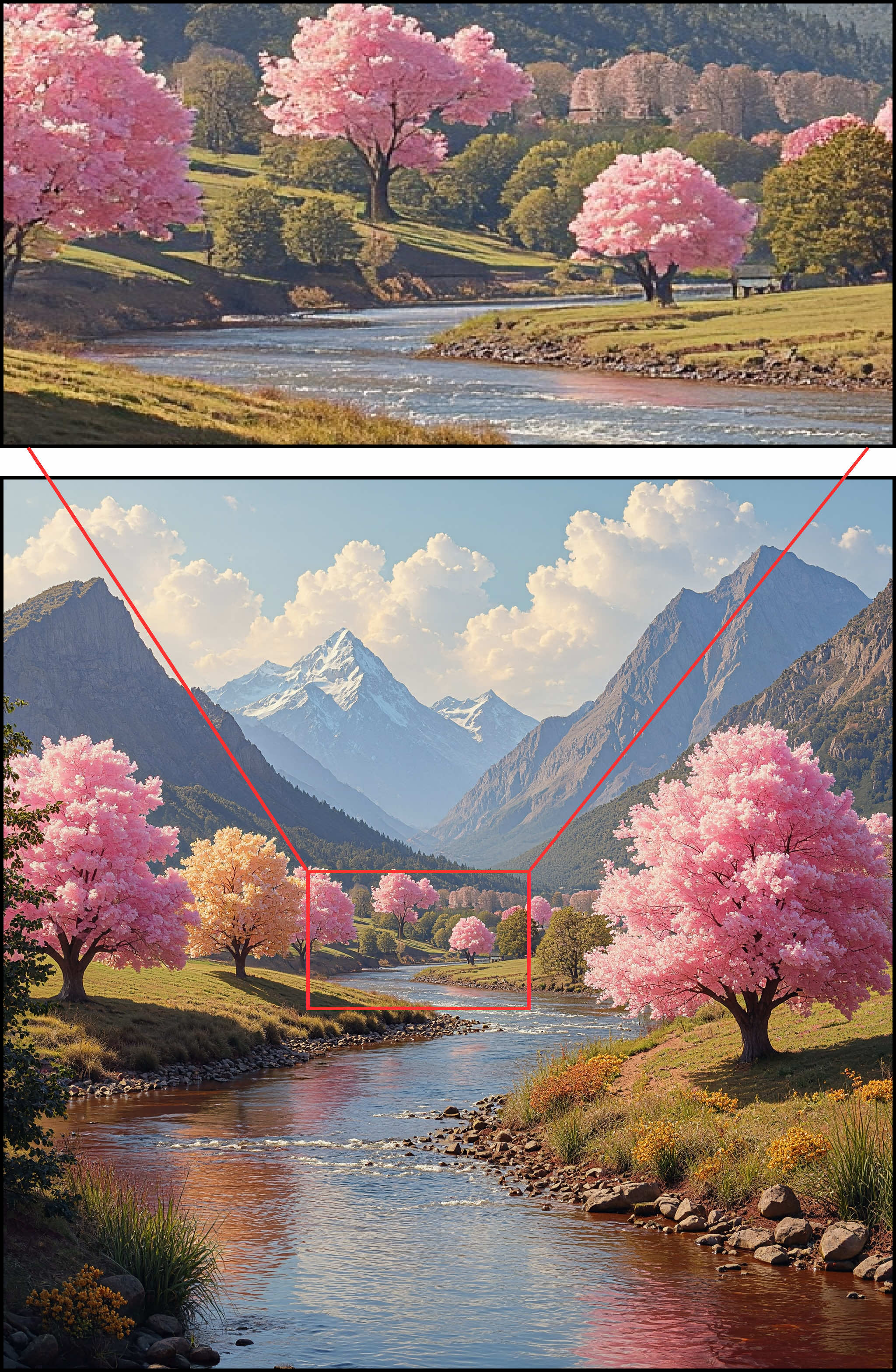}{211.2 ms}\\[0.5em]
    \imgwithlabel{0.2\textwidth}{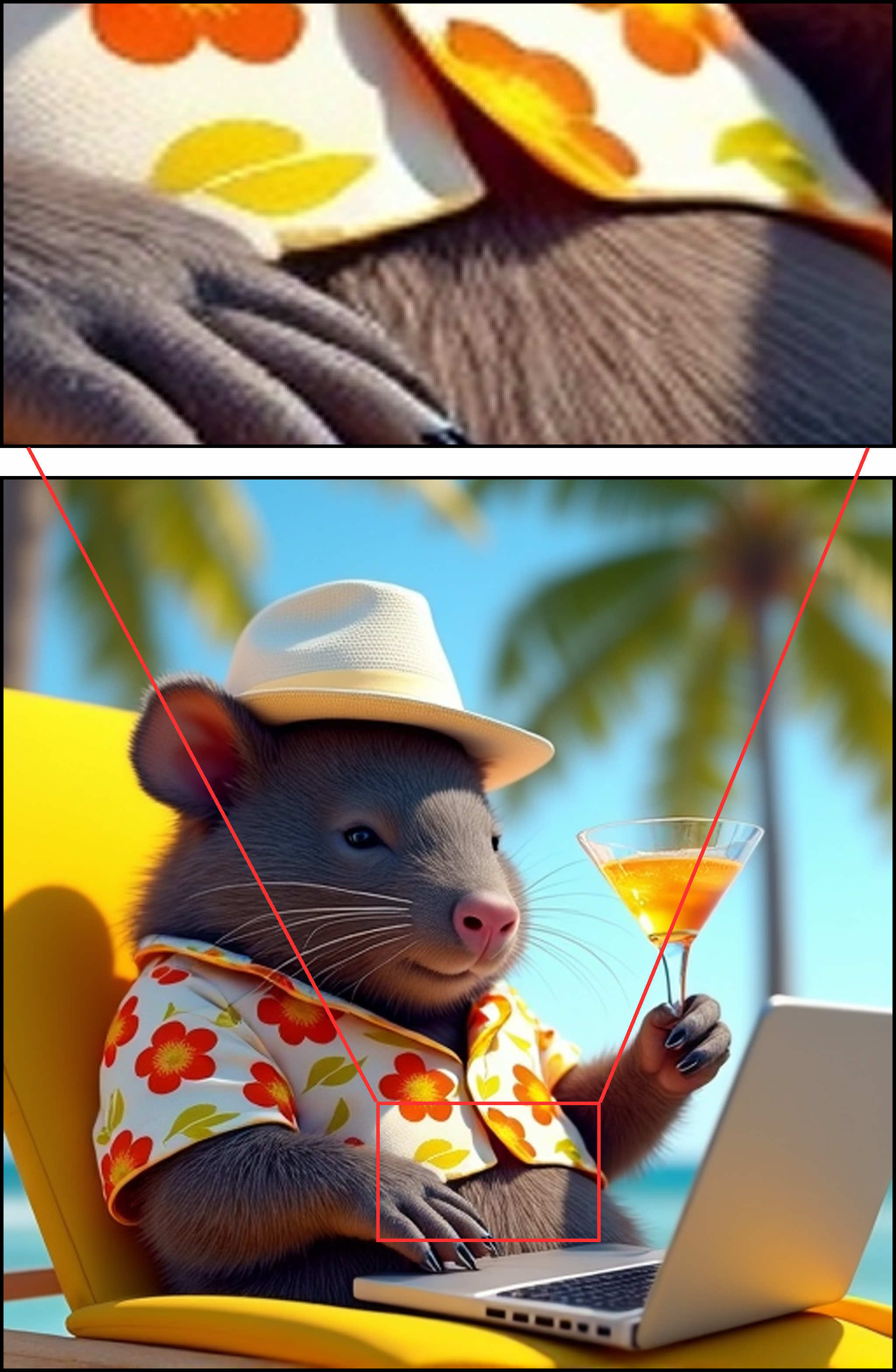}{18.25 ms}&
    \imgwithlabel{0.2\textwidth}{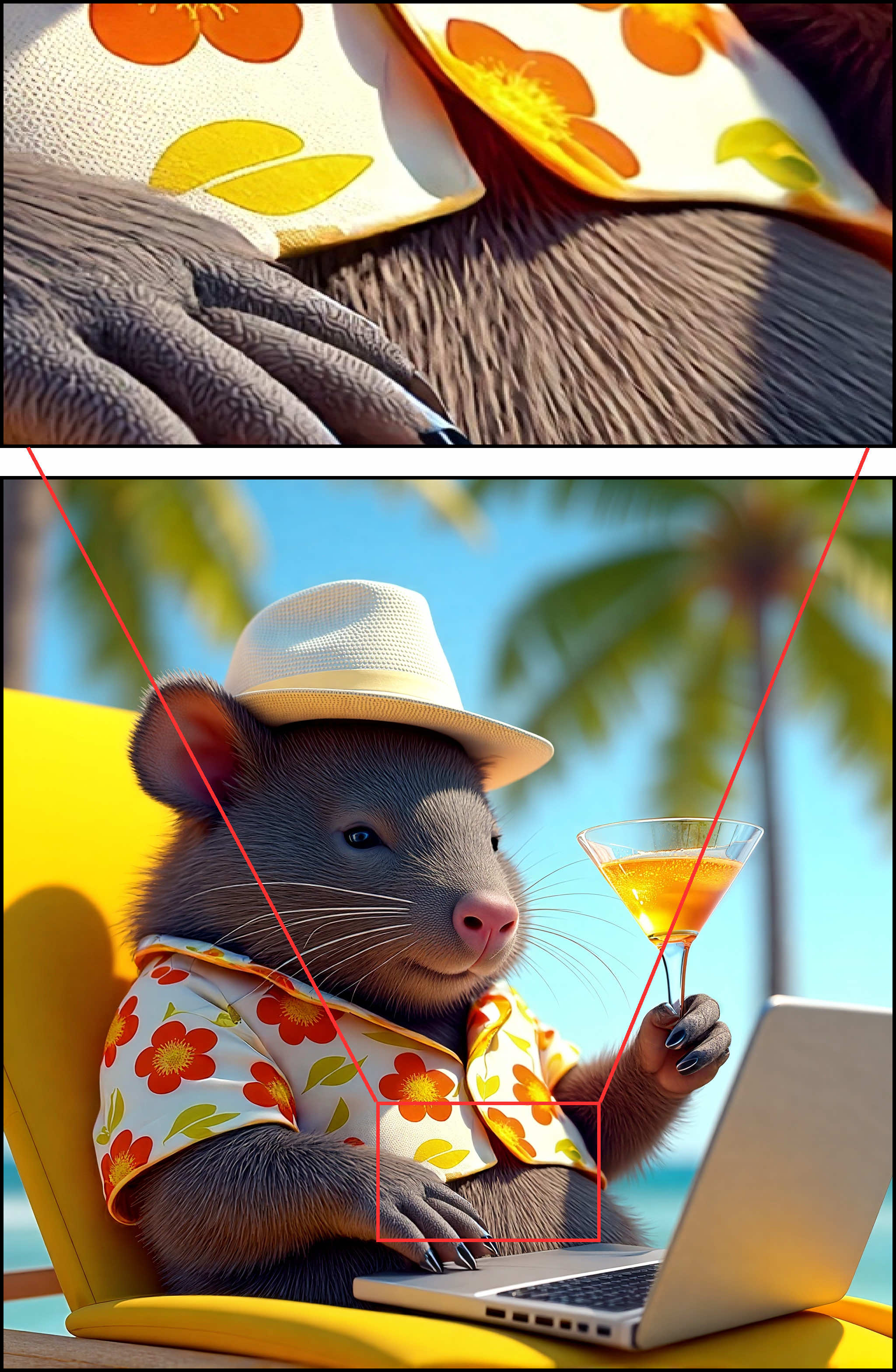}{1237.4 ms}&
    \imgwithlabel{0.2\textwidth}{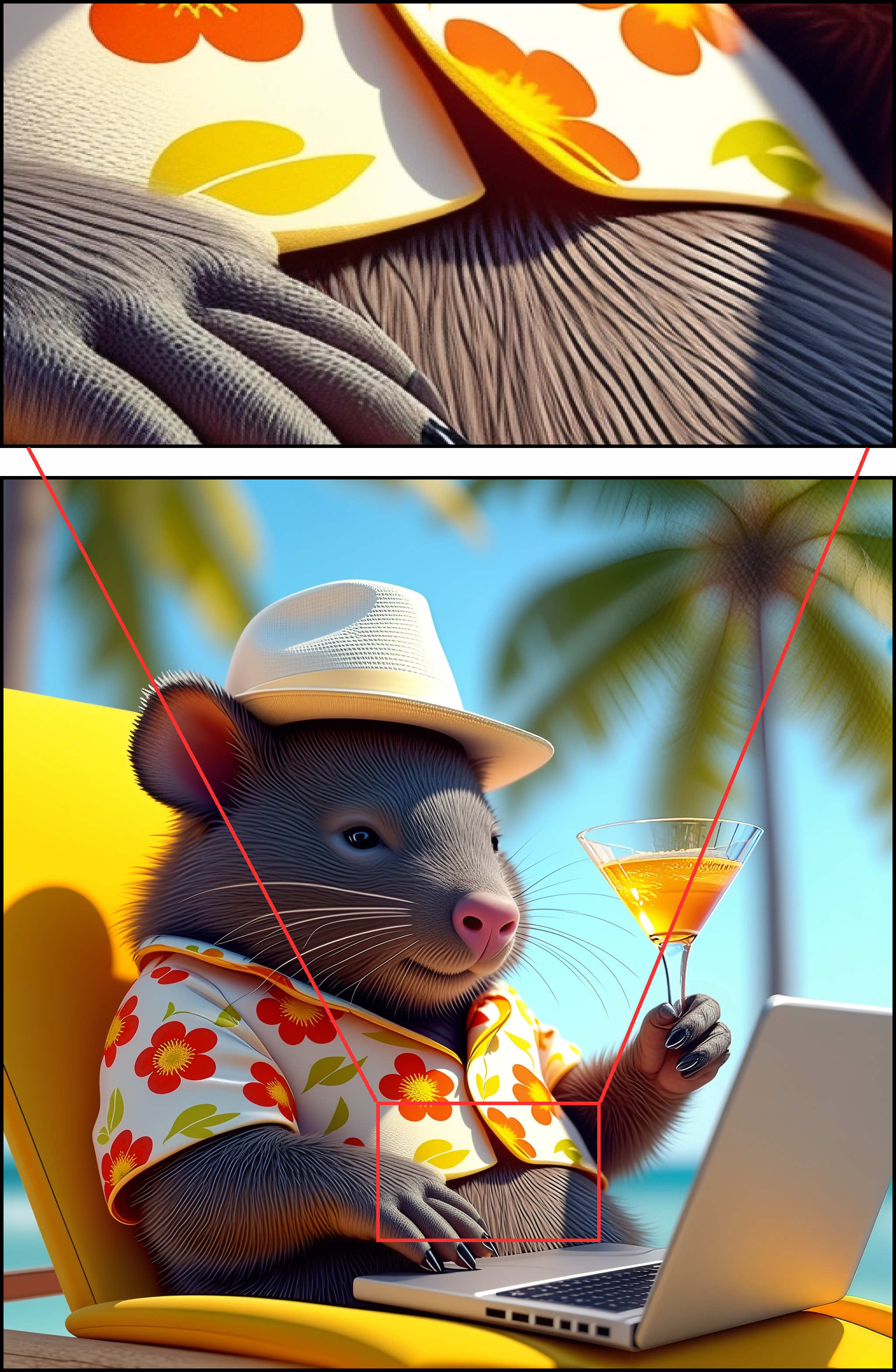}{1017.7 ms}&
    \imgwithlabel{0.2\textwidth}{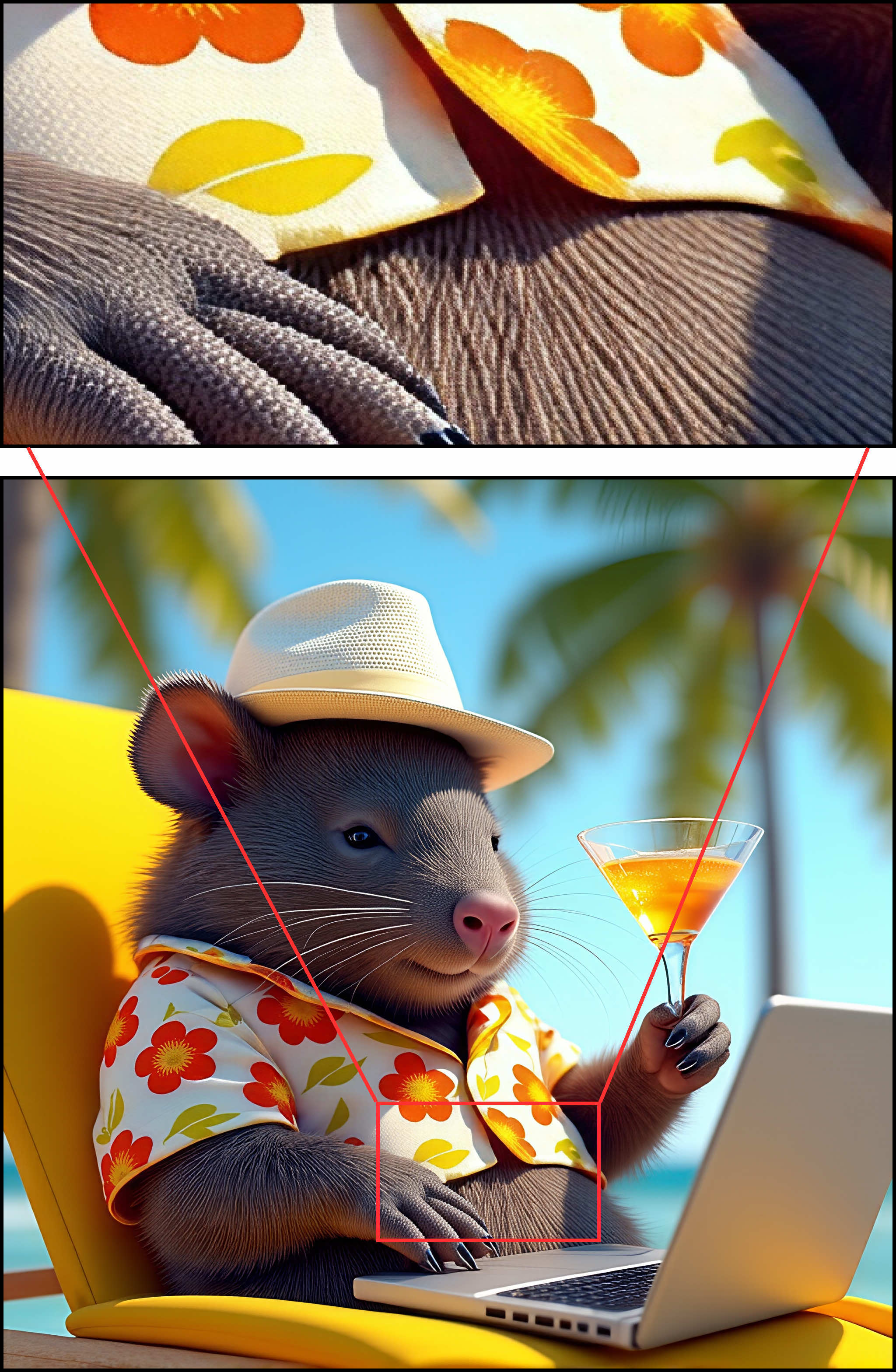}{724.7 ms}&
    \imgwithlabel{0.2\textwidth}{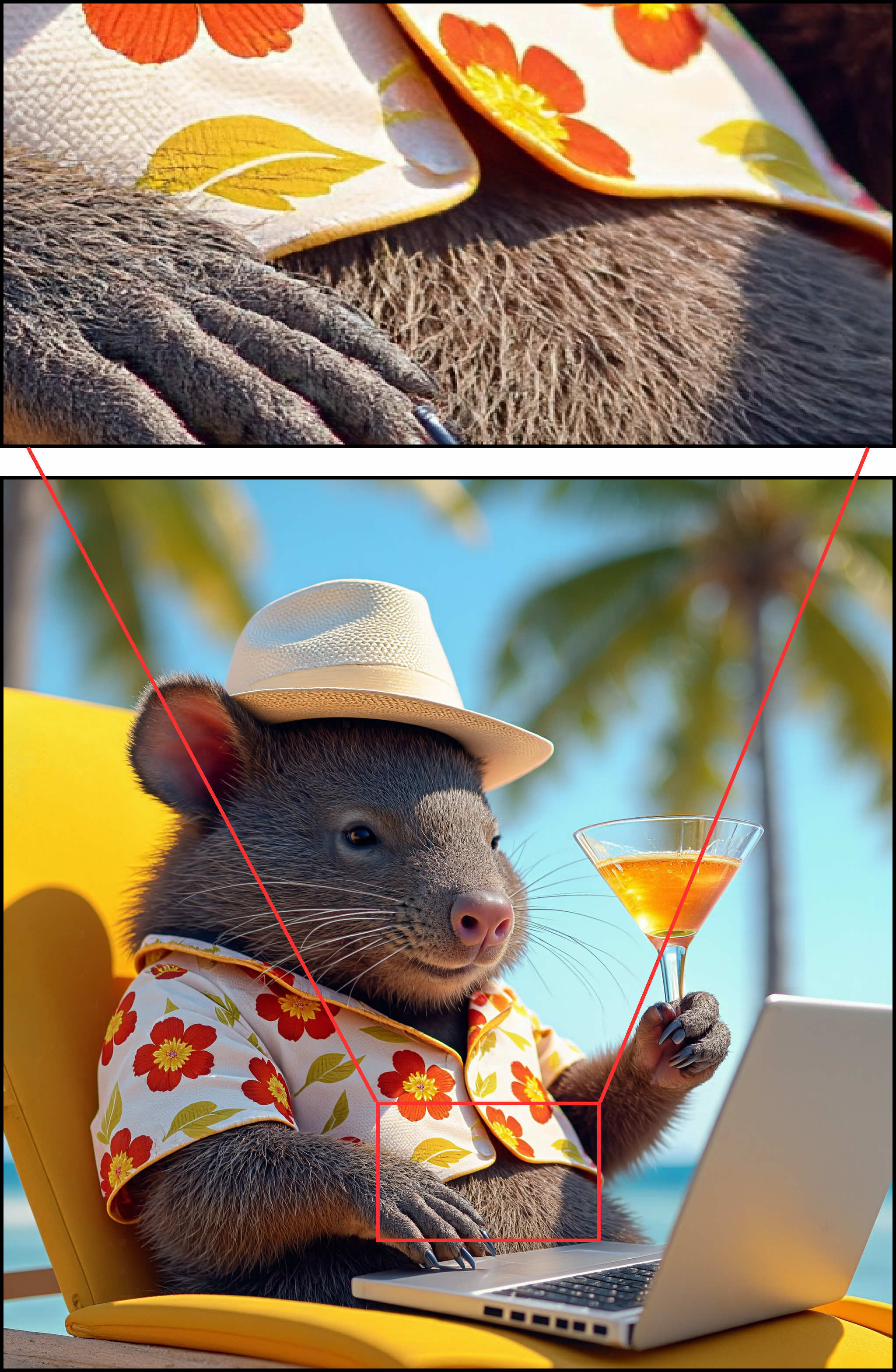}{211.2 ms}\\
    \scriptsize\strut VAE Decode $(512^2)$ & \scriptsize\strut +SeedVR2-3B $(2048^2)$ & \scriptsize\strut +InvSR $(2048^2)$ & \scriptsize\strut +TSD-SR $(2048^2)$ & \scriptsize\strut Ours $(2048^2)$\\[0.8em]
  \end{tabular}
  \vspace{-0.8em}
  \caption{\textbf{\ours vs.\ cascaded super-resolution.} From a FLUX.1 [dev] latent of a $512^2$ image, baselines apply a super-resolution model on VAE decoding output, while \ours decodes directly to $2048^2$. \ours produces sharper detail at lower latency. Latency is measured on a single GB200 GPU with \texttt{torch.compile}.}
  \label{fig:generation-vis}
  \vspace{-1em}
\end{figure*}

\noindent\textbf{Compare with super-resolution baselines.}
We further compare image-generation decoding in \Cref{fig:generation-vis}. Starting from FLUX.1 [dev] latents of $512^2$ images, we compare \ours with cascaded super-resolution baselines, showing that \ours yields richer details with lower inference latency than diffusion-based SR methods. Here we use \ours\texttt{(24/28)} in the figure.

\clearpage

\begin{figure*}[t]
  \centering

  \sbox{\infcosttblbox}{%
    \begin{minipage}{0.66\textwidth}
      \resizebox{0.95\columnwidth}{!}{%
\begin{tabular}{@{}lcccccc@{}}
\toprule
\multicolumn{7}{c}{\textbf{PiD Latency (ms)}}                               \\ \midrule
\textbf{GPU Type} & \textbf{Compile} & \textbf{256px} & \textbf{512px} & \textbf{1024px} & \textbf{2048px} & \textbf{4096px} \\ \midrule
RTX 5090  & \xmark & 79.1  & 114.1 & 273.1 & 1388.8 & OOM    \\
RTX 5090  & \cmark & 52.5  & 78.4  & 188.2 & 979.3  & 9238.0 \\
H100      & \xmark & 272.2 & 279.3 & 211.6 & 797.0  & 4763.4 \\
H100      & \cmark & 36.5  & 45.3  & 88.4  & 446.0  & 3754.6 \\
GB200     & \xmark & 265.1 & 260.8 & 251.2 & 505.1  & 2944.1 \\
GB200     & \cmark & 32.2  & 33.0  & 57.0  & 208.8  & 1927.3 \\ \midrule
\multicolumn{7}{c}{\textbf{Decoder Memory Usage (GB)}}                      \\ \midrule
\textbf{Decoder}  & \textbf{Compile} & \textbf{256px} & \textbf{512px} & \textbf{1024px} & \textbf{2048px} & \textbf{4096px} \\ \midrule
FLUX.1 VAE & \xmark & 0.3   & 0.7   & 2.6   & 37.0   & OOM    \\
FLUX.1 VAE & \cmark & 0.4   & 0.8   & 2.4   & 16.7   & OOM    \\
PiD       & \xmark & 12.6  & 12.8  & 13.6  & 16.5   & 28.6   \\
PiD       & \cmark & 10.3  & 10.3  & 10.9  & 13.0   & 22.5   \\ \bottomrule
\end{tabular}%
}
    \end{minipage}%
  }

  \begin{minipage}[t]{0.55\textwidth}
    \vspace{0pt}
    \centering
    \captionof{table}{\textbf{\ours latency and decoder memory usage} across output resolutions, compared with the original VAE decoder.}
    \label{tab:pid_latency_memory}
    \resizebox{0.95\columnwidth}{!}{%
\begin{tabular}{@{}lcccccc@{}}
\toprule
\multicolumn{7}{c}{\textbf{PiD Latency (ms)}}                               \\ \midrule
\textbf{GPU Type} & \textbf{Compile} & \textbf{256px} & \textbf{512px} & \textbf{1024px} & \textbf{2048px} & \textbf{4096px} \\ \midrule
RTX 5090  & \xmark & 79.1  & 114.1 & 273.1 & 1388.8 & OOM    \\
RTX 5090  & \cmark & 52.5  & 78.4  & 188.2 & 979.3  & 9238.0 \\
H100      & \xmark & 272.2 & 279.3 & 211.6 & 797.0  & 4763.4 \\
H100      & \cmark & 36.5  & 45.3  & 88.4  & 446.0  & 3754.6 \\
GB200     & \xmark & 265.1 & 260.8 & 251.2 & 505.1  & 2944.1 \\
GB200     & \cmark & 32.2  & 33.0  & 57.0  & 208.8  & 1927.3 \\ \midrule
\multicolumn{7}{c}{\textbf{Decoder Memory Usage (GB)}}                      \\ \midrule
\textbf{Decoder}  & \textbf{Compile} & \textbf{256px} & \textbf{512px} & \textbf{1024px} & \textbf{2048px} & \textbf{4096px} \\ \midrule
FLUX.1 VAE & \xmark & 0.3   & 0.7   & 2.6   & 37.0   & OOM    \\
FLUX.1 VAE & \cmark & 0.4   & 0.8   & 2.4   & 16.7   & OOM    \\
PiD       & \xmark & 12.6  & 12.8  & 13.6  & 16.5   & 28.6   \\
PiD       & \cmark & 10.3  & 10.3  & 10.9  & 13.0   & 22.5   \\ \bottomrule
\end{tabular}%
}
  \end{minipage}\hfill
  \begin{minipage}[t]{0.4\textwidth}
    \vspace{2pt}
    \centering
    \includegraphics[
      height=\dimexpr\ht\infcosttblbox+\dp\infcosttblbox\relax,
      width=\linewidth,
      keepaspectratio
    ]{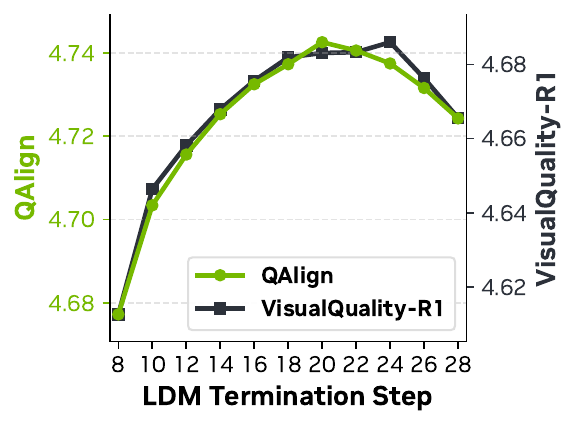}
    \captionof{figure}{\textbf{Image quality of \ours at different LDM termination step} for FLUX.1 [dev] ($28$ denoising steps in total).}
    \label{fig:best-ldm-exit-step}
  \end{minipage}
\end{figure*}

\subsection{Inference Cost}
\ours is efficient with respect to both latency and memory consumption. \Cref{tab:pid_latency_memory} reports the runtime and peak memory usage when decoding to different output resolutions. 
Compared with standard VAE decoding, \ours scales more favorably with resolution and consistently remains within a practical memory budget, requiring less than $30\,$GB even at 4K decoding. In contrast, the VAE decoder runs out of memory at resolutions of approximately $2500^2$ pixels on an $80\,$GB GPU without tiling.

\vspace{-1em}
\subsection{Ablation and Discussion}

\noindent\textbf{Ablation study.}
\Cref{tab:ablation} studies two key design choices: (i) the text-to-image pixel prior and (ii) the sigma-aware gating module. Removing the T2I prior severely degrades latent decoding (e.g., NIQE rises from $5.43$ to $7.79$, VisualQuality-R1~\cite{wu2025visualquality} drops from $4.649$ to $2.587$), showing that the high-resolution pretraining prior is crucial for generative SR. Removing the sigma-aware gate consistently worsens perceptual quality and fidelity, confirming that explicitly controlling the decoder’s reliance on the latent is beneficial. The gap is even larger in small-text reconstruction, where local fidelity is critical. Metrics are computed with the pre-distillation model over 25 denoising steps.

\begin{table}[t]
\centering
\caption{\textbf{Ablation study} on FLUX.1 [dev] decoding (left) and small-text reconstruction (right).}
\label{tab:ablation}
\resizebox{\columnwidth}{!}{%
\begin{tabular}{@{}c|cccccccc|ccc@{}}
\toprule
\textbf{Tasks} &
  \multicolumn{8}{c|}{\textbf{PiD\texttt{(24/28)}} (\textit{FLUX.1 [dev]})} &
  \multicolumn{3}{c}{\textbf{Small Text Reconstruction}} \\ \midrule
\multirow{2}{*}{Method} &
  \multirow{2}{*}{\begin{tabular}[c]{@{}c@{}}MUSIQ $\uparrow$\\ (paq2piq)\end{tabular}} &
  \multirow{2}{*}{NIQE$\downarrow$} &
  \multirow{2}{*}{DEQA $\uparrow$} &
  \multirow{2}{*}{MANIQA$\uparrow$} &
  \multirow{2}{*}{QA.$\uparrow$} &
  \multirow{2}{*}{\begin{tabular}[c]{@{}c@{}}Uni.$\uparrow$ \\ (IAA)\end{tabular}} &
  \multirow{2}{*}{\begin{tabular}[c]{@{}c@{}}Uni.$\uparrow$\\ (IQA)\end{tabular}} &
  \multirow{2}{*}{VQ-R1$\uparrow$} &
  \multirow{2}{*}{PSNR$\uparrow$} &
  \multirow{2}{*}{SSIM$\uparrow$} &
  \multirow{2}{*}{LPIPS$\downarrow$} \\
 &
   &
   &
   &
   &
   &
   &
   &
   &
   &
   &
   \\ \midrule
w/o T2I prior &
  59.52 &
  7.79 &
  2.649 &
  0.282 &
  2.58 &
  52.25 &
  46.93 &
  2.587 &
  17.21 &
  0.292 &
  0.584 \\
w/o sigma-aware gate &
  70.84 &
  5.84 &
  \textbf{4.292} &
  0.472 &
  4.75 &
  \textbf{63.49} &
  73.21 &
  4.647 &
  24.28 &
  0.956 &
  0.202 \\
Ours &
  \textbf{71.63} &
  \textbf{5.43} &
  4.289 &
  \textbf{0.487} &
  \textbf{4.75} &
  63.36 &
  \textbf{73.26} &
  \textbf{4.649} &
  \textbf{25.00} &
  \textbf{0.965} &
  \textbf{0.179} \\ \bottomrule
\end{tabular}%
}
\vspace{-1em}
\end{table}

\noindent\textbf{Optimal LDM termination step.}
We study the optimal number of LDM denoising steps for decoding partially denoised latents. Terminating too early leaves the latent semantically under-formed and harms output quality, while running full denoising leaves little room for the decoder to synthesize fine detail and increases overall latency. \Cref{fig:best-ldm-exit-step} shows the image quality of \ours at different LDM termination step for FLUX.1 [dev], which has $28$ denoising steps in total. It shows that the last 3 to 5 steps provide better visual quality than other steps. Other VAE latents show similar behaviors.

\noindent\textbf{LDM + \ours\ vs.\ native 2K generation.}
\begin{figure*}[t]
  \centering
  \setlength{\tabcolsep}{0pt}
  \renewcommand{\arraystretch}{0}
  \begin{tabular}{@{}cccc@{}}
    \imgwithlabel{0.25\textwidth}{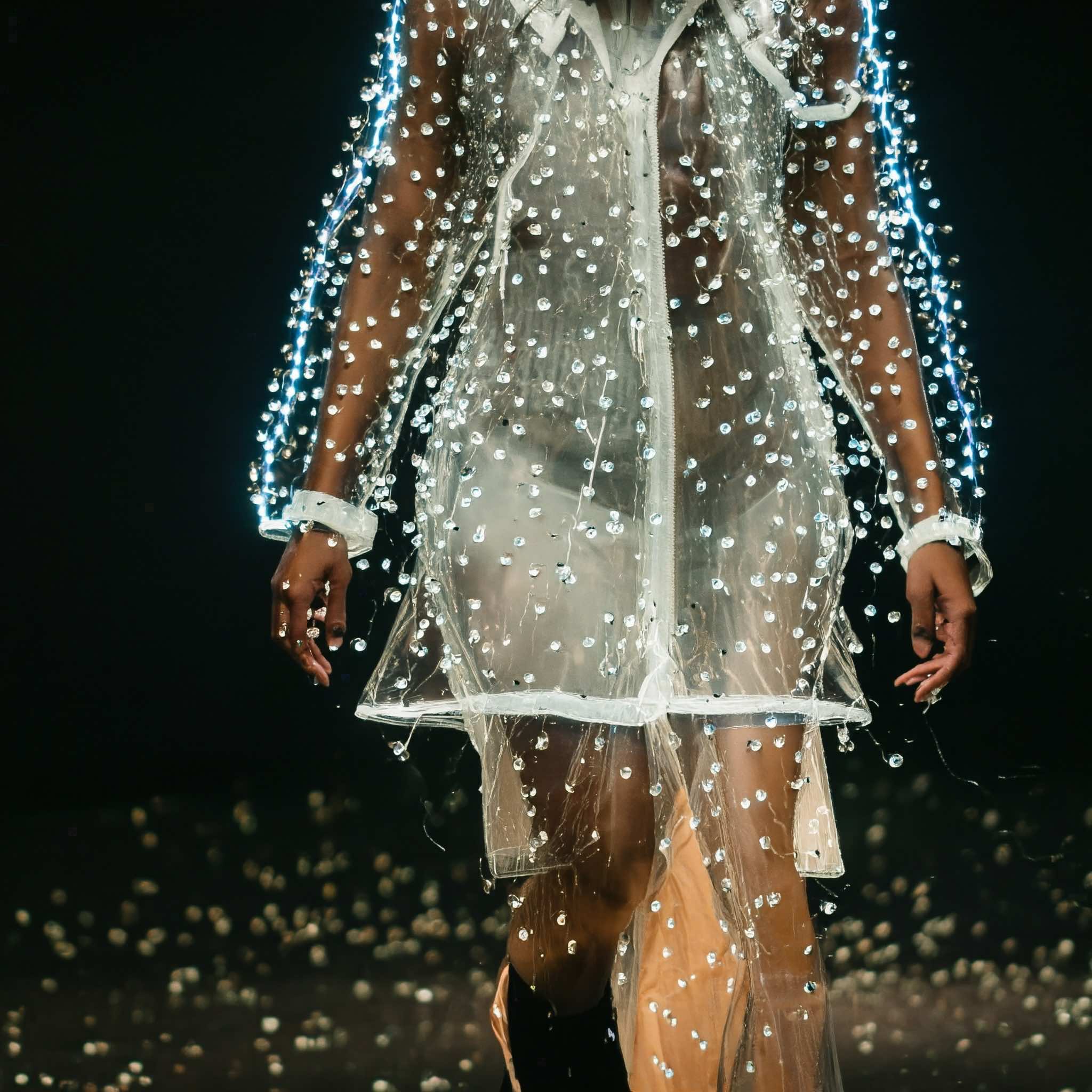}{13.3s} &
    \imgwithlabel{0.25\textwidth}{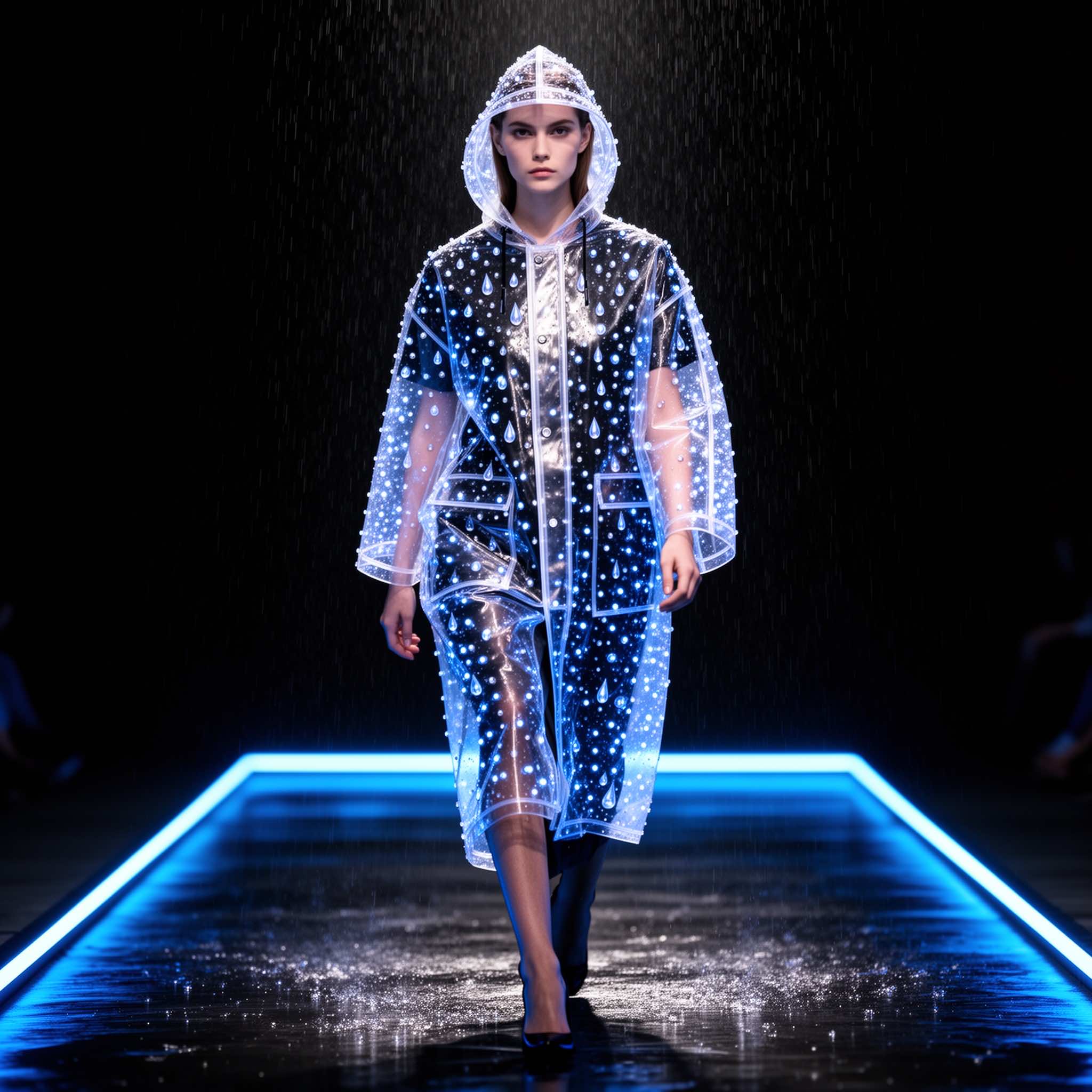}{102.2s} &
    \imgwithlabel{0.25\textwidth}{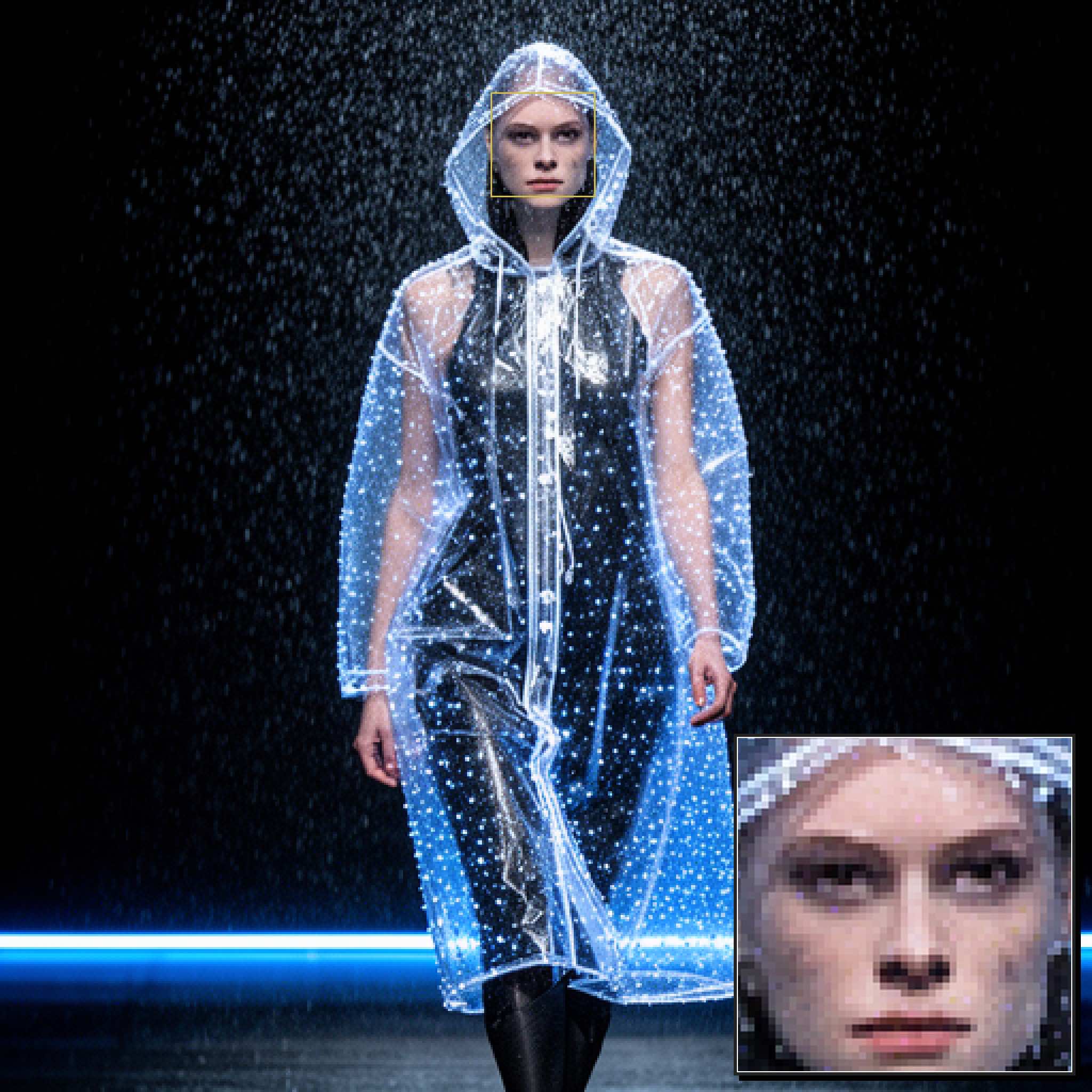}{6.6s} &
    \imgwithlabel{0.25\textwidth}{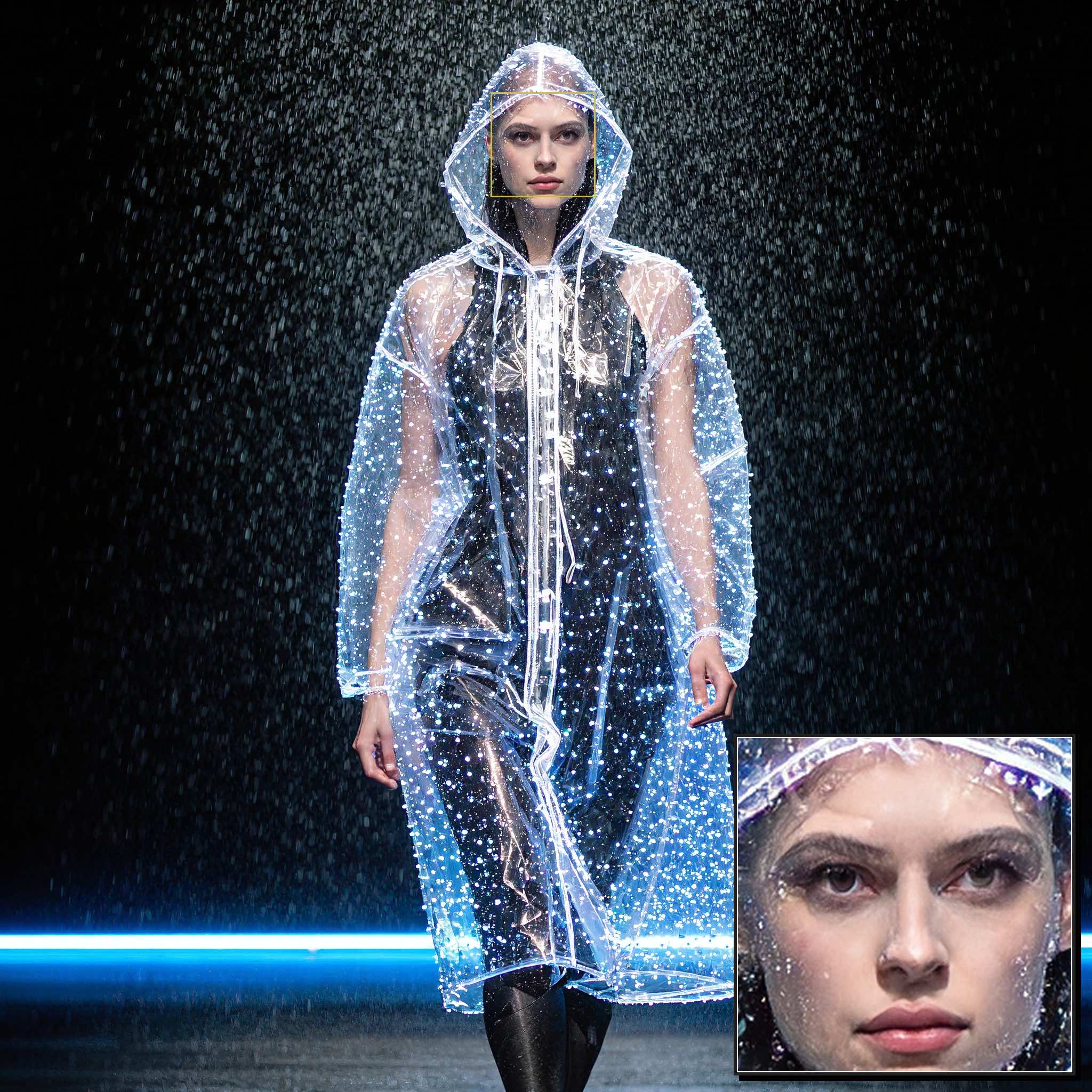}{7.1s}\\
    \imgwithlabel{0.25\textwidth}{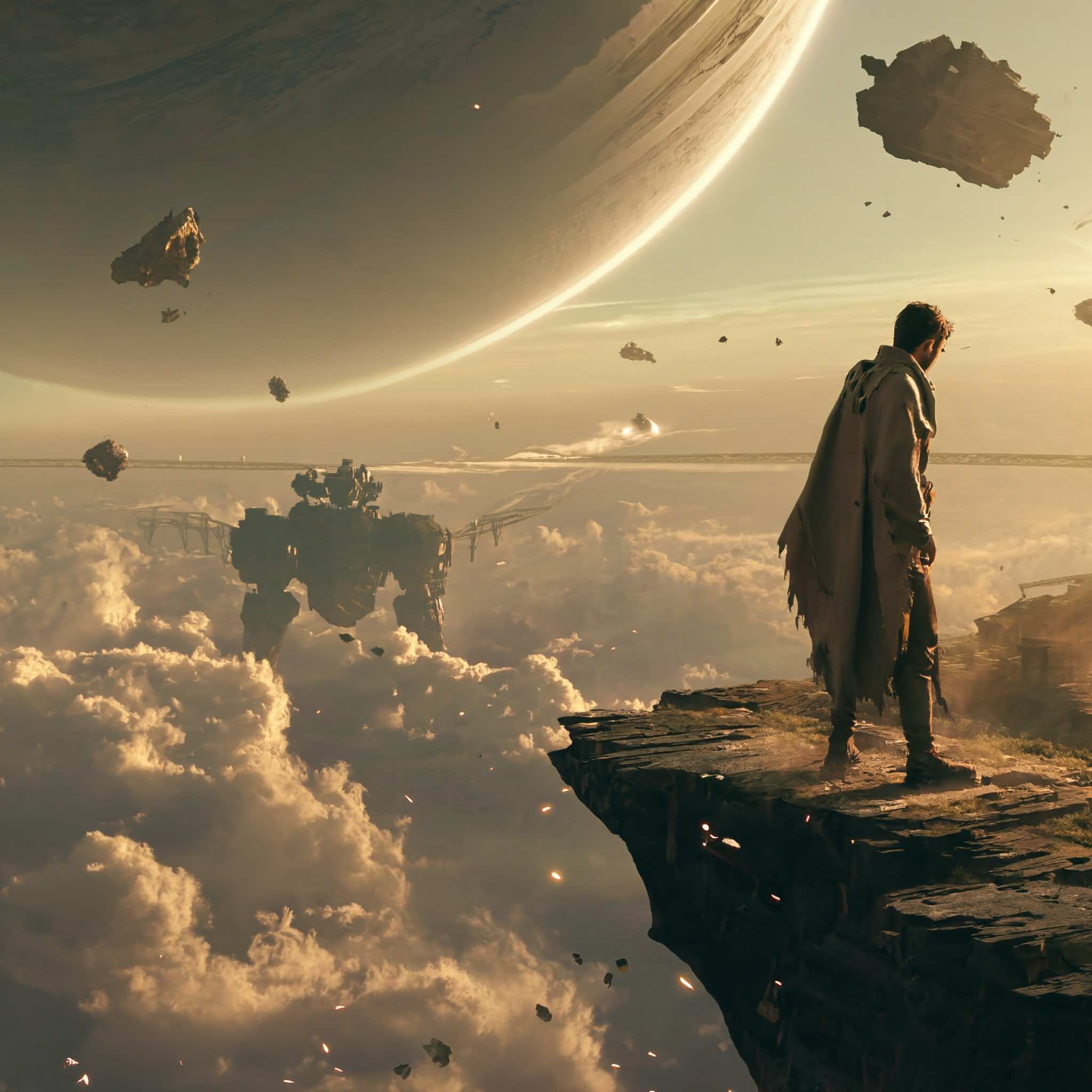}{13.3s} &
    \imgwithlabel{0.25\textwidth}{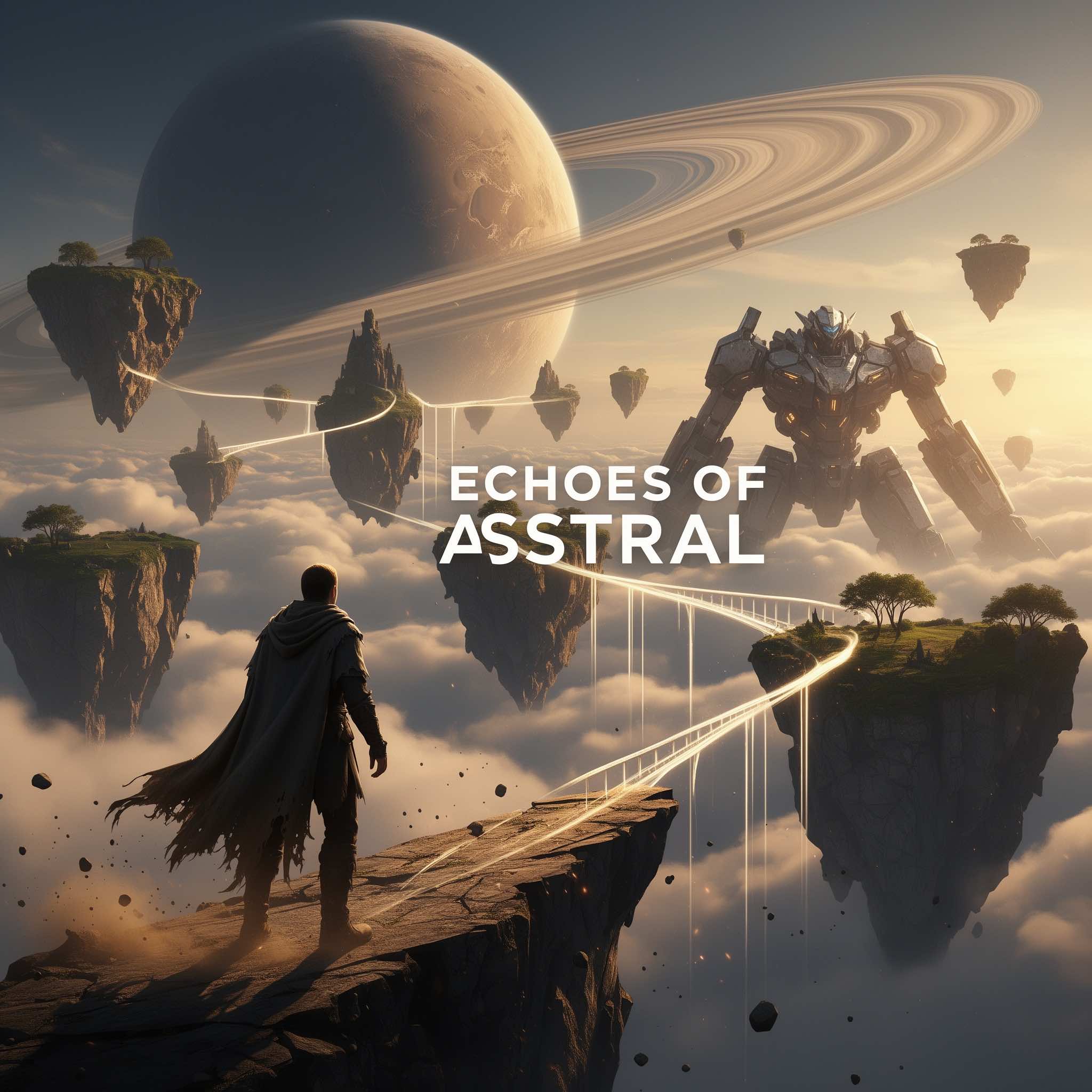}{102.2s} &
    \imgwithlabel{0.25\textwidth}{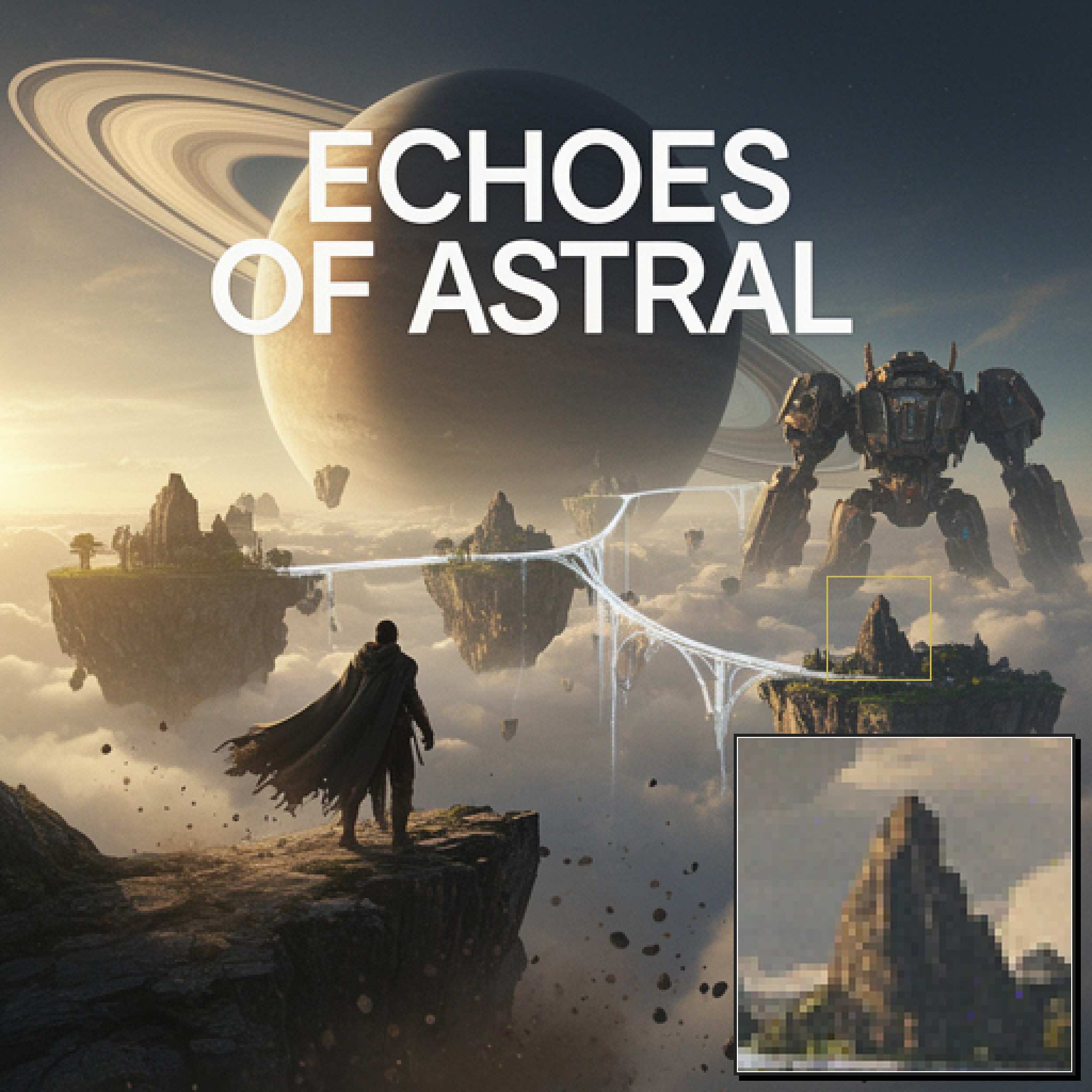}{6.6s} &
    \imgwithlabel{0.25\textwidth}{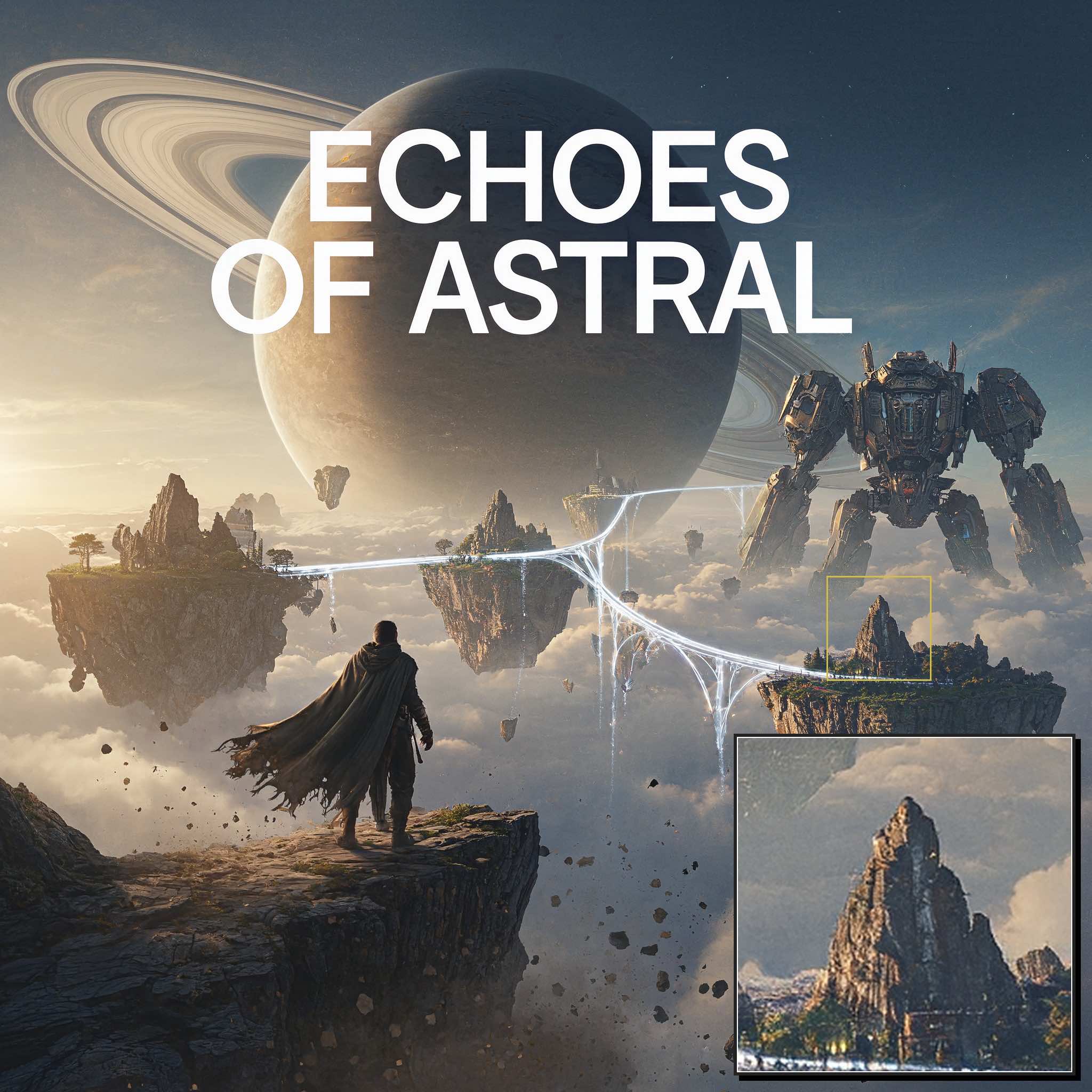}{7.1s}\\[2pt]
    \small PixelDiT (native $2048^2$) & \small FLUX.2  (native $2048^2$) & \small FLUX.2  $(512^2)$ + VAE Dec. & \small FLUX.2  $(512^2)$ + \ours
  \end{tabular}
  \caption{\textbf{Comparison with native $2048\times2048$ px  generation.} Coupling a low-resolution LDM with \ours substantially reduces inference time while maintaining image quality competitive to native high-resolution generation, and in some cases surpassing it in fine-grained details. Latency (in lower left corner) is measured on a single GB200 GPU without \texttt{torch.compile}.}
  \label{fig:pid-better-than-ldm}
  \vspace{-1em}
\end{figure*}
For high-resolution generation, running an LDM with \ours provides a practical alternative that achieves competitive visual quality with substantially lower sampling cost. We compare native 2K generation of FLUX.2, PixelDiT, and FLUX.2${(512^2)}$ + \ours. FLUX.2 and PixelDiT use 50 sampling steps at 2K or 512 resolution, while \ours runs 4 steps on a FLUX.2 generated latent of $512^2$ pixels. As a result (see~\Cref{fig:pid-better-than-ldm}), FLUX.2${(512^2)}$ + \ours only needs $7.1\,$s per sample, making it $1.87\times$ faster than PixelDiT and $\mathbf{14.3}\times$ faster than FLUX.2. Qualitatively, it shows better prompt following and visual quality than PixelDiT, and produces sharper details than native FLUX.2 in some cases. Given the substantial difference in model size (FLUX.2: 32B vs. PiD: 1.3B parameters) and inference cost (FLUX.2: 102.2s vs. w/ \ours 7.1s), we interpret these results as evidence that latent diffusion model equipped with \ours decoding strategy achieves favorable performance.

\FloatBarrier
\subsection{Extend to 4K Decoding}
\ours can easily extend to 4K decoding. We follow the same training recipe: first train a 4K resolution text-to-image pixel diffusion prior, then add latent projection adapter on the top of the backbone, and finally distill it into a 4-step student model. The experimental configuration and hyper-parameters stay the same as the 2K-resolution model mentioned in~\ref{sec:implementation-details}. We use 96 GB200 GPUs with context parallel 2 for text-to-image model and teacher model training, and 96 GB200 GPUs with context parallel 4 for distillation. We show qualitative results of 4K decoding in~\Cref{fig:4k-decoding}.

\vspace{-6pt}
\section{Conclusion}
\vspace{-6pt}

\label{sec:conclusion}

We introduced \ours, a pixel diffusion decoder that replaces reconstruction-oriented latent decoding with conditional pixel diffusion. This formulation unifies decoding and super-resolution in a single generative module, improving both speed and quality of high-resolution image synthesis from VAE and RAE-style latent spaces. \ours can also decode partially denoised latents, which allows early termination of base latent diffusion while preserving high-fidelity outputs.

\clearpage
\newpage
\appendix
\section{MLLM Judgment Details}
\label{app:mllm-judge-details}

For each paired sample $(x_i, y_i)$, where $x_i$ denotes the image from our
method and $y_i$ denotes the baseline image, we ask a closed-source MLLM to
select the image with better perceptual quality and details. The exact prompt
used for pairwise judgment is:

\begin{quote}
\small
\ttfamily
You are a strict image-quality judge. Two images follow, labeled Image A and Image B.

Compare them on perceptual quality and detail: sharpness, fine texture, noise,
compression artifacts, ringing, blocking, over-smoothing, over-sharpening,
edge halos.

Where to look (anchor your judgement on these regions):
\begin{itemize}
    \item Fine textures: hair, foliage, fabric weave, skin pores, small printed text.
    \item Edges: clean transitions with no ringing, halos, or stair-stepping.
    \item Flat / smooth regions: free of blocking, banding, color noise, or chroma smear.
    \item Repetitive patterns (grilles, bricks, screens): no moire / aliasing.
\end{itemize}

Position-bias rule: do NOT favor A or B because of their order. Mentally swap
A and B and re-check; only commit to a side if the same side still looks better.

You MUST pick a side -- no ties, no hedging. If you're genuinely unsure, pick
the side that wins on the largest number of the anchor regions above.

Output format (strict -- the FIRST line MUST be the verdict):

Line 1 must be exactly one of:

VERDICT: A \quad -- A is better in quality / detail

VERDICT: B \quad -- B is better in quality / detail

From line 2 onward: 1--3 sentences of justification in English.
\end{quote}

The MLLM output is parsed by extracting the first valid verdict from the
response, where only \texttt{VERDICT: A} and \texttt{VERDICT: B} are accepted.
Ties are not allowed. Invalid or unparseable responses are treated as failed
judgments and are excluded from the final win-rate calculation.

To reduce position bias, each image pair is evaluated twice with swapped input
order. In the first round, our result is shown as Image A and the baseline
result is shown as Image B. In the second round, the order is reversed: the
baseline result is shown as Image A and our result is shown as Image B. The raw
A/B verdicts are then converted into method-level preferences, i.e., whether
the MLLM prefers \textsc{Ours} or \textsc{Base} in each round.

For each baseline and each judge model, the win rate of our method is computed
over all valid judgments from the two rounds. Specifically, we count how many
times the MLLM selects our result and divide it by the total number of valid
non-failed judgments. Thus, each image pair can contribute up to two valid
votes, one from the original order and one from the swapped order.

\begin{figure}[p]
  \centering
  \includegraphics[width=\linewidth]{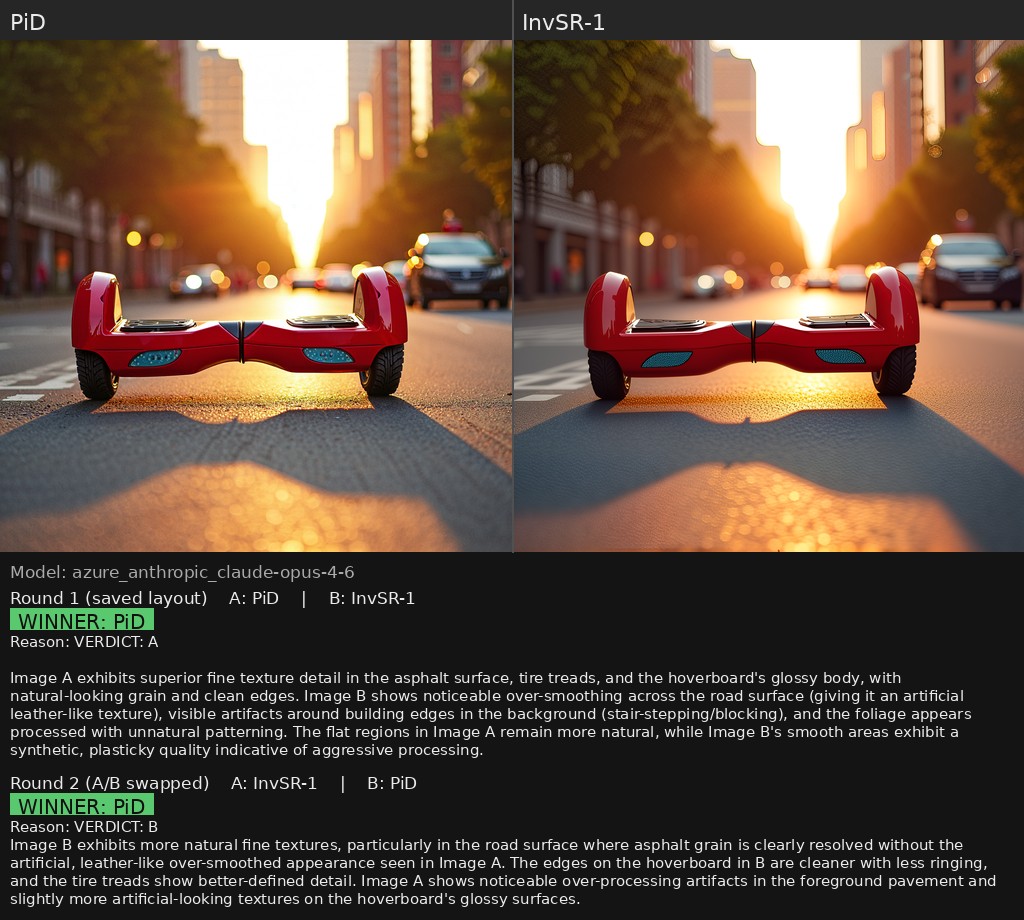}
  \caption{MLLM evaluation sample 1.}
  \label{fig:mllm-details-1}
\end{figure}

\begin{figure}[p]
  \centering
  \includegraphics[width=\linewidth]{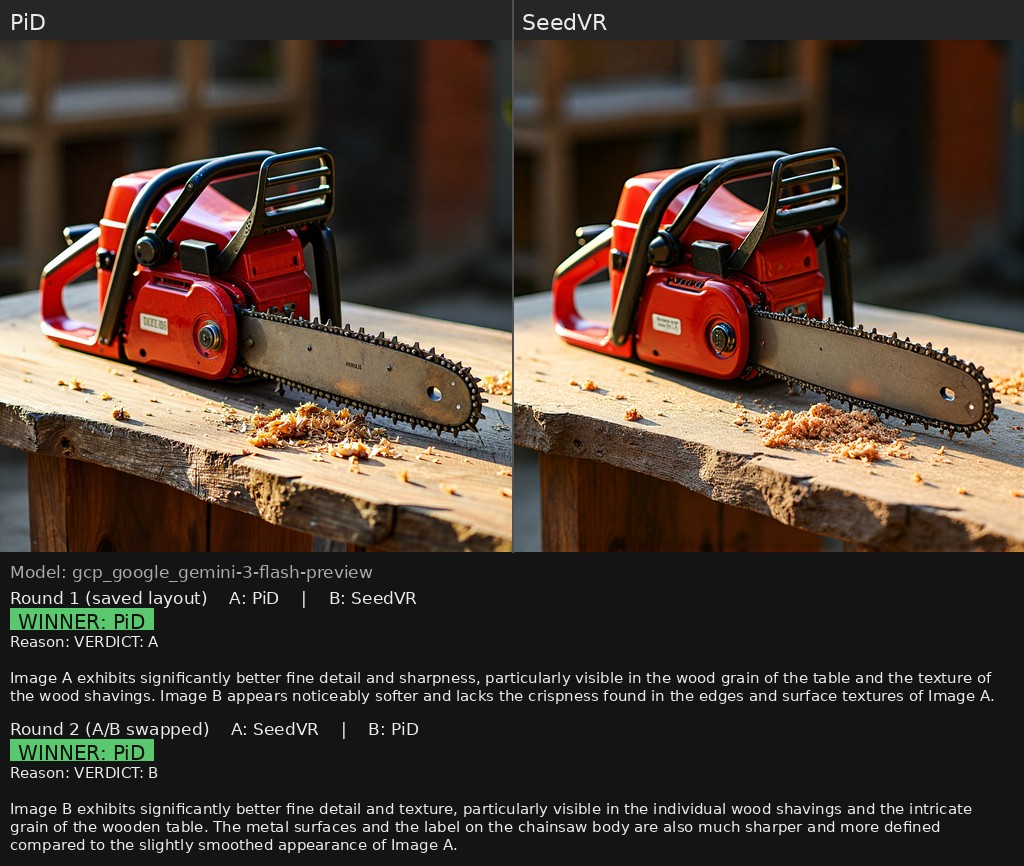}
  \caption{MLLM evaluation sample 2.}
  \label{fig:mllm-details-2}
\end{figure}

We also report the consistency rate to measure whether the MLLM gives a stable
preference under order swapping. For each image pair, we compare the
method-level preference from the two rounds. If both rounds are valid and they
select the same method, the pair is counted as consistent. A higher consistency rate indicates that the judge's preference is
less sensitive to the input order. Samples of evaluation can be found in~\Cref{fig:mllm-details-1,fig:mllm-details-2}.

\clearpage
{
\small
\bibliographystyle{abbrv}
\bibliography{bibs/intro,bibs/related_works,bibs/others}
}
\end{document}